\documentclass[aps,prapplied,reprint,superscriptaddress,floatfix,showkeys,longbibliography]{revtex4-2}

\usepackage{amsmath,amssymb,amsthm,amsfonts}
\usepackage{mathtools}
\usepackage{array}
\usepackage{textcomp}
\usepackage{url}
\usepackage{verbatim}
\usepackage{graphicx}
\usepackage{enumerate}
\usepackage{bm}
\usepackage{tabularx}
\usepackage{multirow}
\usepackage{threeparttable}
\usepackage[caption=false]{subfig}
\usepackage{psfrag}
\usepackage{color}
\usepackage[table,dvipsnames]{xcolor}
\usepackage[hidelinks]{hyperref}

\newtheoremstyle{colonstyle}%
    {3pt}{3pt}%
    {\normalfont}%
    {}%
    {\bfseries}%
    {:}%
    {.5em}%
    {}%

\theoremstyle{colonstyle}
\newtheorem{dfn}{Definition}

\newtheorem{proposition}{Proposition}
\newtheorem{corollary}{Corollary}
\newtheorem{lemma}{Lemma}

\newtheorem{ex}{Example}

\hypersetup{
  pdftitle={Architectural Principles for Parameter-Efficient Unitary Transformations\\
 in Coherent Optical Neural Networks},
  pdfauthor={Kazuo Aoyama, Hiroshi Sawada, Shota Kita, Akihiko Shinya, and Masaya Notomi},
  pdfsubject={Parameter-efficient coherent optical neural networks},
  pdfkeywords={silicon photonics, optical neural networks, parameterized unitary matrices, butterfly topology, Mach-Zehnder interferometers}
}
\AtEndDocument{%
  \special{ps: [ /Title (Architectural Principles for Parameter-Efficient Unitary Transformations in Coherent Optical Neural Networks)
  /Author (Kazuo Aoyama, Hiroshi Sawada, Shota Kita, Akihiko Shinya, and Masaya Notomi)
  /Creator (LaTeX with REVTeX 4.2)
  /Subject (Parameter-efficient coherent optical neural networks)
  /Keywords (silicon photonics, optical neural networks, parameterized unitary matrices, butterfly topology, Mach-Zehnder interferometers)
  /DOCINFO pdfmark}%
}

\begin{document}

\title{Architectural Principles for Parameter-Efficient Unitary Transformations\\
in Coherent Optical Neural Networks}

\author{Kazuo Aoyama}
\email{kazuo.aoyama@ntt.com}
\affiliation{NTT Communication Science Laboratories, NTT, Inc.}
\author{Hiroshi Sawada}
\affiliation{NTT Communication Science Laboratories, NTT, Inc.}
\author{Shota Kita}
\affiliation{NTT Basic Research Laboratories, NTT, Inc.}
\author{Akihiko Shinya}
\affiliation{NTT Basic Research Laboratories, NTT, Inc.}
\author{Masaya Notomi}
\affiliation{NTT Basic Research Laboratories, NTT, Inc.}

\begin{abstract}
Coherent optical neural networks based on interferometric architectures
offer a promising route toward high-throughput, energy-efficient computation
by exploiting the multiple degrees of freedom of light, 
including spatial, temporal, and spectral modes.
However, practical implementations are constrained by the number of tunable parameters
due to fabrication complexity, device footprint, and control overhead.
In such parameter-constrained settings,
achieving high expressive power remains a central challenge.
We identify two key architectural requirements
for maintaining high expressive power under these constraints:
(i) statistical uniformity in the implemented transformations and
(ii) strong mixing capability per parameter, which enables rapid and unbiased
redistribution of information across modes.
To realize these requirements,
we develop a logarithmic-depth butterfly-topology architecture 
whose fundamental building block is a single phase-shifter--beam-splitter pair,
in contrast to conventional rectangular mesh architectures
based on Mach-Zehnder interferometers.
Computational evaluations show that, particularly at restricted parameter
counts, the proposed architecture achieves lower matrix-reconstruction errors
and higher classification accuracies than conventional meshes in both
feedforward and recurrent networks.
We further quantify these design principles using graph-theoretic measures,
expected squared moduli of transformation-matrix entries, and
matrix-reconstruction error. The resulting framework provides a quantitative
and physically interpretable basis
for parameter-efficient optical neural networks.
\end{abstract}


\maketitle

\section{Introduction}\label{sec:intro}
Optical neural networks (ONNs) offer a promising alternative 
to conventional electronic implementations by enabling high-throughput computation 
with potentially reduced energy consumption \cite{lin-ozcan,bogaerts,wetzstein}.
Among these, coherent ONNs based on interferometric architectures 
in silicon photonics are particularly promising \cite{shen}, 
as they leverage multiple degrees of freedom of light, 
including spatial, temporal, and spectral modes.
In particular, exploiting spectral modes enables parallel processing \cite{kita}
and improved accuracy \cite{ikeda_cleo}, representing a key advantage of 
coherent ONNs.
However, practical implementations remain constrained by the number of tunable parameters
due to fabrication complexity, device footprint, and control overhead.
As a result, existing coherent ONN implementations have so far been limited 
to relatively small-scale systems \cite{band,tang-chen,zhang-nanyang}.

Coherent ONNs have previously relied on interferometric mesh architectures 
to implement configurable linear transformations \cite{reck,clements}, 
most notably rectangular meshes.
The fundamental building block of these architectures is 
the programmable Mach-Zehnder interferometer (MZI) unit,
which consists of \emph{two} tunable phase shifters (PSs) and two beam splitters (BSs)
with fixed power-splitting ratios \cite{clements,pai,fang}.
We refer to this programmable unit simply as an MZI.
By leveraging well-established decomposition algorithms \cite{reck,clements}, 
such meshes can realize arbitrary unitary transformations. 
For example, 
a rectangular mesh with $n$ input and $n$ output ports can implement 
an arbitrary $n\!\times\!n$ unitary matrix 
using $n$ stages of MZIs followed by a final stage of $n$ phase shifters \cite{clements}.
Despite their universality, these architectures require $n^{2}$ tunable parameters,
resulting in substantial implementation overhead for a single linear unit.

In parameter-efficient neural networks, however, 
the ability of a single linear unit to implement an arbitrary unitary transformation
is not always required.
This can be understood from two perspectives:
(1) direct optimization of parameters within a given architecture 
without relying on decomposition algorithms,
and (2) the use of multiple linear units 
whose combined effect can realize complex transformations,
as discussed in Sections~\ref{sec:pre} and \ref{sec:exp}. 
Related nonuniversal architectures, including shallow MZI-based designs with
butterfly topologies and truncated meshes, have been explored primarily to
mitigate device imperfections \cite{fang}.
However, these architectures were not designed specifically for
parameter-efficient computation:
their reliance on MZIs with two tunable parameters limits efficiency,
and truncated meshes further reduce mixing across modes.

In this work, we explore architectures capable of implementing unitary matrices 
with high \emph{expressive power} using fewer than $n^2$ tunable parameters 
per linear unit.
The expressive power of a parameterized unitary matrix is quantified by
the average reconstruction error between Haar-random matrices \cite{zyczkowski} 
and optimized parameterized matrices,
as defined in Section~\ref{ssec:reconst}.
We identify two key architectural requirements
for achieving high expressive power in parameter-constrained settings:
(i) statistical uniformity in the implemented transformations and
(ii) strong mixing capability per parameter, which enables rapid and unbiased
redistribution of information across all modes.
The notion of statistical uniformity is related to prior work on
Haar initialization in mesh architectures,
which enables uniform sampling over the unitary group for fully parameterized 
transformations \cite{pai}.
However, such approaches were developed for universal meshes
and do not provide a framework for achieving statistical uniformity 
in parameter-constrained or non-universal architectures.
By contrast, strong mixing capability per parameter is not captured by existing schemes 
and becomes critical in reduced-parameter settings,
where each parameter must contribute effectively to global mode coupling.
These requirements provide a principled foundation 
for designing parameter-efficient ONNs.

To realize these requirements,
we propose the fine-grained butterfly (FGB) architecture shown in
Fig.~\ref{fig:8in_bfly}. This architecture employs the minimal
phase-shifter--beam-splitter (PSBS) pair shown in Fig.~\ref{fig:8in_bfly}(b)
as the fundamental building block of each stage.
Unlike an MZI, which constitutes a complete two-mode interferometer, 
a PSBS does not form an interferometer on its own 
but instead provides a minimal parameterized mixing operation.
The sequential stages are connected in a butterfly topology.
For $n$ inputs ($n\!=\!2^{t}$, $t\!\in\!\mathbb{N}$), 
the FGB architecture mixes input modes across all outputs 
within $\log_{2}n$ stages using only $(n/2)\log_{2}n$ tunable parameters.
Statistical uniformity follows from structural properties of the topology,
including an equal number of PSBS units per stage 
and a unique path from each input port to every output port.

Our contributions are fourfold:
\begin{enumerate}
\item
We identify two key architectural requirements for achieving high expressive power 
under a limited number of tunable parameters:
(i) statistical uniformity in the transformation of input modes
and (ii) strong mixing capability per parameter.
We quantify these requirements using graph-theoretic measures, including
path-count distributions and reachability.
\item
We propose the FGB architecture
based on a butterfly topology with PSBS building blocks:
the topology promotes statistical uniformity, while the PSBS blocks
enhance mixing capability per parameter.
\item
Through computational experiments on Haar-random matrix reconstruction,
we demonstrate the high expressive power of the parameterized unitary
transformations implemented by the FGB architecture.
\item
Through computational experiments on image-classification tasks
using both feedforward and recurrent networks,
we show that
the FGB architecture achieves higher classification accuracy than
conventional rectangular mesh architectures at restricted parameter counts.
\end{enumerate}

The remainder of this paper is organized as follows.
Section \ref{sec:pre} reviews conventional ONNs,
including training strategies and MZI-based mesh architectures.
Section \ref{sec:limit} analyzes their limitations 
using the graph-theoretic approach.
Section \ref{sec:prop} introduces the proposed FGB architecture.
Section \ref{sec:exp} presents computational evaluations,
including matrix reconstruction and image classification tasks.
Section \ref{sec:disc} discusses the results in terms of statistical uniformity,
mixing capability per parameter, reachability, and prospects for physical
implementation.
Finally, Section~\ref{sec:conc} concludes the paper and outlines directions
for future research.

\section{Conventional Optical Neural Networks}\label{sec:pre}
This section reviews widely used approaches for realizing MZI-based
interferometric ONNs in silicon photonics, including training methods and mesh
architectures.
Here, a mesh architecture is a gridlike topology in which the terminals of a
processing unit, such as an MZI, connect directly to those of physically
adjacent units in the subsequent stage without waveguide crossings.
Examples include triangular \cite{reck}, rectangular \cite{clements},
diamond \cite{shokraneh}, and Bokun \cite{mojaver} architectures.
Particular attention is given to conventional rectangular mesh architectures
\cite{clements,fang},
which serve as the baseline for subsequent comparisons.
\subsection{ONN Training}\label{ssec:train}
A two-step method is commonly used to train ONN linear units with arbitrary
weight matrices \cite{shen}. Such units are constructed from unitary matrices
implemented by triangular or rectangular meshes. In the first step, an
arbitrary weight matrix $W$ is optimized
using standard techniques and then 
factorized via singular value decomposition (SVD) 
as $W\!=\!U\Sigma V^{\dagger}$,
where $U$ and $V$ are unitary matrices, 
$V^{\dagger}$ denotes the conjugate transpose of $V$, and
$\Sigma$ is a diagonal matrix containing the singular values.
In the second step, 
the unitary matrices $U$ and $V^{\dagger}$ are decomposed and mapped 
onto optical circuits configured as meshes based on MZIs,
which can implement arbitrary unitary transformations \cite{clements}.
The diagonal matrix $\Sigma$ is physically realized 
using attenuators combined with phase shifters.

By contrast,
we use direct parameter optimization, which can be applied to arbitrary
ONN architectures \cite{jing,aoyama,sawada_aaai,ikeda} without decomposition
or circuit-mapping algorithms. Specifically,
given a predefined linear architecture, 
such as a mesh configuration that directly represents 
a parameterized unitary transformation,
we optimize the phase-shifter parameters end to end.
Such direct parameter-optimization methods for parameterized unitary matrices
have also been employed in recurrent neural-network (RNN) models
\cite{arjovsky}.
\subsection{Conventional Mesh Architectures}\label{ssec:mesh}
\begin{figure}[t]
\begin{center}
	\psfrag{A}[c][c][1.0]{(a)}
	\psfrag{B}[c][c][1.0]{(b)}
	\psfrag{C}[c][c][1.0]{(c)}
	\psfrag{X}[r][r][0.9]{$n\!=\!8$}
	\psfrag{Y}[c][c][0.9]{$m\!=\!8$}
	\psfrag{Z}[c][c][0.9]{$m\!=\!4\!<\!n$}
	\psfrag{P}[c][c][0.85]{PS$_1\,(\phi)$}
	\psfrag{Q}[l][l][0.85]{PS$_2\,(\theta)$}
	\psfrag{R}[c][c][0.85]{BS$_1$}
	\psfrag{S}[c][c][0.85]{BS$_2$}
	\includegraphics[width=74mm]{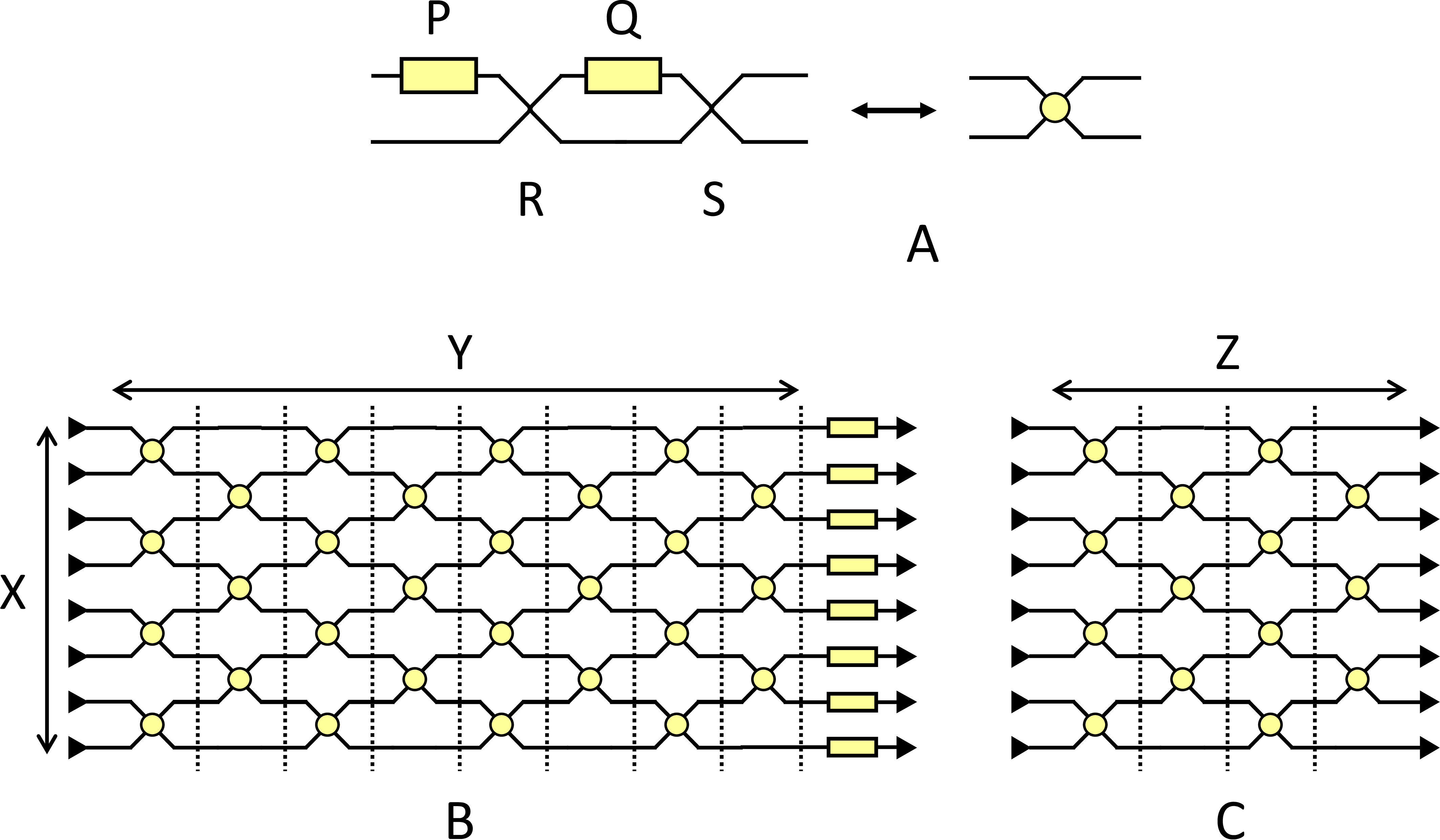}
\end{center}
\vspace*{-2mm}
\caption{
Rectangular mesh configurations:
(a) a conventional MZI circuit composed of
two cascaded pairs of a programmable phase shifter and a beam splitter (left)
and its symbol (right),
(b) a programmable eight-input ($n\!=\!8$), eight-stage ($m\!=\!n\!=\!8$) rectangular mesh
followed by a phase-shifter stage,
and (c) a truncated rectangular mesh with four stages ($m\!=\!4\!<\!n$).
}\label{fig:mesh}
\vspace*{-2mm}
\end{figure}

Programmable MZIs have been widely adopted as fundamental components 
in optical switch fabrics for communication systems \cite{lee,bao}.
An MZI functions as a switch between output ports when its phase parameter is
set to binary values.
Beyond binary operation, 
continuous tuning of parameters enables general interferometric transformations 
\cite{perez,bogaerts}.  
This parameterization is the basis of the MZI used in ONNs.

Figure~\ref{fig:mesh}(a) shows a programmable MZI used in ONNs,
which consists of two phase shifters (PS$_1$ and PS$_2$) and 
two beam splitters (BS$_1$ and BS$_2$) with fixed 50:50 power-splitting 
ratios \cite{clements}.\footnote{
Alternative configurations include 
interleaved phase shifters and beam splitters \cite{pai} 
and phase shifters placed on both internal arms \cite{bogaerts}.
}
The phase angles of 
external phase shifter PS$_1$ and internal phase shifter PS$_2$ 
are tunable parameters, 
with $\phi\!\in\![0,2\pi)$ and $\theta\!\in\![0,\pi]$, respectively.
The corresponding transfer matrix $T_{(\text{MZI})}(\phi,\theta)$ 
is given by
\begin{align}
T_{(\text{MZI})}(\phi,\theta) &= 
\underbrace{\frac{1}{\sqrt{2}}
\setlength{\arraycolsep}{2.5pt}
\begin{pmatrix}1&i\\i&1\end{pmatrix}}_{\text{BS}_2}\,
\underbrace{
\setlength{\arraycolsep}{2.5pt}
\begin{pmatrix}e^{i\theta}&0\\0&1\end{pmatrix}}_{\text{PS}_2}\,
\underbrace{\frac{1}{\sqrt{2}}
\setlength{\arraycolsep}{2.5pt}
\begin{pmatrix}1&i\\i&1\end{pmatrix}}_{\text{BS}_1}\,
\underbrace{
\setlength{\arraycolsep}{2.5pt}
\begin{pmatrix}e^{i\phi}&0\\0&1\end{pmatrix}}_{\text{PS}_1}   \notag\\
&= ie^{i\frac{\theta}{2}}
\begin{pmatrix}e^{i\phi}\sin{\frac{\theta}{2}}& \cos{\frac{\theta}{2}}\\
e^{i\phi}\cos{\frac{\theta}{2}}& -\sin{\frac{\theta}{2}}
\end{pmatrix} \, , \label{eq:mzi}
\end{align}
where $i$ denotes the imaginary unit.

Among mesh architectures, we focus on conventional rectangular meshes
\cite{clements,fang}. A rectangular mesh consists of cascaded programmable MZIs
arranged in a layer topology. 
Figure~\ref{fig:mesh}(b) illustrates a rectangular mesh
with eight input ports and eight stages, 
where solid circles and lines represent MZIs and waveguides, respectively.
The mesh architecture can implement an arbitrary $8\!\times\!8$ unitary matrix,
followed by a programmable phase-shifter stage featuring a phase shifter
at every port. 
Generalizing this to an arbitrary $n\!\times\!n$ unitary transformation 
requires $n$ stages of MZIs followed by $n$ output phase shifters, 
forming a programmable diagonal unitary matrix \cite{clements}.

One option for reducing the number of phase shifters and the device footprint
is a truncated rectangular mesh architecture \cite{fang}.
Figure~\ref{fig:mesh}(c) illustrates an example
with eight input ports and four stages.
This architecture employs a shallower network and thus fewer parameters,
but it no longer realizes arbitrary unitary transformations. 
Furthermore, it can no longer mix input signals across all modes.

\section{Limitations of Conventional Meshes}\label{sec:limit}
\begin{figure}[t]
\begin{center}
	\psfrag{a}[r][r][0.9]{$x_0$} \psfrag{b}[r][r][0.9]{$x_1$}
	\psfrag{c}[r][r][0.9]{$x_2$} \psfrag{d}[r][r][0.9]{$x_3$}
	\psfrag{A}[l][l][0.9]{$y_0$} \psfrag{B}[l][l][0.9]{$y_1$}
	\psfrag{C}[l][l][0.9]{$y_2$} \psfrag{D}[l][l][0.9]{$y_3$}
	\psfrag{E}[c][c][0.9]{$0$}
	\psfrag{F}[c][c][0.9]{$1$}
	\psfrag{G}[c][c][0.9]{$2$}
	\psfrag{H}[c][c][0.9]{$3$}
	\psfrag{W}[c][c][0.9]{Stage number}
	\psfrag{P}[c][c][0.8]{${\bm p}_{00}$}
	\psfrag{Q}[c][c][0.8]{${\bm p}_{02}$}
	\psfrag{R}[c][c][0.8]{${\bm p}_{11}$}
	\psfrag{S}[c][c][0.8]{${\bm p}_{20}$}
	\psfrag{T}[c][c][0.8]{${\bm p}_{22}$}
	\psfrag{U}[c][c][0.8]{${\bm p}_{31}$}
	\psfrag{X}[c][c][1.0]{(a)}
	\psfrag{Y}[c][c][1.0]{(b)}
	\includegraphics[width=75mm]{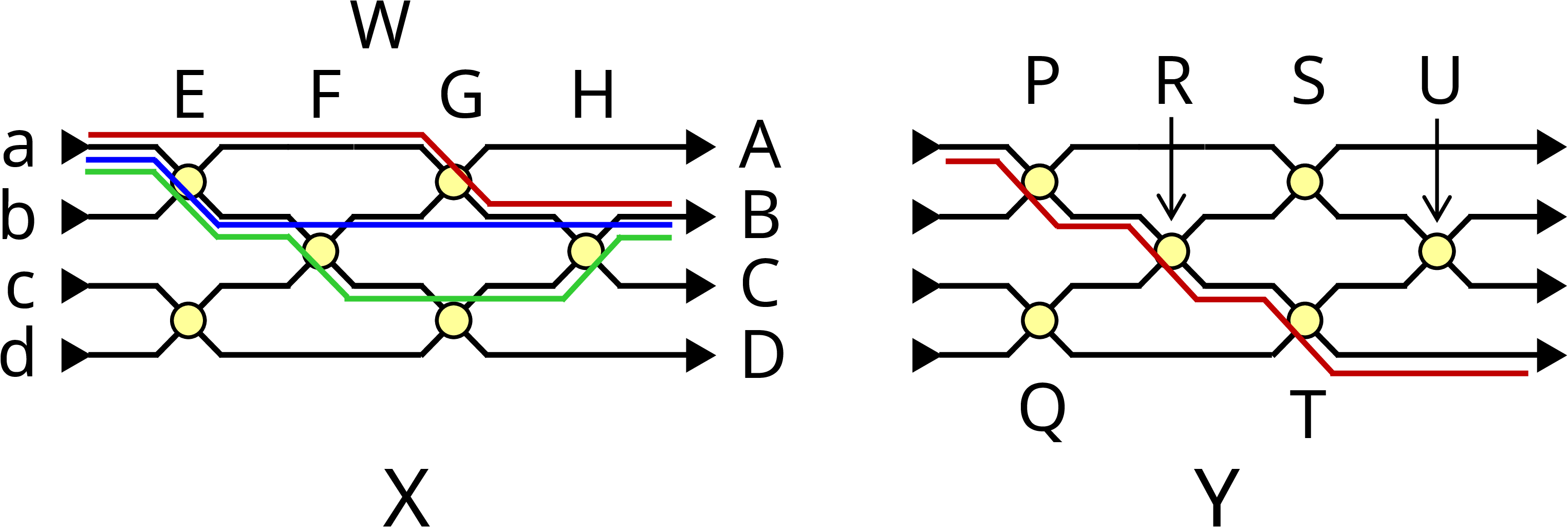}
\end{center}
\vspace*{-2mm}
\caption{
Paths between input and output ports in a mesh architecture with four input
ports and four stages:
(a) three paths from $x_0$ to $y_1$ and (b) a single path from $x_0$ to $y_3$.
${\bm p}_{su}$ denotes the MZI at the $s$th stage and the $u$th position
from the top; its two parameters are denoted by $(\phi_{su},\theta_{su})$.
}\label{fig:meshpath}
\end{figure}

Mesh architectures exhibit three primary limitations:
\begin{enumerate}[(1)]
\item Nonuniform path multiplicities across input-output pairs.
\item Lack of \emph{reachability} in truncated meshes,
defined below.
\item Inefficiency of the MZI as the fundamental building block.
\end{enumerate}

\begin{dfn}[Reachability] \label{dfn:reachability}
A network with distinct input and output terminals
is said to satisfy the \emph{reachability} property if there exists 
at least one path from each input terminal to every output terminal.
\end{dfn}

\subsection{Multiple Paths Between an Input-Output Pair}
We consider path multiplicity in rectangular mesh architectures
and its adverse effect on the statistical uniformity of the implemented
transformations.
Such meshes inherently allow 
multiple distinct paths between a given input-output pair,
depending on the indices of the ports.
For example, as illustrated in Fig.~\ref{fig:meshpath},
(a) three distinct paths exist 
from the input port $x_0$ to the output port $y_1$,
whereas (b) only a single path exists from $x_0$ to $y_3$.
The subscripts of $x_h$ and $y_j$ denote the input and output indices, 
respectively.
Furthermore, in a rectangular mesh with 32 inputs and 32 outputs ($n\!=\!32$),
the path multiplicity increases substantially and becomes highly nonuniform,
with paths concentrated toward the central output ports,
as shown in Fig.~\ref{fig:pathcount}.
The path counts for general $n$ are analyzed in Appendix~\ref{app_sec:path}.
This imbalance causes the circuit parameters to contribute unevenly across
input-output pairs, thereby degrading statistical uniformity in the
transformation.

\begin{figure}[t]
\begin{center}
	\psfrag{X}[c][c][1.0]{Output index}
	\psfrag{Y}[c][c][1.0]{
		\begin{picture}(0,0)
			\put(0,0){\makebox(-6,36)[c]{
				Number of paths}} 
			\put(-15,0){\makebox{($\times 10^{8}$)}}
		\end{picture}
	}
	\psfrag{A}[c][c][0.9]{0} \psfrag{B}[c][c][0.9]{5}
	\psfrag{C}[c][c][0.9]{10} \psfrag{D}[c][c][0.9]{15}
	\psfrag{E}[c][c][0.9]{20} \psfrag{F}[c][c][0.9]{25}
	\psfrag{G}[c][c][0.9]{30}
	\psfrag{H}[r][r][0.9]{0} \psfrag{J}[r][r][0.9]{1}
	\psfrag{K}[r][r][0.9]{2} \psfrag{L}[r][r][0.9]{3}
	\psfrag{M}[r][r][0.9]{4}
	\psfrag{P}[r][r][0.9]{Input index: 2}
	\psfrag{Q}[r][r][0.9]{8}
	\psfrag{R}[r][r][0.9]{16}
	\includegraphics[width=55mm]{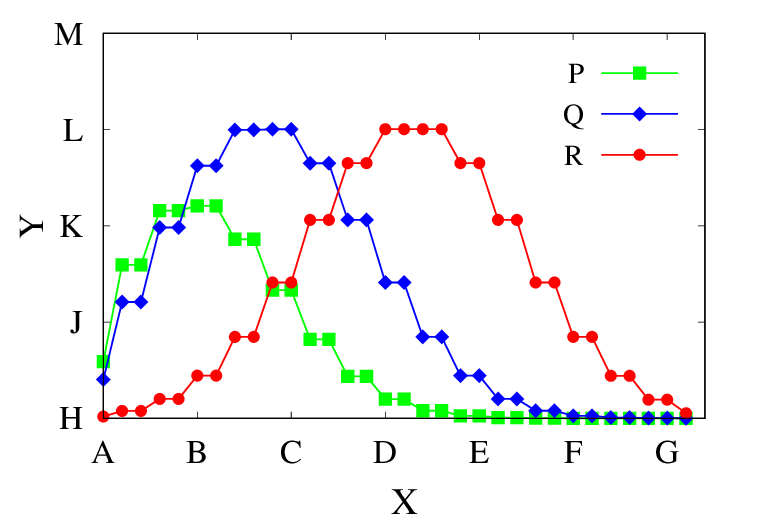}
\end{center}
\vspace*{-2mm}
\caption{Numbers of paths from the specified input ports (indices 2, 8, and 16)
to the output ports of a mesh architecture with 32 inputs and 32 outputs.
}\label{fig:pathcount}
\end{figure}

From a graph-theoretic perspective,
we consider a parameterized unitary matrix $G$ implemented by the mesh,
where the $(j,h)$ entry is denoted by $g_{jh}$.
Let $P_{(j,h)}$ denote the set of paths from the $h$th input terminal to 
the $j$th output terminal, 
and let $p\!\in\!P_{(j,h)}$ represent an individual path.
Define $F(p)$ as the number of MZIs along path $p$.
When the parameters of $G$ are initialized uniformly at random,
the expected squared modulus of $g_{jh}$ is expressed as
\begin{equation}
\mathbb{E}\left[\,|g_{jh}|^2\,\right] = 
\sum_{p\in P_{(j,h)}} \frac{1}{2^{\,F(p)}}\, .
\label{eq:expected}
\end{equation}
For paths of comparable length, this expectation increases with the path
multiplicity $|P_{(j,h)}|$, leading to nonuniformity across input-output pairs
and thus degrading statistical uniformity in the transformation.
A detailed derivation of Eq.~(\ref{eq:expected}) is provided 
in Appendix~\ref{app_sec:heatmap}.

We compare the analytical $n\!\times\!n$ matrix $\widehat{M}$, whose
$(j,h)$ entry is $\mathbb{E}[|g_{jh}|^2]$ as given by
Eq.~(\ref{eq:expected}), with the corresponding computational average
$\overline{M}$ over $10^{6}$ mesh instances whose phase-shifter parameters
are sampled uniformly at random.
We use the root-mean-square error (RMSE) to assess the agreement
between the analytical matrix $\widehat{M}$ and the computational average
$\overline{M}$:
\begin{equation}
\text{RMSE}(\overline{M},\widehat{M}) = 
\frac{1}{n}\,\| \overline{M} -\widehat{M} \|_F\, ,
\label{eq:rmse}
\end{equation}
where $\|\overline{M}\!-\!\widehat{M}\|_F$ is the Frobenius norm of the
residual matrix.

\begin{figure}[t]
  \centering
	\captionsetup[subfloat]{margin={3pt,0pt}}
	\subfloat[$n\!=\!8$ Anal.]{
  	  \includegraphics[width=0.18\columnwidth]{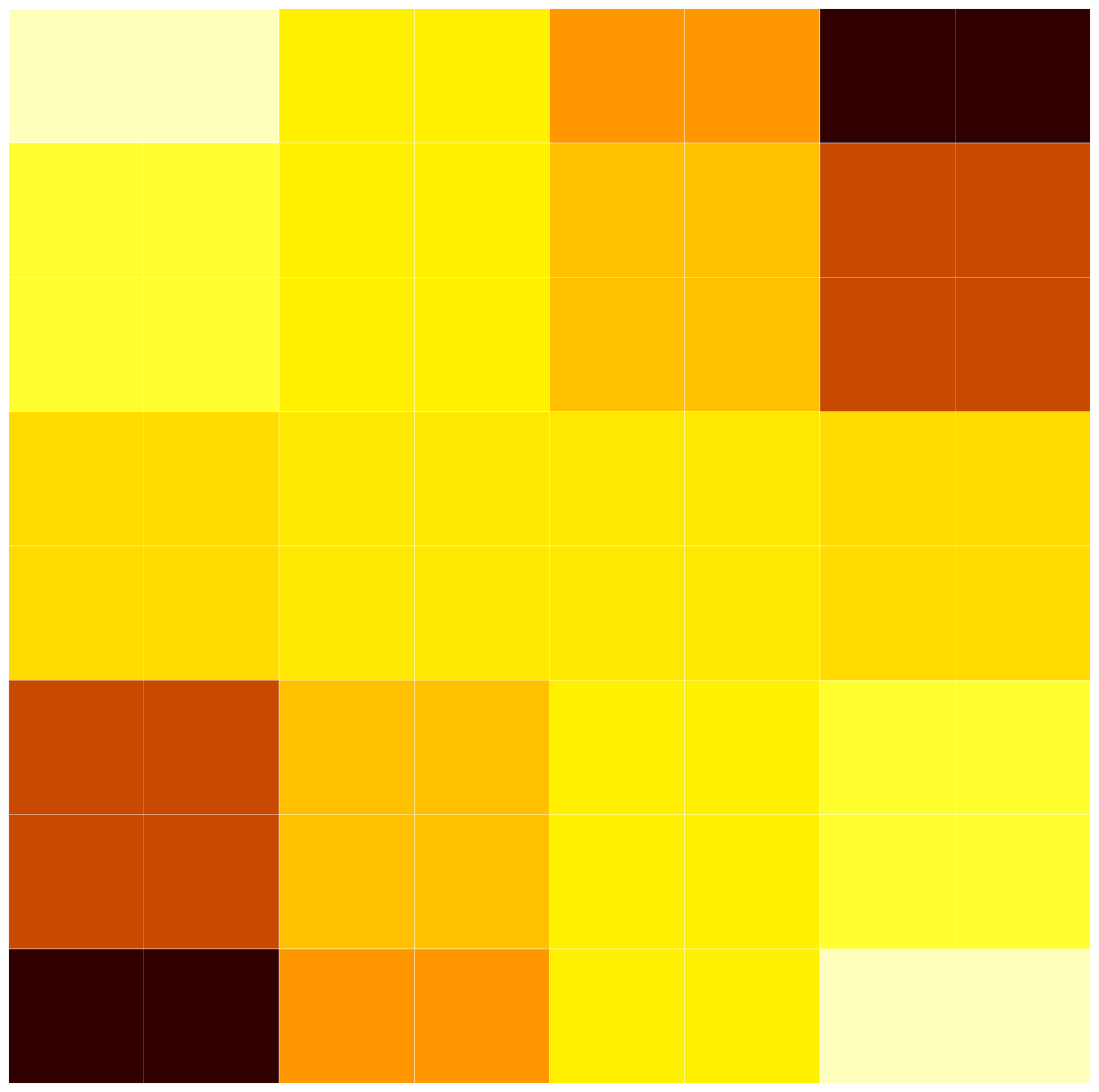}
	}
	\subfloat[$n\!=\!16$]{
  	  \includegraphics[width=0.18\columnwidth]{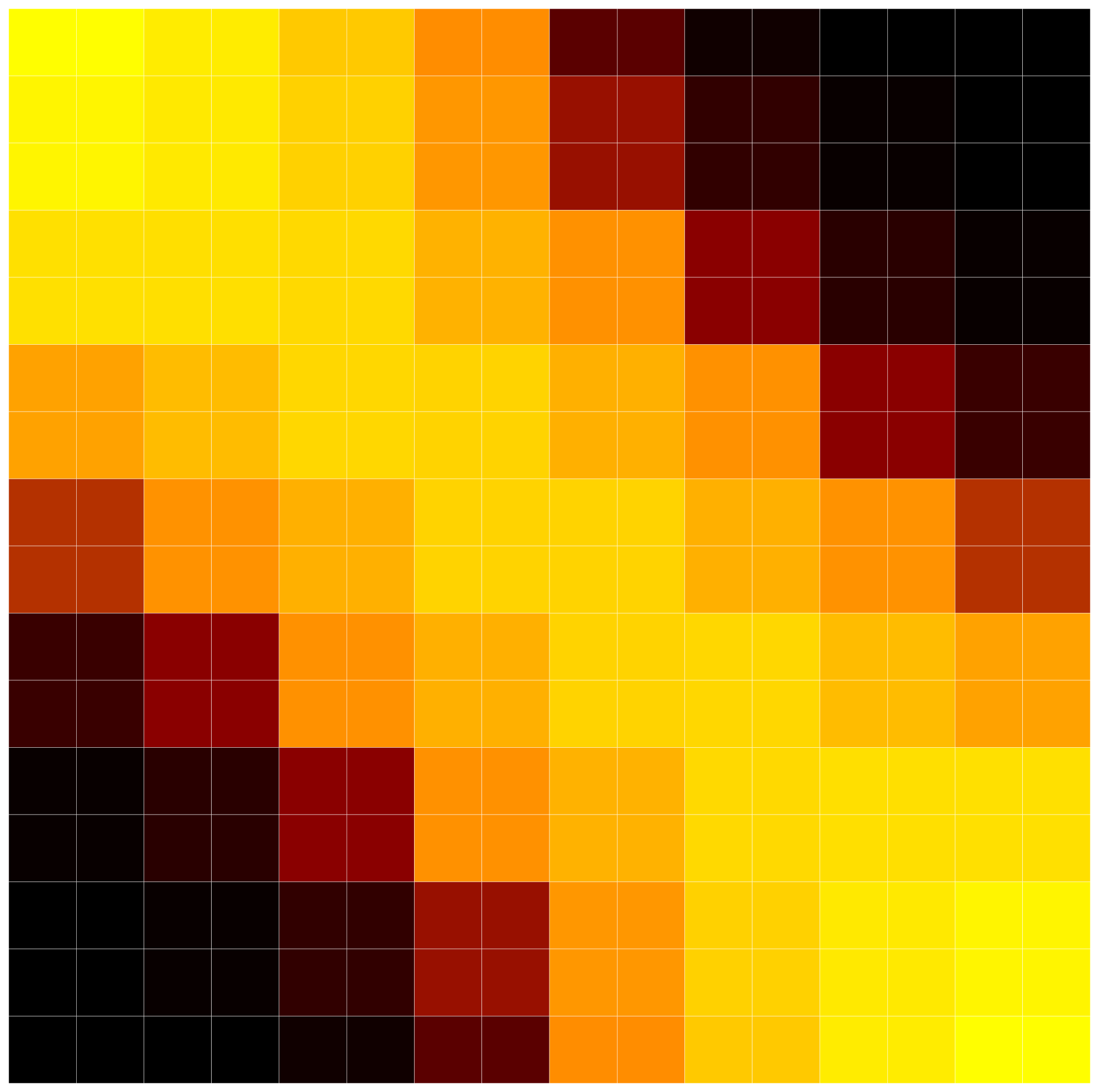}
	}
	\subfloat[$n\!=\!32$]{
  	  \includegraphics[width=0.18\columnwidth]{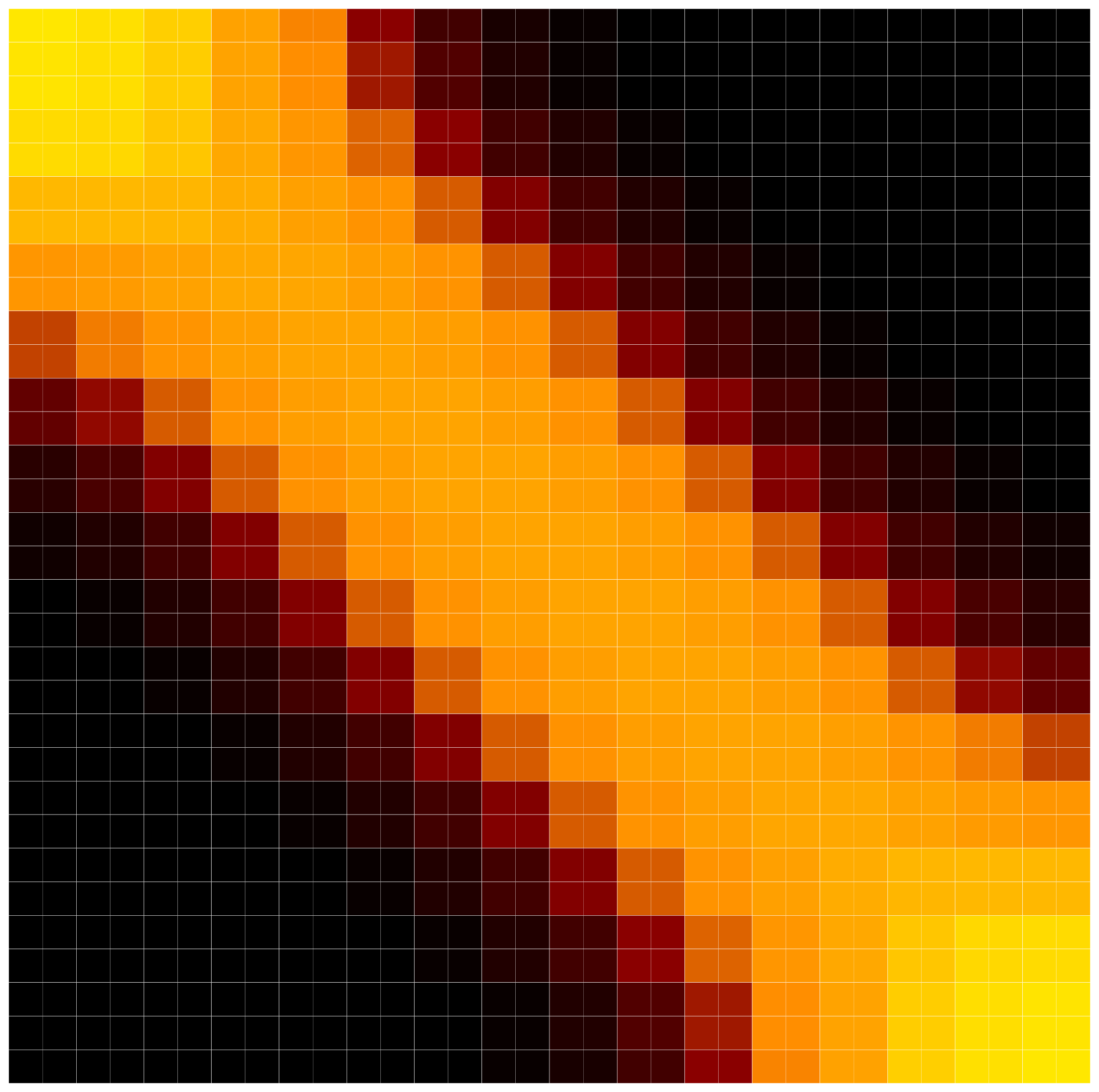}
	}
	\subfloat[$n\!=\!64$]{
  	  \includegraphics[width=0.18\columnwidth]{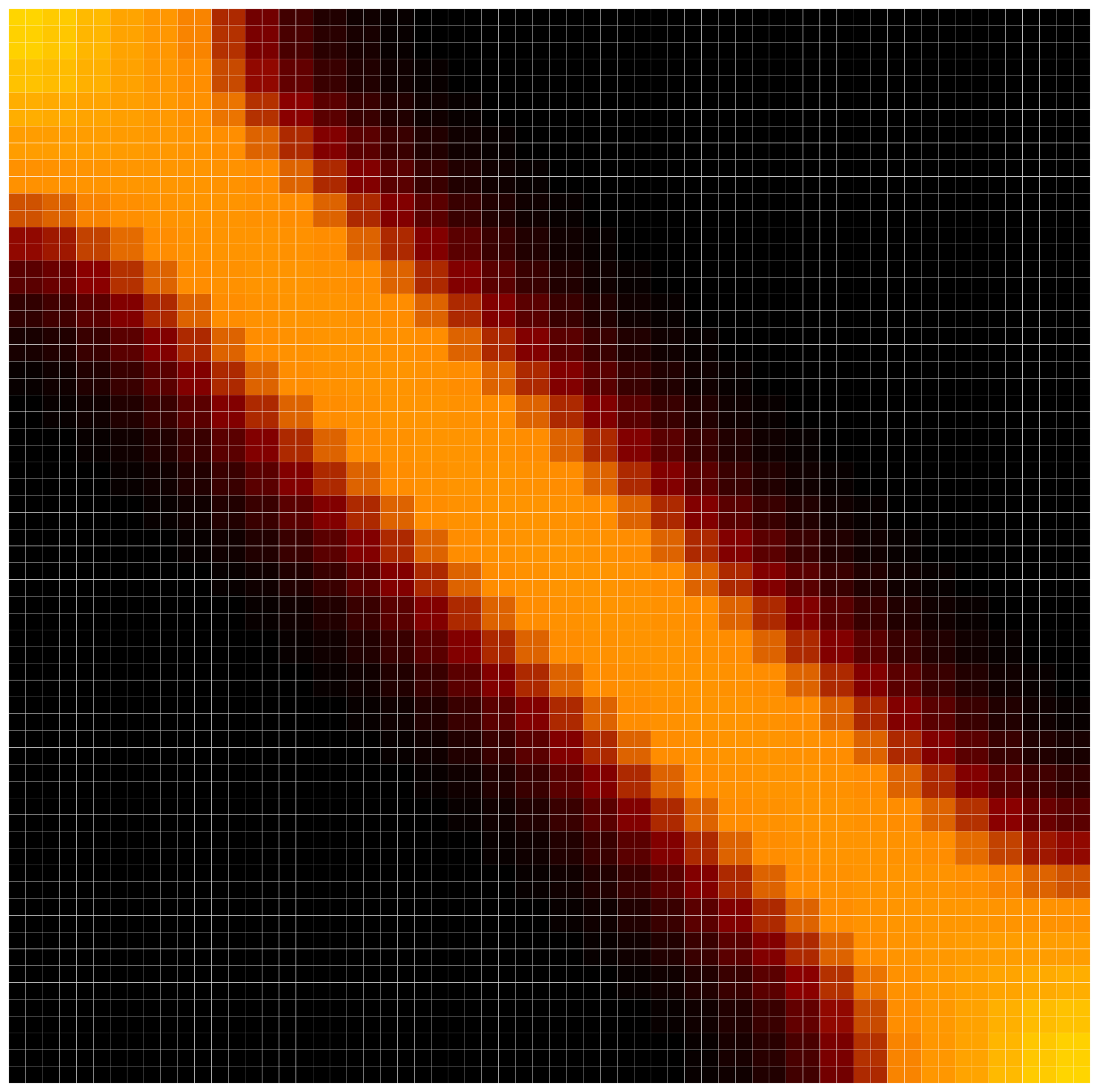}
	}
	\subfloat{
		\psfrag{a}[l][l][0.7]{0}
		\psfrag{d}[l][l][0.7]{0.7}
		\psfrag{g}[l][l][0.7]{3.0}
		\psfrag{h}[l][l][0.72]{$\times 10^{-1}$}
  	  \includegraphics[height=0.21\columnwidth]{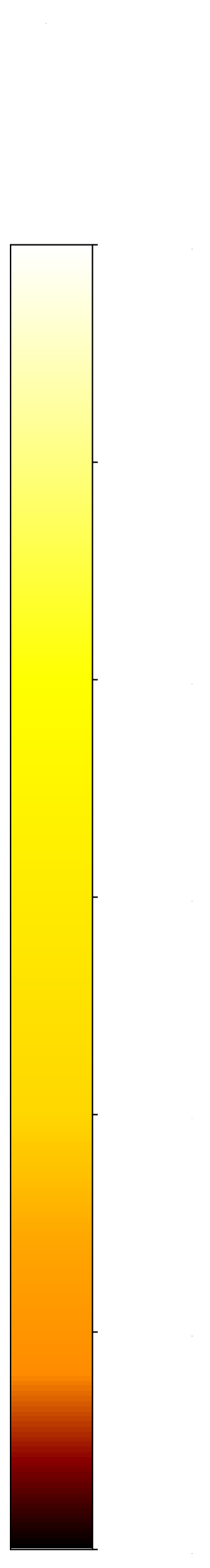}
	}
	\\[-3mm]
	\setcounter{subfigure}{4}
	\subfloat[$n\!=\!8$ Comp.]{
  	  \includegraphics[width=0.18\columnwidth]{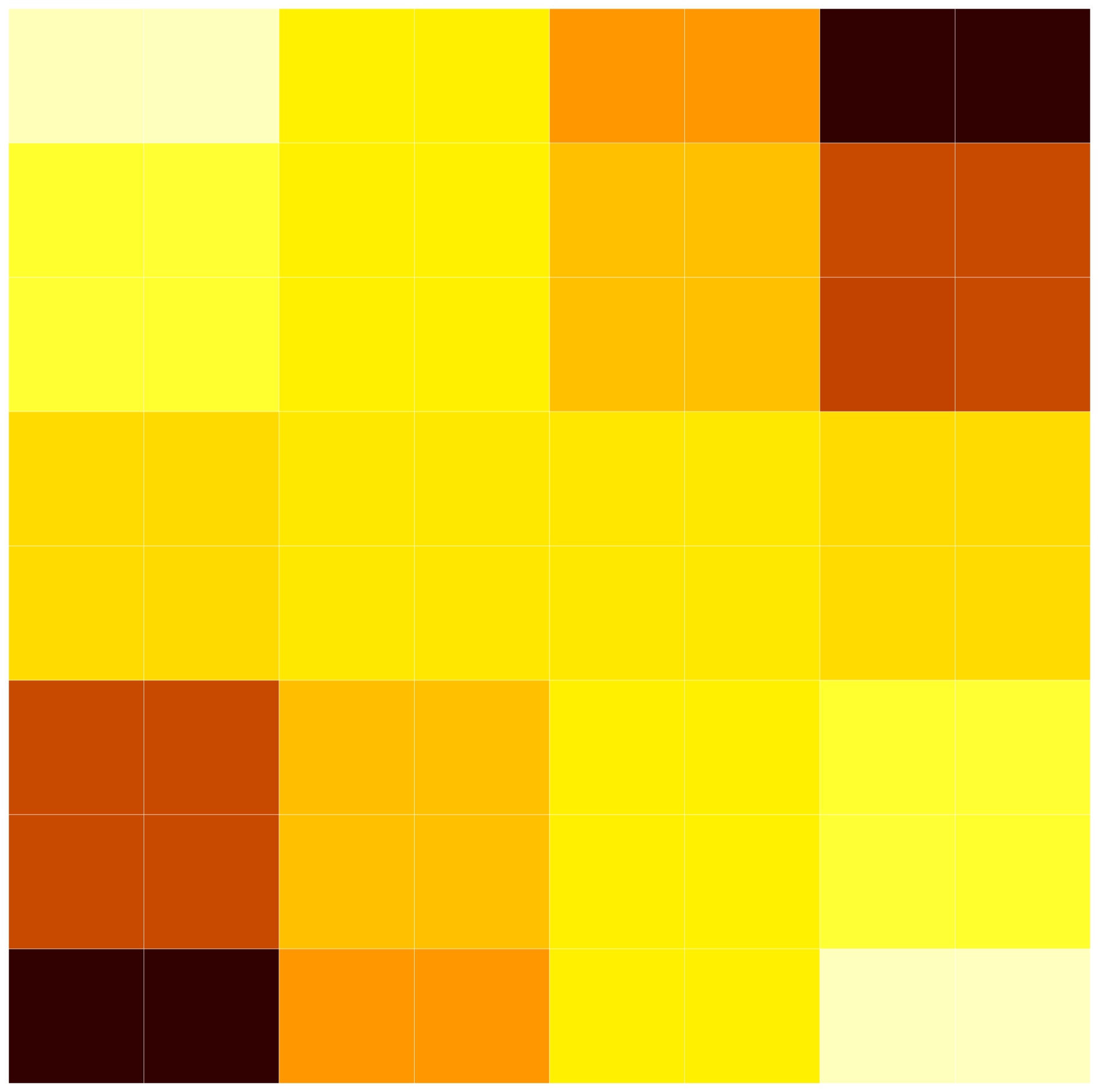}
	}
	\subfloat[$n\!=\!16$]{
  	  \includegraphics[width=0.18\columnwidth]{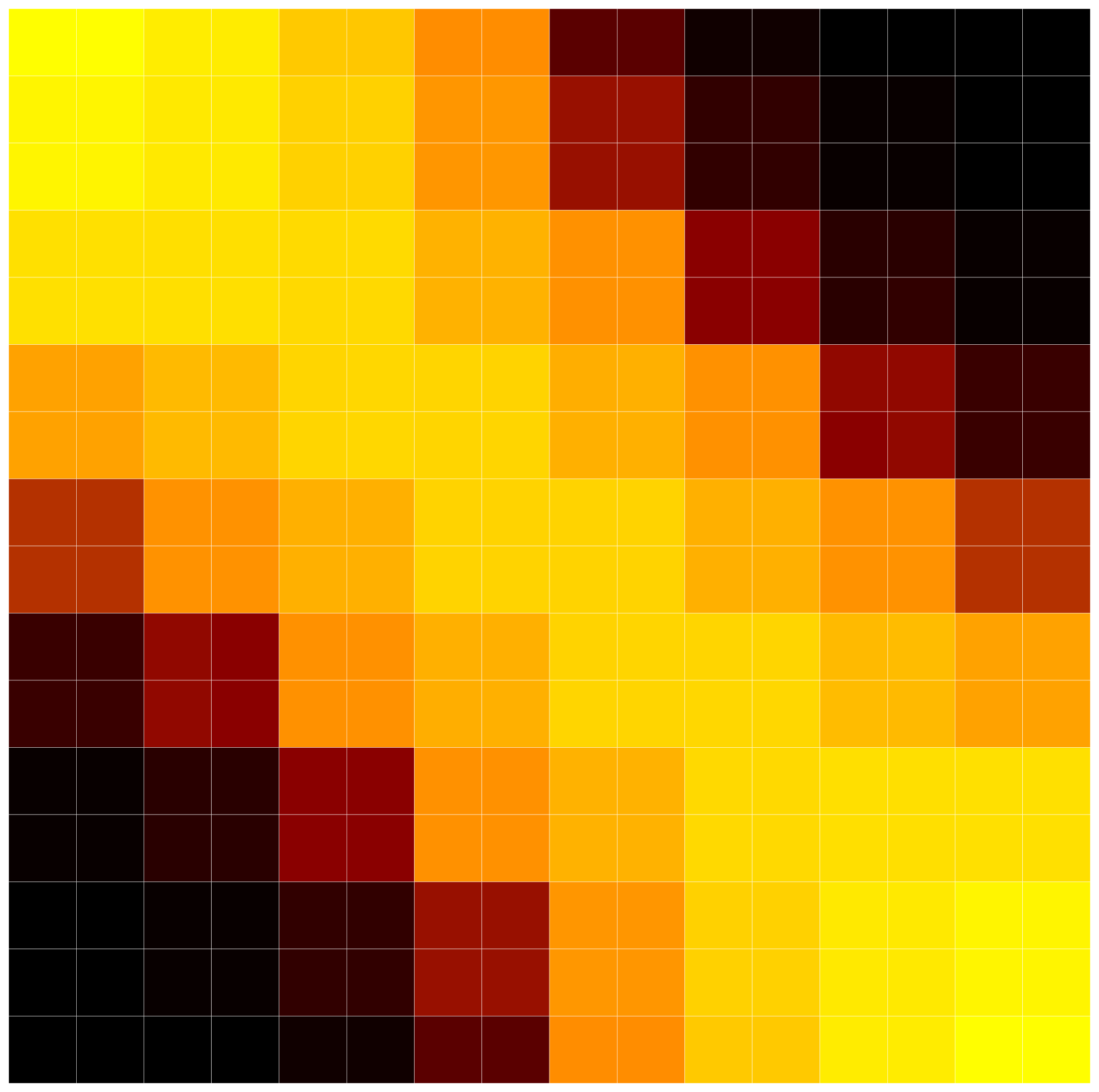}
	}
	\subfloat[$n\!=\!32$]{
  	  \includegraphics[width=0.18\columnwidth]{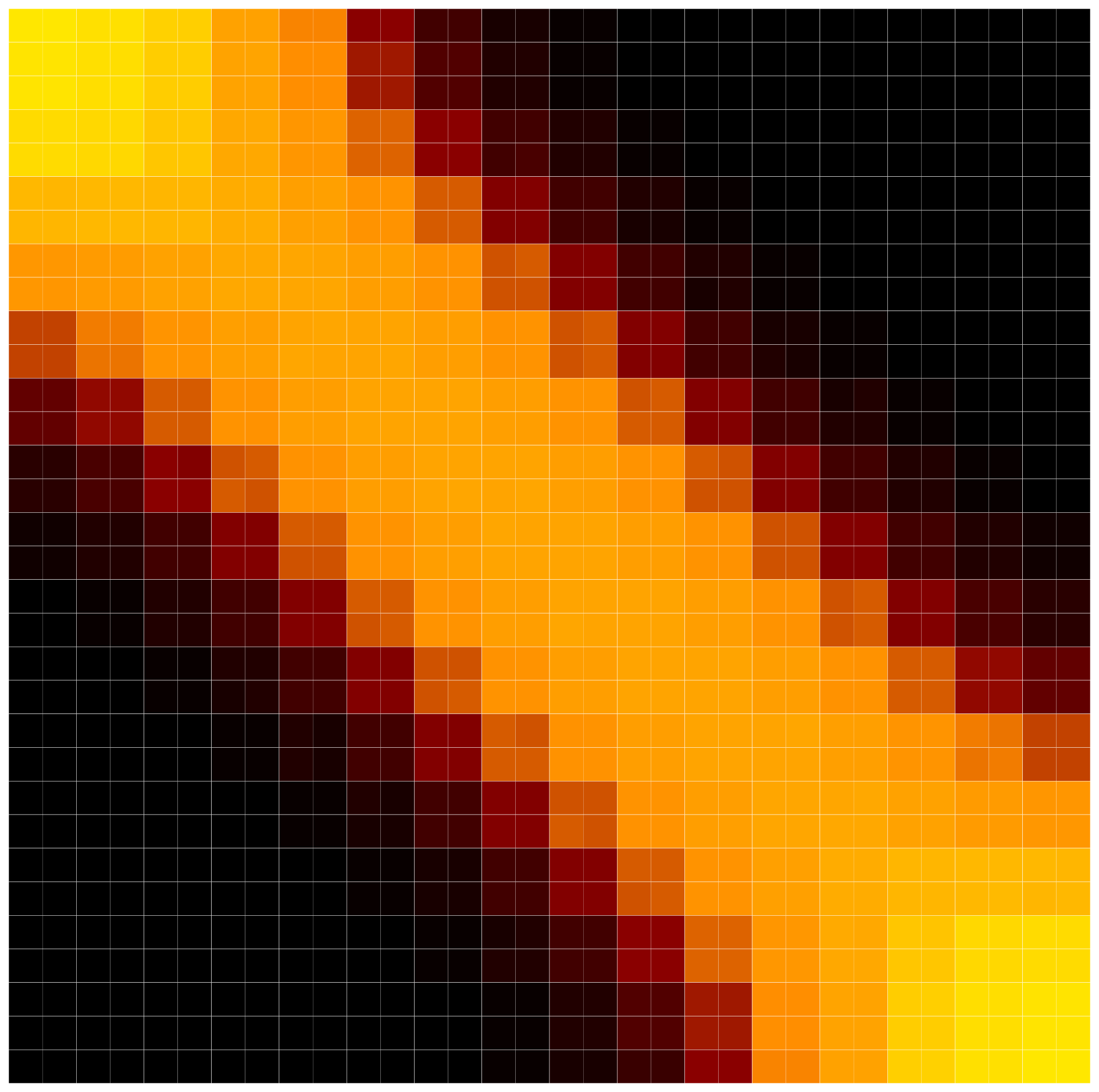}
	}
	\subfloat[$n\!=\!64$]{
  	  \includegraphics[width=0.18\columnwidth]{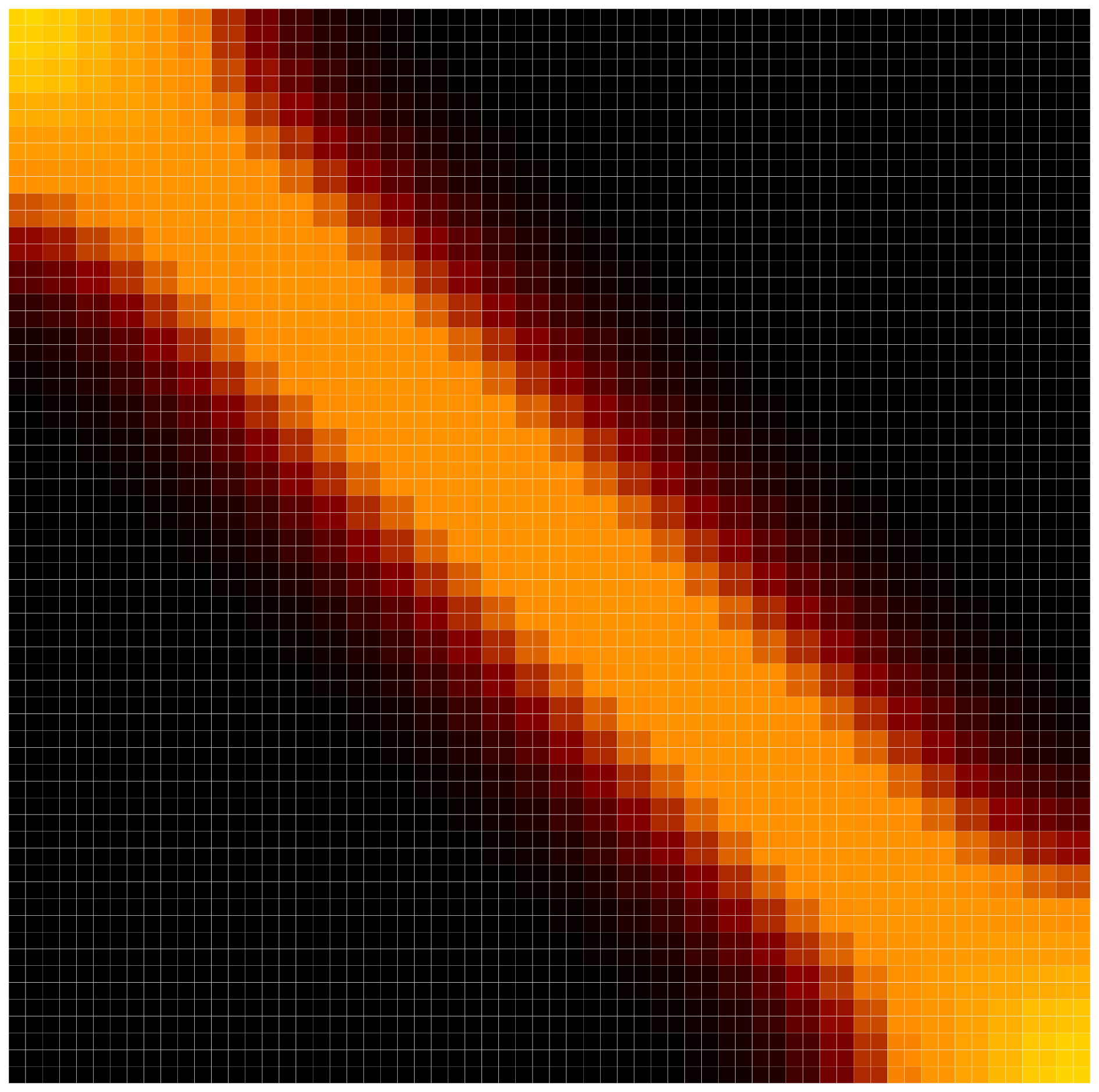}
	}
	\subfloat{
		\psfrag{a}[l][l][0.7]{0}
		\psfrag{d}[l][l][0.7]{0.7}
		\psfrag{g}[l][l][0.7]{3.0}
		\psfrag{h}[l][l][0.72]{$\times 10^{-1}$}
  	  \includegraphics[height=0.21\columnwidth]{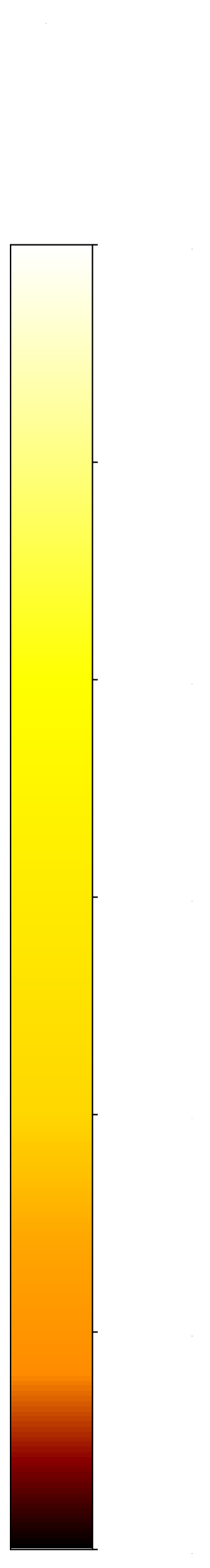}
	}

  \caption{Analytical expectations $\widehat{M}$ (top) and computational
averages $\overline{M}$ (bottom) of the squared moduli of the matrix entries
for $n\!\in\!\{8,16,32,64\}$ and stage count $m\!=\!n$.
Matrix size increases from left to right.
The horizontal and vertical axes of each heatmap represent matrix columns and
rows, respectively.
The color scale ranges from black (0) through red and yellow to white (0.30).
}
  \label{fig:comp_heatmap}
\end{figure}

\begin{table}[t]
\centering
\caption{Root-mean-square error between $\overline{M}$ and $\widehat{M}$}
\label{table:heatmap_error}
\renewcommand{\arraystretch}{1.4}
\begin{tabular}{ccccc}\hline
Matrix size $n$ &          8 & 16 & 32 & 64\\ \hline 
RMSE ($\times\! 10^{-4}$) $\downarrow$\,&\,4.84\, &\, 2.30\, &\, 1.35\, &\, 0.826\,\\ \hline
\end{tabular}
\end{table}

Figure~\ref{fig:comp_heatmap} shows the expected matrices $\widehat{M}$
[(a)--(d)] and average matrices $\overline{M}$ [(e)--(h)] for mesh
architectures with $n\!=\!8,16,32,64$ and stage count $m\!=\!n$.
For every matrix size, the two sets of heatmaps agree closely, as quantified
by the RMSE values in Table~\ref{table:heatmap_error}.
These small values demonstrate close numerical agreement between the
analytical expectations and computational averages.

The matrices in Fig.~\ref{fig:comp_heatmap} exhibit a \emph{banded structure}
that becomes more pronounced as the matrix size increases, consistent with
Ref.~\cite{pai}. Equation~(\ref{eq:expected}) provides an intuitive explanation
for the saddle-like form produced by uniform random initialization. The form
has two principal features: entries with large expectations are concentrated
near the diagonal, and the largest expectations occur near the upper-left and
lower-right corners.
The former feature originates from the number of paths $|P_{(j,h)}|$.
Equation~(\ref{eq:expected}) shows that, when $F(p)$ is the same for all
relevant paths, the expectation increases with path multiplicity. Ordinarily,
a path traverses one MZI per stage; the exception is a path that follows a top
or bottom boundary in an odd-numbered stage (e.g., $s\!=\!1,3$ in
Fig.~\ref{fig:meshpath}). As Fig.~\ref{fig:pathcount} shows, paths originating
from the $h$th input are concentrated near the $h$th output.
The latter feature arises from the number of MZIs along each path, $F(p)$.
Equation~(\ref{eq:expected}) indicates that 
if the MZI count $F(p)$ decreases by 1, the term $1/2^{F(p)}$ doubles.
If a path includes a top or bottom boundary at an odd-numbered stage,
the path has fewer MZIs than the stage count $m$.
The upper-left and lower-right entries have many such paths and therefore
larger expectations, producing the saddle-like banded structure.

For a fully parameterized mesh architecture with $n^2$ parameters, the Haar
initialization method \cite{pai} prevents this banded structure from forming
in $\overline{M}$ and yields a statistically uniform ensemble of implemented
unitary matrices.
However, Haar initialization does not address the parameter-constrained
architectures considered here.

\subsection{Lack of Reachability}
Truncated mesh architectures fail to satisfy the reachability property,
as they employ fewer than the $(n\!-\!1)$ stages
required to ensure full connectivity,
as detailed in Appendix~\ref{app_sec:reach}.
An example illustrating this limitation is shown in Fig.~\ref{fig:mesh}(c), 
where the top input port has no path to the bottom three output ports; 
consequently, signals from this input cannot reach those outputs.
This structural constraint restricts the set of realizable transformations 
and limits the mixing of signals across modes. 
Consequently, the structure yields \emph{trivial zero} entries 
in an implemented unitary matrix.
They cannot be eliminated by parameter initialization schemes,
including those designed to improve statistical properties.
This highlights that architectural design, not initialization, is the primary factor 
governing reachability and mixing capability in parameter-constrained settings.

\begin{figure}[t]
\begin{center}
	\psfrag{X}[c][c][1.0]{Normalized stage count: $m/n$}
	\psfrag{Y}[c][c][.95]{
		\begin{picture}(0,0)
			\put(-39,31){\makebox{Normalized trivial-}}
			\put(-32,22){\makebox{zero-entry count}}
			\put(-12,8){\makebox{$\zeta_{0}/n^{2}$}}
		\end{picture}
	}
	\psfrag{A}[c][c][0.9]{0} \psfrag{F}[c][c][0.9]{1} 
	\psfrag{B}[c][c][0.9]{0.2} \psfrag{C}[c][c][0.9]{0.4}
	\psfrag{D}[c][c][0.9]{0.6} \psfrag{E}[c][c][0.9]{0.8}
	\psfrag{O}[r][r][0.9]{0} \psfrag{T}[r][r][0.9]{1} 
	\psfrag{P}[r][r][0.9]{0.2} \psfrag{Q}[r][r][0.9]{0.4}
	\psfrag{R}[r][r][0.9]{0.6} \psfrag{S}[r][r][0.9]{0.8}
	\psfrag{U}[r][r][0.9]{$n=64$}
	\psfrag{V}[r][r][0.9]{$32$}
	\psfrag{W}[r][r][0.9]{$16$}
	\psfrag{Z}[r][r][0.9]{Parabola}
	\includegraphics[width=50mm]{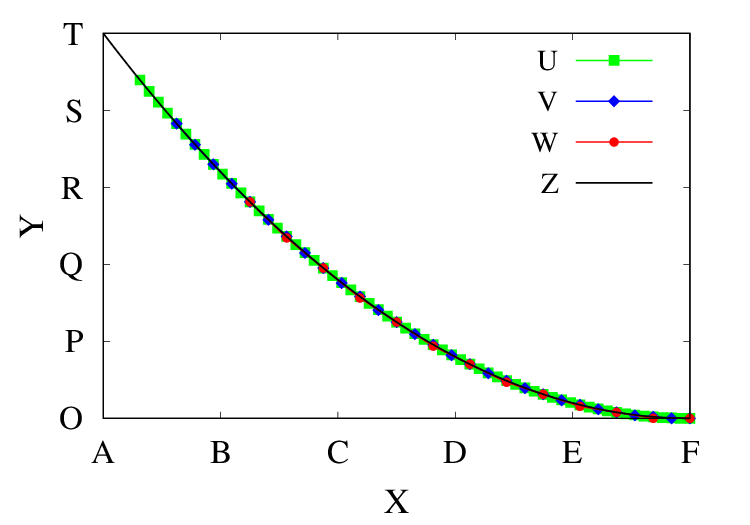}
\end{center}
\vspace*{-2mm}
\caption{
Number of trivial zero entries normalized by $n^2$ 
against stage count normalized by $n$, where $n\!=\!\{16,32,64\}$.
The solid line shows the quadratic function $(1\!-\!m/n)^{2}$.
}
\label{fig:nz}
\end{figure}

For a truncated mesh architecture with $n$ inputs and $m$ stages,
where $m\!\in\!\mathbb{N}$ and $m\!<\!n$, 
the matrix entry $g_{jh}$ is nonzero only if the following condition holds:
\begin{align}
\max( 2\hat{h}\!+\!1 -\!m,\, 0 ) &\leq j \leq 
\min( 2\hat{h}\!+\!m,\, n\!-\!1 ) \label{eq:trunc_nonzero}\\
\hat{h} &= \lfloor h/2 \rfloor\, . \nonumber
\end{align}
Based on Eq.~(\ref{eq:trunc_nonzero}), 
the numbers of nonzero entries $\zeta_{\lnot 0}$ and zero entries $\zeta_{0}$
are obtained as
\begin{equation}
\zeta_{\lnot 0} = \sum_{h=0}^{n-1}\left\{
\min( 2\hat{h}\!+\!m,n\!-\!1 )
-\!\max( 2\hat{h}\!+\!1 -\!m,0 )+\!1\right\} \label{eq:zeta_nonzero}
\end{equation}
\begin{equation}
\zeta_0 = n^{2} -\zeta_{\lnot 0}\, . \label{eq:zeta0}
\end{equation}
Figure~\ref{fig:nz} shows the normalized number of trivial zero entries,
$\zeta_{0}/n^{2}$, as a function of the normalized stage count $m/n$ for
$n\!=\!\{16,32,64\}$, as calculated using Eq.~(\ref{eq:zeta0}).
This result shows that reducing the number of stages leads to 
a quadratic increase in the number of trivial zero entries,
thereby significantly limiting connectivity 
and degrading mixing capability across modes. 

\begin{figure}[t]
  \centering
	\captionsetup[subfloat]{format=plain,margin={3pt,0pt}}
		\subfloat[$m\!=\!16$, $\zeta_{0}\!=\!0$]{
  	  \includegraphics[width=0.18\columnwidth]{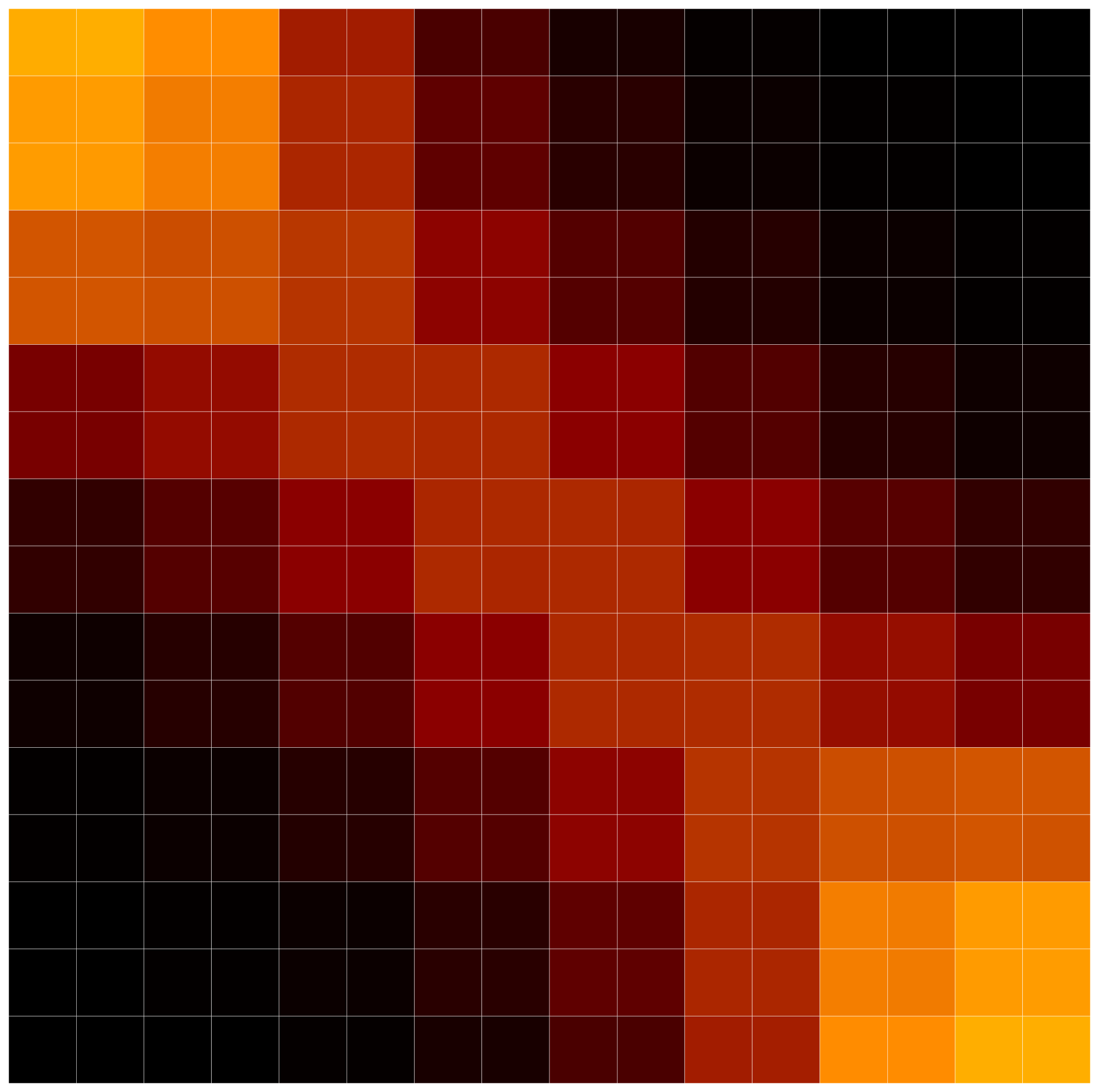}
	}
		\subfloat[$m\!=\!12$, $\zeta_{0}\!=\!16$]{
  	  \includegraphics[width=0.18\columnwidth]{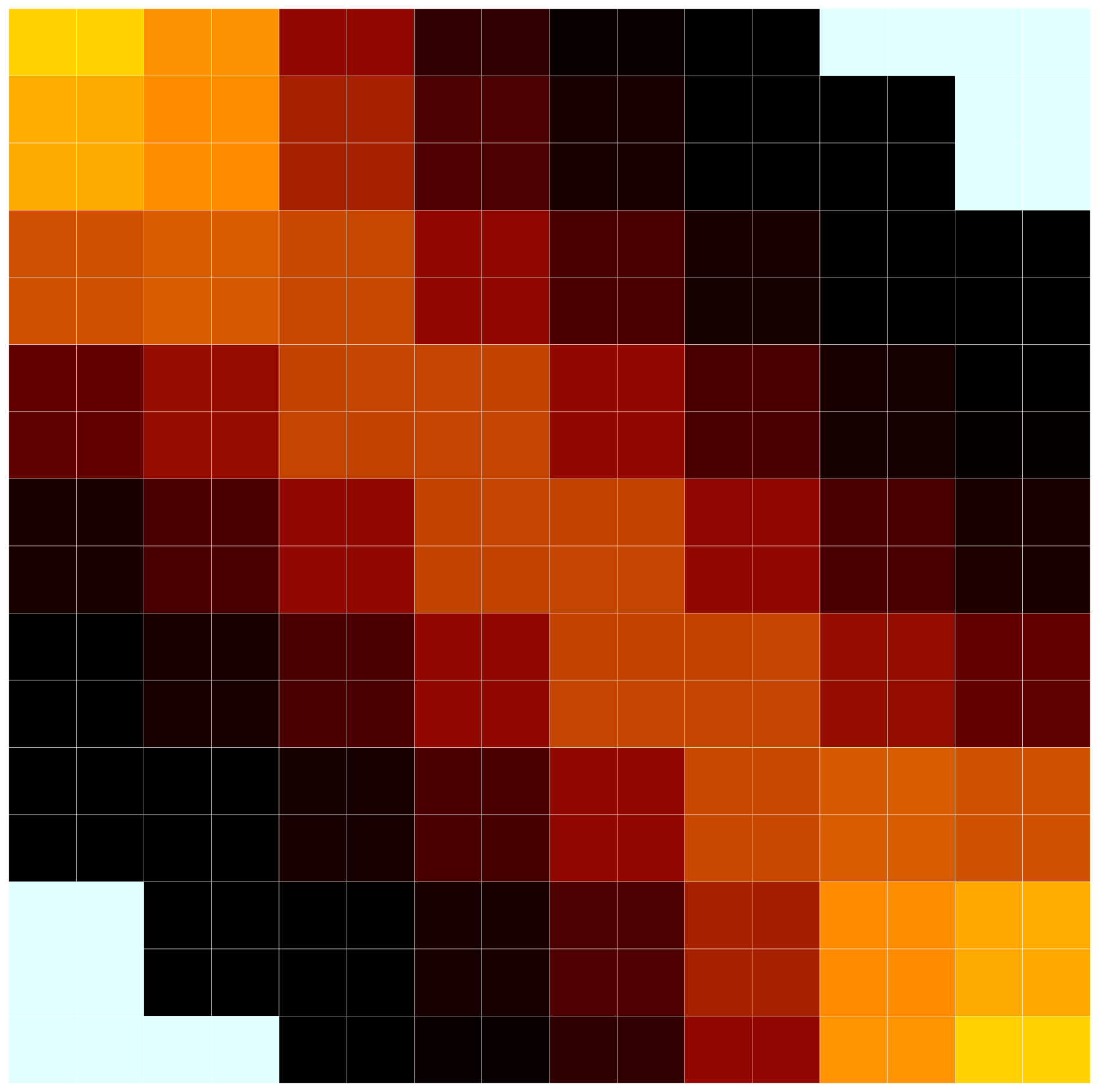}
	}
		\subfloat[$m\!=\!8$, $\zeta_{0}\!=\!64$]{
  	  \includegraphics[width=0.18\columnwidth]{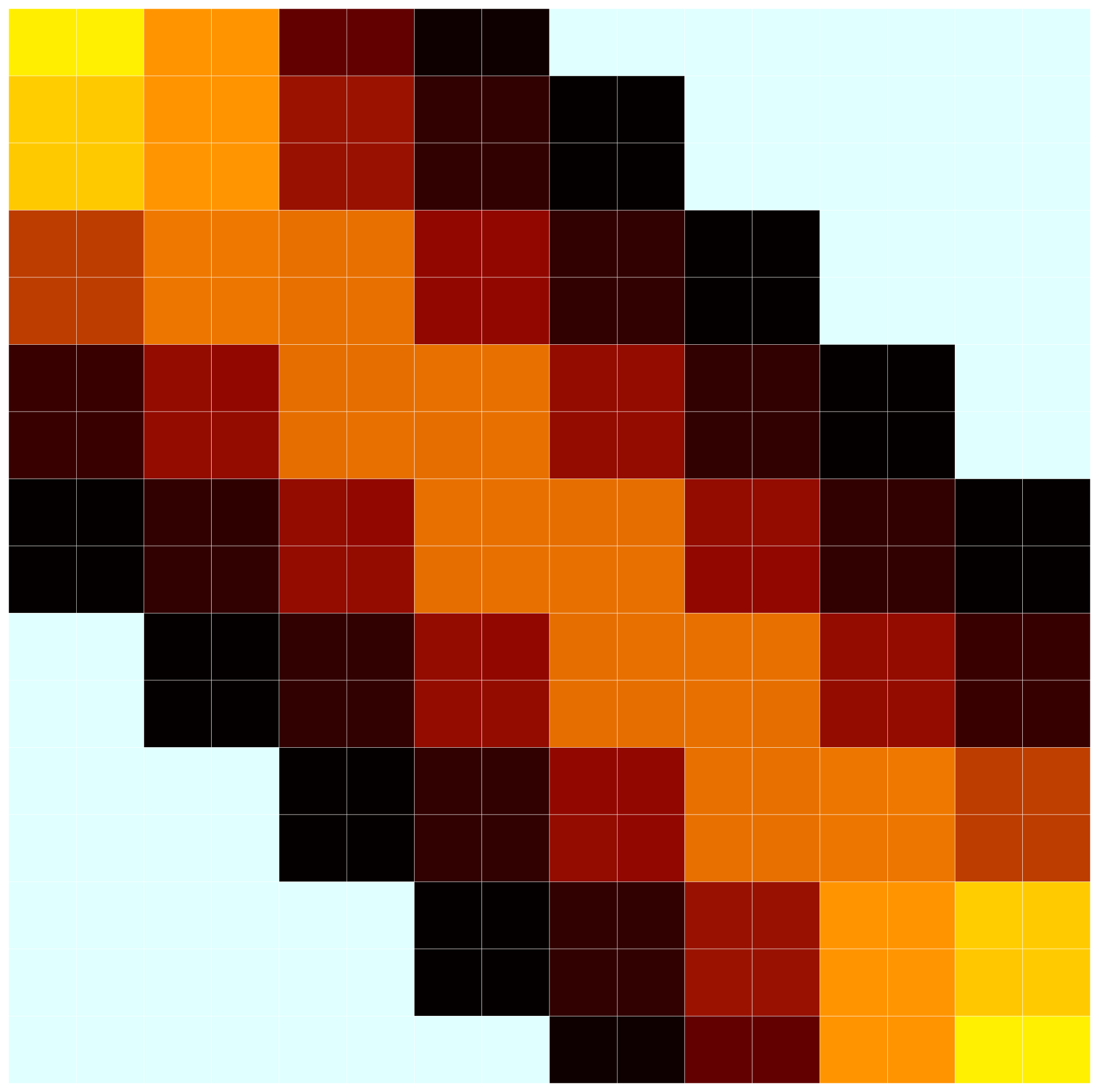}
	}
		\subfloat[$m\!=\!4$, $\zeta_{0}\!=\!144$]{
  	  \includegraphics[width=0.18\columnwidth]{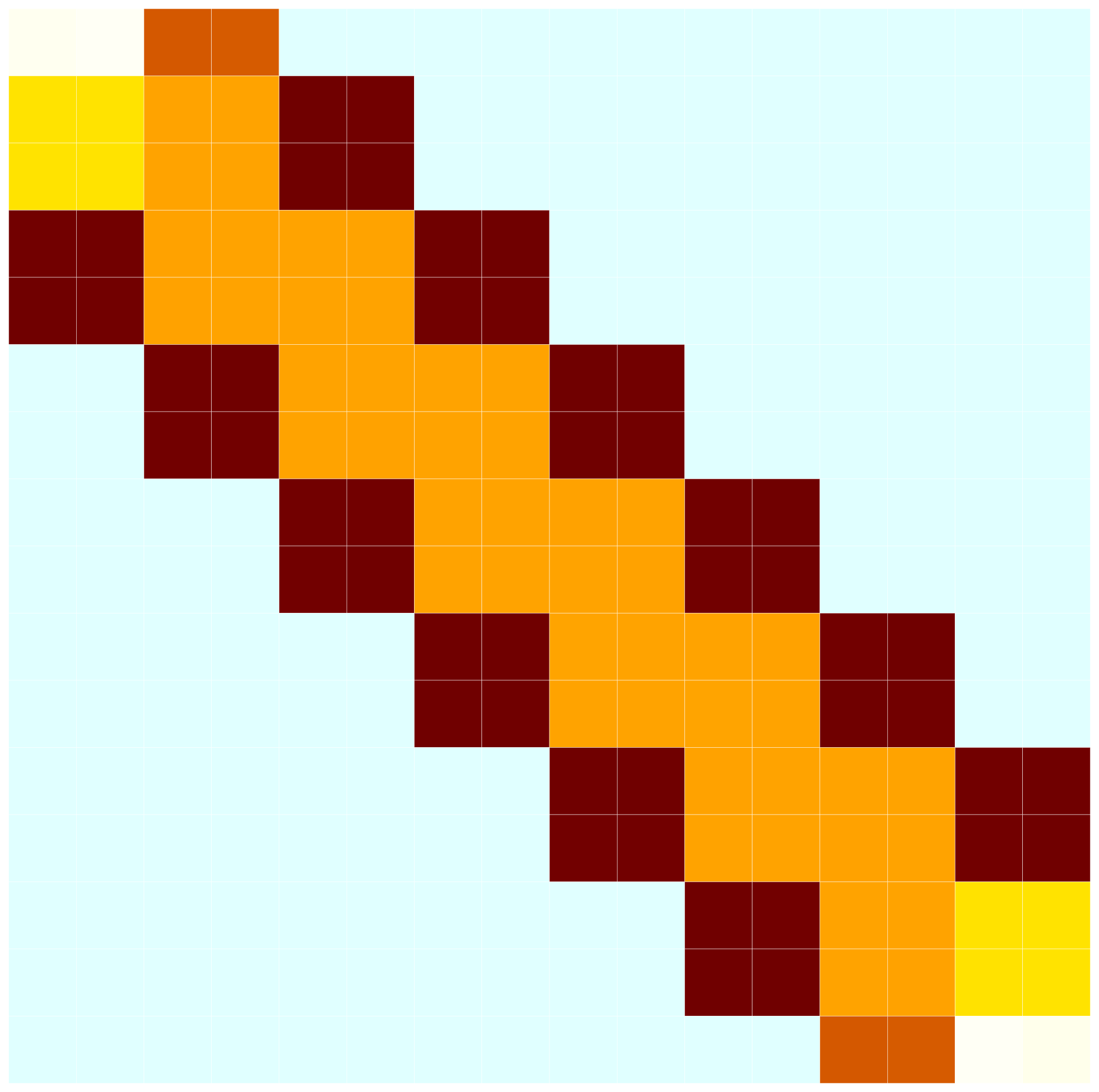}
	}
	\subfloat{
		\psfrag{a}[l][l][0.7]{0}
		\psfrag{d}[l][l][0.7]{1.9}
		\psfrag{g}[l][l][0.7]{3.8}
		\psfrag{h}[l][l][0.72]{$\times 10^{-1}$}
  	  \includegraphics[height=0.21\columnwidth]{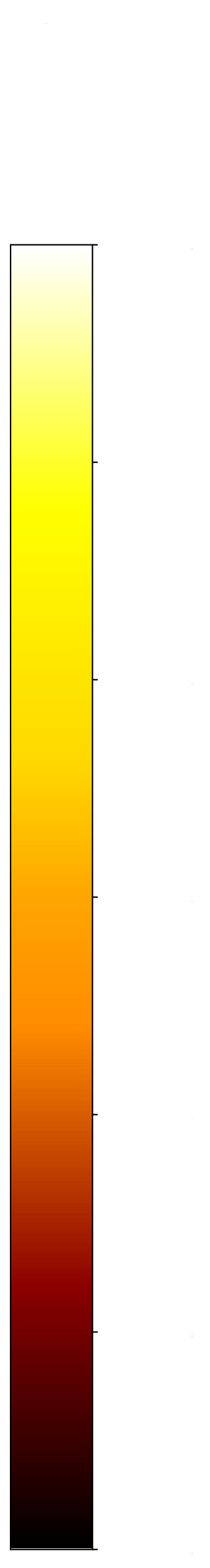}
	}

  \caption{Average squared moduli of matrix entries 
for the mesh architecture and its truncated variants with $n\!=\!16$ and 
stage counts $m\!=\!\{16,12,8,4\}$, computed over $10^{6}$ instances
whose parameters are initialized uniformly at random.
The horizontal and vertical axes correspond to input (column) and output (row) indices, 
respectively.
The color scale ranges from black (low: 0) through red and yellow to white (high: 0.38). 
Trivial zero entries corresponding to unreachable input--output pairs, 
which are distinguished from extremely small values near zero,
are highlighted in cyan; their total number is denoted by $\zeta_{0}$.
}
  \label{fig:comp_heatmap_trunc}
\end{figure}

To illustrate these properties,
Fig.~\ref{fig:comp_heatmap_trunc} shows 
the average squared moduli of matrix entries for the mesh and its truncated variants 
with $n\!=\!16$, stage counts $m\!=\!\{16,12,8,4\}$, 
computed over $10^{6}$ randomly initialized instances.
As the number of stages decreases, 
trivial zero entries corresponding to unreachable input-output pairs
increase and expand from the corners toward the central region.
This behavior reflects the loss of reachability 
and the resulting degradation of 
both statistical uniformity in the transformation and mixing capability.

\subsection{Inefficiency of MZIs}
We address the inefficiency of the MZI as the fundamental building block.
Since our approach directly optimizes phase parameters 
without relying on the two-step training method 
described in Section~\ref{ssec:train}, 
it is not restricted to a specific architectural template and 
can, in principle, be applied to arbitrary architectures.
In conventional mesh architectures,
each MZI functions as a complete two-mode interferometer
with two tunable parameters, as illustrated in Fig.~\ref{fig:mesh}(a).
Although this structure enables universal unitary transformations,
it is inefficient when the objective is strong mixing
under a limited parameter count.
In particular, using two parameters per MZI reduces the number of coupling
operations available under a fixed parameter budget and therefore limits the
mixing capability per parameter. Consequently, global mixing across all modes
requires a comparatively large number of parameters.
By contrast, our approach does not rely on MZIs as fundamental units. 
A fine-grained structure enables more efficient utilization of parameters 
and improved mixing capability per parameter.

These limitations highlight the need for architectural designs
that ensure both statistical uniformity and strong mixing capability per
parameter under parameter constraints,
motivating the proposed fine-grained butterfly (FGB) architecture.

\section{Proposed Fine-Grained Butterfly Architecture}\label{sec:prop}
We first introduce the butterfly topology,
which serves as the structural basis of the proposed architecture.
Widely used in communication networks,
this topology provides efficient and structured connectivity 
for information propagation.
Building on this,
we present the fine-grained butterfly (FGB) architecture, 
which addresses the three limitations of conventional rectangular mesh
designs identified in the preceding section.

\subsection{Butterfly Topology}\label{ssec:topology}
\begin{figure}[t]
\begin{center}
	\psfrag{0}[c][c][.8]{0} \psfrag{1}[c][c][.8]{1} \psfrag{2}[c][c][.8]{2}
	\psfrag{3}[c][c][.8]{3} \psfrag{4}[c][c][.8]{4} \psfrag{5}[c][c][.8]{5}
	\psfrag{6}[c][c][.8]{6} \psfrag{7}[c][c][.8]{7} \psfrag{8}[c][c][.8]{8}
	\psfrag{9}[c][c][.8]{9} \psfrag{10}[c][c][.8]{10} \psfrag{11}[c][c][.8]{11}
	\psfrag{12}[c][c][.8]{12} \psfrag{13}[c][c][.8]{13} \psfrag{14}[c][c][.8]{14}
	\psfrag{15}[c][c][.8]{15} 
	\psfrag{P}[r][r][.9]{$n\!=\!16$} \psfrag{Q}[c][c][.9]{Diameter: $\log_2 n+1 =\!5$}
	\psfrag{R}[c][c][.88]{\# Stages: $\log_2 n=\!4$} 
	\psfrag{K}[c][c][.77]{Stage~0} \psfrag{L}[c][c][.77]{Stage~1}
	\psfrag{M}[c][c][.77]{Stage~2} \psfrag{N}[c][c][.77]{Stage~3}
	\includegraphics[width=58mm]{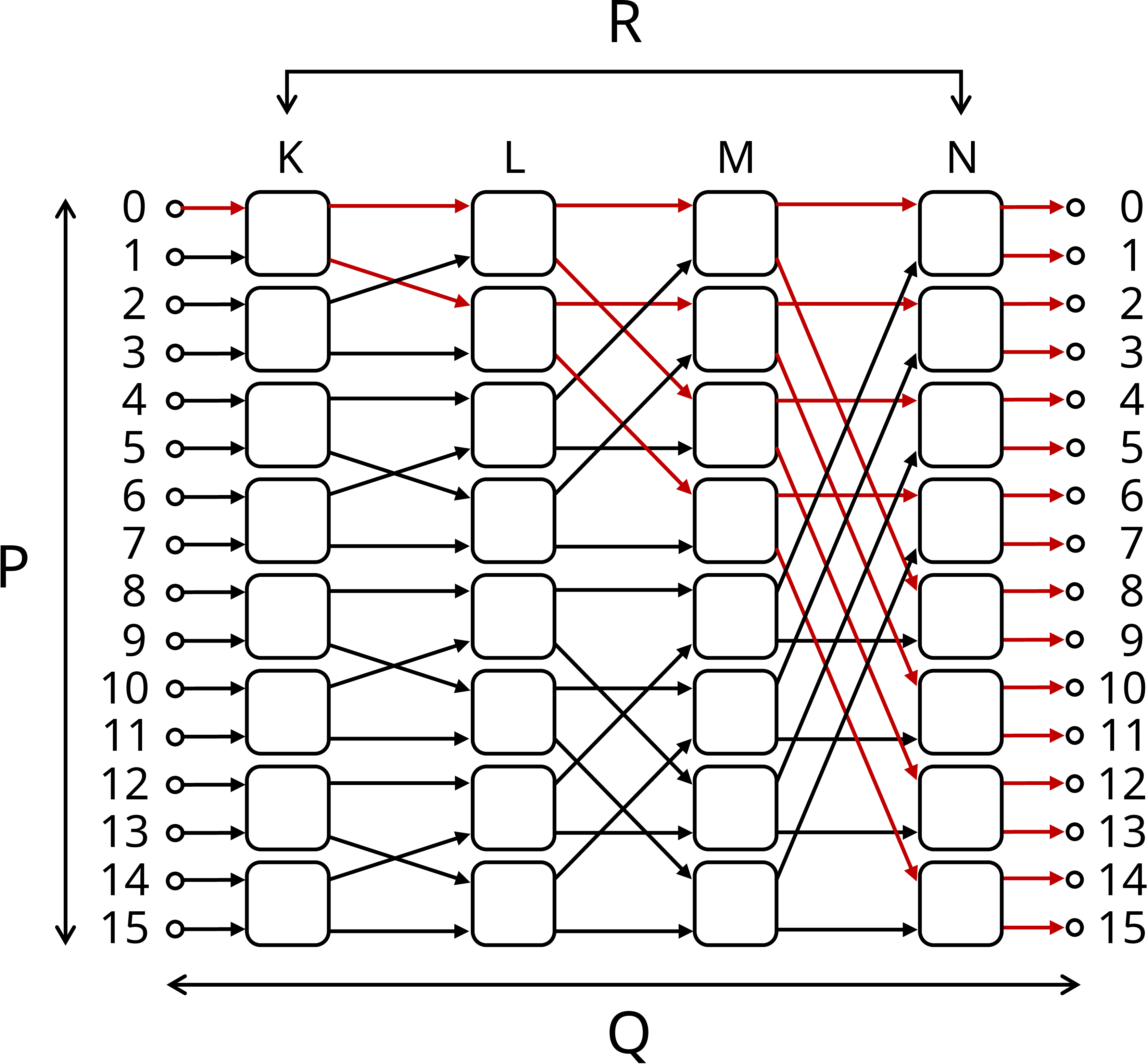}
\end{center}
\vspace*{-2mm}
\caption{Standard butterfly topology with 16 inputs, 16 outputs, and four stages.
}\label{fig:std_bflyarch}
\end{figure}
A butterfly topology has been widely adopted in various domains,
including interconnection networks in communication systems \cite{thomson,dally,wu,kruskal}, 
fast Fourier transforms in signal processing \cite{oppenheim}, 
and neural network models \cite{vahid,lin,ding}.
In this work, we leverage its structural properties 
to design the proposed architecture.
Figure~\ref{fig:std_bflyarch} illustrates a standard butterfly topology\footnote{
This topology is referred to as a \emph{banyan network} in 
interconnection networks \cite{wu} and has the same connectivity as a
decimation-in-time fast Fourier transform (DIT-FFT) in signal processing
\cite{oppenheim}.
}
with 16 input and 16 output terminals 
(see Appendix~\ref{app_sec:bfly} for the general case).
The network consists of four stages, each containing eight functional nodes.
Each node, represented as a rounded square, has two input and two output terminals, 
with outgoing edges connected to predetermined nodes in the subsequent stage.
This deterministic interstage connectivity yields a recursive
and highly regular structure.
In this example, the network diameter, defined below, is five.
\begin{dfn}[Network diameter] \label{dfn:diameter}
The network diameter is the maximum number of edges 
traversed along a path from any input terminal to any output terminal.
\end{dfn}

The butterfly topology exhibits the following key properties \cite{dally}:
\begin{enumerate}[(a)]
\item \emph{Unique path.}
For each input terminal, there exists exactly one path to every output terminal.
This eliminates path redundancy and ensures uniform path structure 
across all input-output pairs.
As illustrated in Fig.~\ref{fig:std_bflyarch},
the connections originating from a given input terminal (e.g., the top terminal) form
a perfect binary tree, where a unique path exists from the root to each leaf.
The butterfly topology can be regarded as a superposition of such trees.
\item \emph{Minimum diameter.}
The butterfly topology achieves a minimum network diameter of $(\log_2 n +\!1)$ 
for $n$ input and $n$ output terminals, 
where $n\!=\!2^{t}$, $t\!\in\!\mathbb{N}$,
while preserving the reachability property in Definition~\ref{dfn:reachability}.
This logarithmic scaling enables efficient propagation across the network.
\item \emph{Long edges.}
To maintain its logarithmic depth, the butterfly topology incorporates 
long-range connections.
The maximum \emph{node stride}, defined below,  
is $n/4$.
\end{enumerate}

\begin{dfn}[Node stride] \label{dfn:stride}
Each functional node in a stage is assigned an index from 0 to $(n/2 \!-\!1)$.
The node stride is defined as the absolute difference between the indices 
of a node and its connected node in the subsequent stage.
In Fig.~\ref{fig:std_bflyarch},
the maximum node stride is four, occurring between nodes in Stages 2 and 3.
\end{dfn}

These properties collectively ensure uniform path structure and 
efficient global connectivity, 
which are essential for achieving statistical uniformity and 
strong mixing capability per parameter.

\subsection{PSBS-Based Butterfly Implementation}\label{ssec:bfly}
\begin{figure}[t]
\begin{center}
	\psfrag{A}[c][c][1.0]{(a)}
	\psfrag{B}[c][c][1.0]{(b)}
	\psfrag{X}[r][r][0.9]{$n\!=\!8$}
	\psfrag{Y}[c][c][0.9]{$m\!=\!\log_{2}n\!=\!3$}
	\psfrag{P}[c][c][0.9]{PS}
	\psfrag{R}[c][c][0.9]{BS}
	\includegraphics[width=67mm]{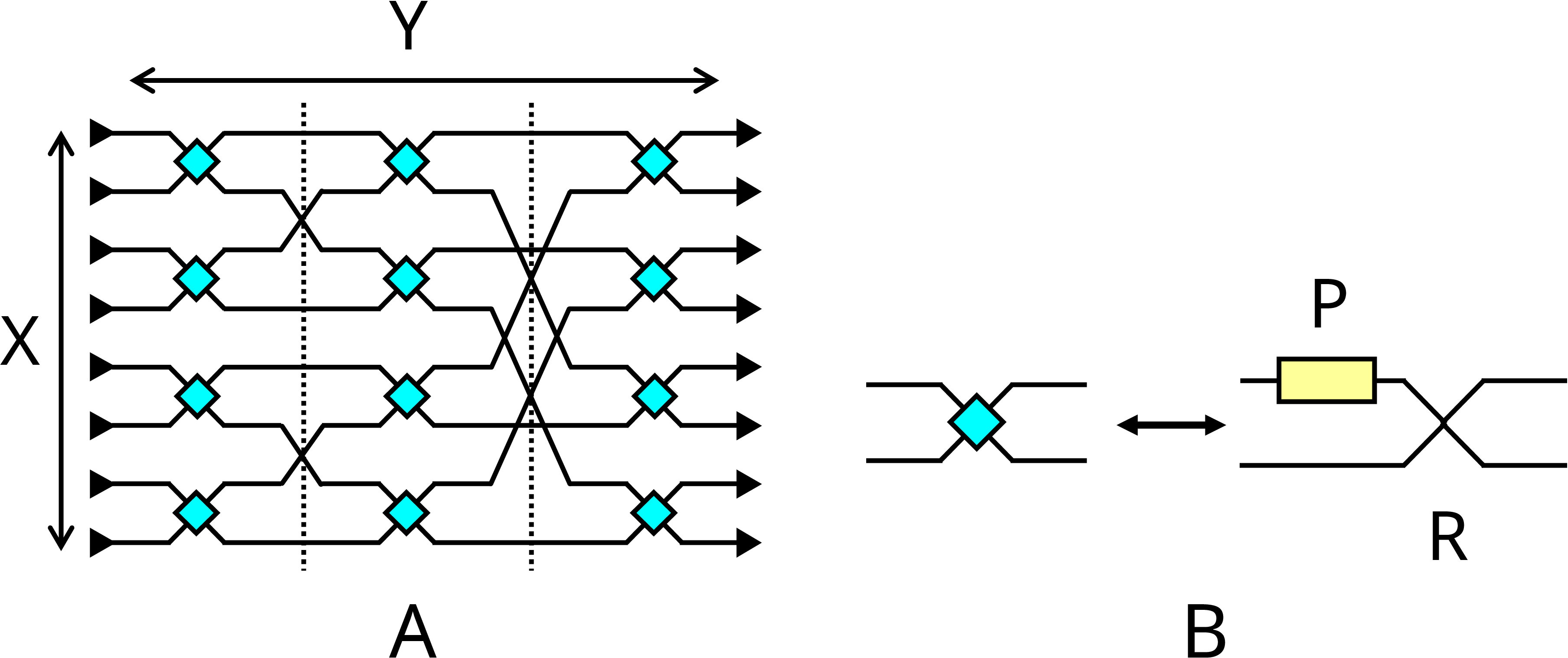}
\end{center}
\vspace*{-2mm}
\caption{FGB architecture:
(a) butterfly structure with eight inputs ($n\!=\!8$) and three stages
($m\!=\!\log_{2}n\!=\!3$), and (b) PSBS symbol (solid diamond) and its
physical circuit, composed of a programmable phase shifter (PS) and a beam
splitter (BS). Lines represent waveguides.
}\label{fig:8in_bfly}
\end{figure}

The FGB architecture adopts the butterfly topology, 
in which each functional node is realized by a minimal building block
consisting of a programmable phase shifter (PS) and a beam splitter (BS)
with a fixed 50:50 power-splitting ratio.
An $n$-input butterfly structure with $\log_2 n$ stages 
is referred to as a butterfly block, 
where $n\!=\!2^t$, $t\!\in\!\mathbb{N}$.
Figure~\ref{fig:8in_bfly}(a) shows a butterfly-block example 
with $n\!=\!8$.
The building block in Fig.~\ref{fig:8in_bfly}(b), referred to as a PSBS unit, 
implements a parameterized unitary transformation with transfer matrix  
\begin{equation}
\setlength{\arraycolsep}{2.2pt}
T_{(\text{PSBS})}(\phi)\! =
\underbrace{\frac{1}{\sqrt{2}}\begin{pmatrix}1&i\\i&1\end{pmatrix}}_{\text{BS}}
\,\underbrace{\begin{pmatrix}e^{i\phi}&0\\0&1\end{pmatrix}}_{\text{PS}}
\!=\! \frac{1}{\sqrt{2}}
\begin{pmatrix}e^{i\phi}& i\\ ie^{i\phi}&1 \end{pmatrix}\, ,
\label{eq:psbs}
\end{equation}
where $\phi$ is the sole tunable parameter.\footnote{
As with MZIs, alternative configurations are possible, such as placing the
phase shifter on the lower arm or reversing the order of the phase shifter and
beam splitter.
}
In contrast to the MZI in Eq.~(\ref{eq:mzi}), 
the PSBS unit employs only a single parameter.
This fine-grained parameterization addresses the third limitation by improving
the use of parameters relative to MZI-based designs. As a trade-off, a PSBS
unit alone neither performs switching nor constitutes a complete interferometer.

\begin{figure}[t]
\begin{center}
	\psfrag{a}[r][r][0.95]{$x_0$}
	\psfrag{b}[r][r][0.95]{$x_1$}
	\psfrag{c}[r][r][0.95]{$x_2$}
	\psfrag{d}[r][r][0.95]{$x_3$}
	\psfrag{A}[l][l][0.95]{$y_0$}
	\psfrag{B}[l][l][0.95]{$y_1$}
	\psfrag{C}[l][l][0.95]{$y_2$}
	\psfrag{D}[l][l][0.95]{$y_3$}
	\psfrag{P}[c][c][0.9]{$\phi_{00}$}
	\psfrag{Q}[c][c][0.9]{$\phi_{01}$}
	\psfrag{R}[c][c][0.9]{$\phi_{10}$}
	\psfrag{S}[c][c][0.9]{$\phi_{11}$}
	\psfrag{X}[c][c][1.0]{(a)}
	\psfrag{Y}[c][c][1.0]{(b)}
	\includegraphics[width=65mm]{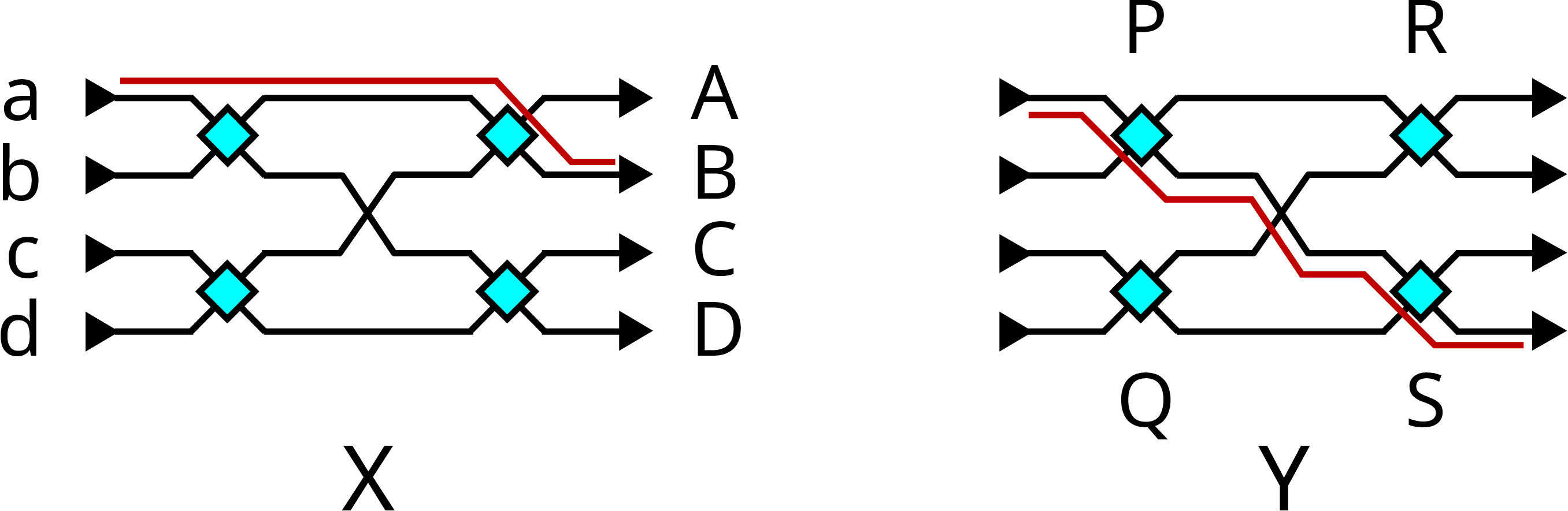}
\end{center}
\vspace*{-2mm}
\caption{Unique paths between input and output ports in an FGB architecture
with four inputs and two stages: (a) the path from $x_0$ to $y_1$ and (b) the
path from $x_0$ to $y_3$. Parameter $\phi_{su}$ is the phase angle of the PSBS
unit at the $s$th stage and the $u$th position from the top.
}\label{fig:bflypath}
\end{figure}

The FGB architecture with $n$ input ports 
inherits the structural properties of the butterfly topology:
it provides a unique path between every input-output pair and 
satisfies the reachability property using only $\log_2 n$ stages,
each containing exactly $n/2$ PSBS units.
Consequently, the architecture addresses the first two limitations using only
$(n/2)\log_2 n$ tunable parameters, substantially fewer than the $n^2$
parameters required by a universal conventional mesh.
As illustrated in Fig.~\ref{fig:bflypath}, 
the butterfly block provides a unique path between every input-output pair 
and satisfies reachability.
Furthermore, each path contains the same number of PSBS units,
equal to the number of stages.
As shown below, this uniform path structure produces statistical uniformity,
while the fine-grained building block enhances mixing capability per parameter.

We relate these structural properties to the statistical uniformity of the
transformation through Eq.~(\ref{eq:expected}), whose extension to the FGB
architecture is derived in Appendix~\ref{app_ssec:heatmap_psbsbfly}.
For an FGB architecture with a single butterfly block,
$\mathbb{E}[|g_{jh}|^2]\!=\!1/n$ for all $j$ and $h$. Every input-output pair
has one path, $|P_{(j,h)}|\!=\!1$, and every path contains
$F(p)\!=\!\log_2 n$ PSBS units. Hence,
\begin{equation}
\mathbb{E}[\,| g_{jh}|^2\,] = \frac{1}{2^{\,(\log_{2} n)}} = \frac{1}{n}\,.
\label{eq:randbfly_expected}
\end{equation}
For $b\!\in\!\mathbb{N}$ butterfly blocks, every input-output pair has
$|P_{(j,h)}|\!=\!n^{b-1}$ paths, and every path has
$F(p)\!=\!b\log_{2}n$. The expectation is therefore
\begin{equation}
\mathbb{E}[\,| g_{jh}|^2\,] = n^{b-1}\,\frac{1}{2^{\,(b\log_{2} n)}} = \frac{1}{n}\,.
\label{eq:b_randbfly_expected}
\end{equation}
Thus, the FGB path structure yields an expected squared modulus of $1/n$ for
every entry, consistent with the unitary normalization $G^{\dagger}G\!=\!I$.
This entrywise equality establishes the statistical uniformity considered in
this work.

\begin{figure}[t]
  \centering
	\captionsetup[subfloat]{margin={2pt,0pt}}
	\subfloat[$m\!=\!16$, $b\!=\!4$]{
  	  \includegraphics[width=0.18\columnwidth]{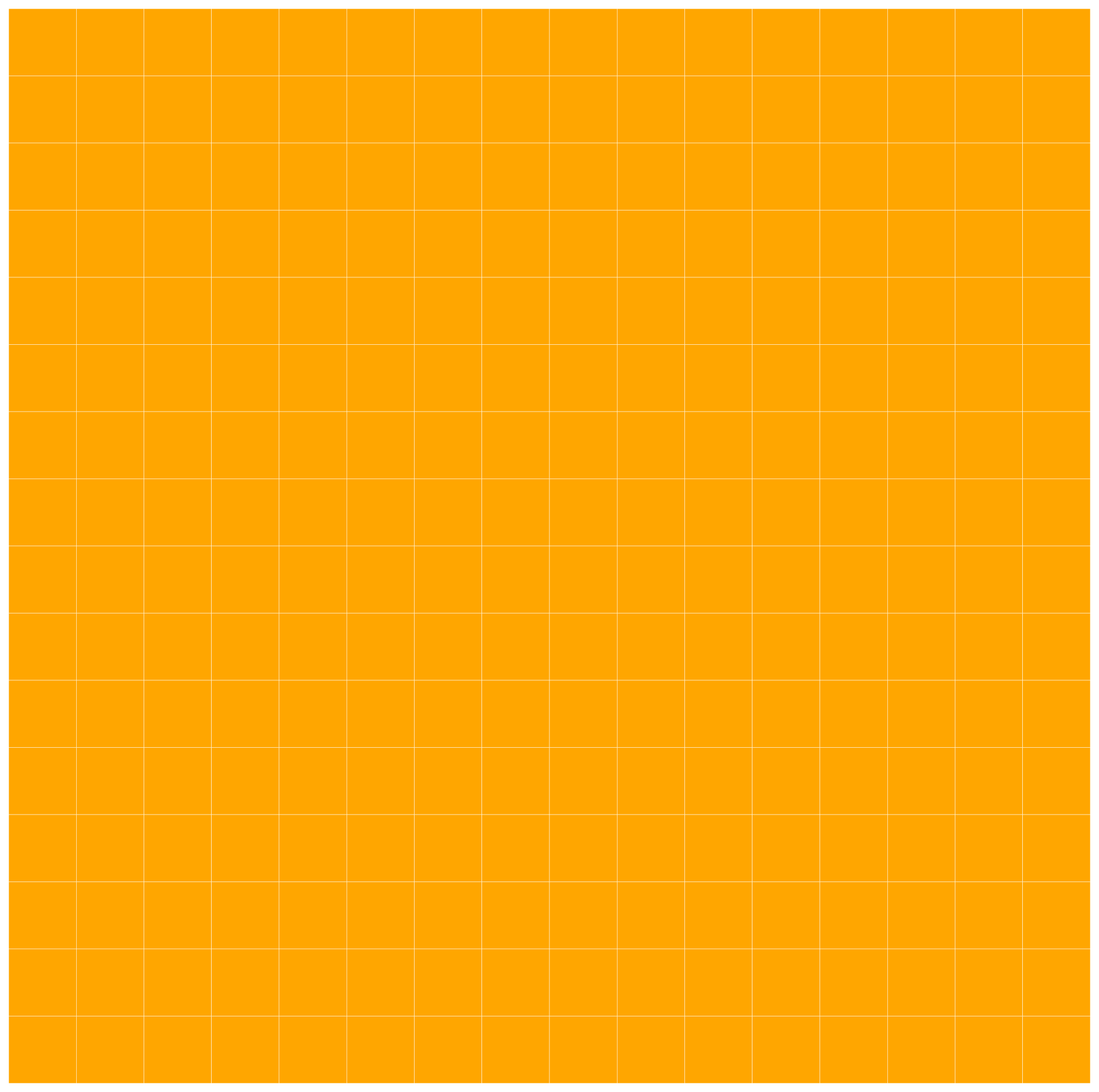}
	}
	\subfloat[$m\!=\!12$, $b\!=\!3$]{
  	  \includegraphics[width=0.18\columnwidth]{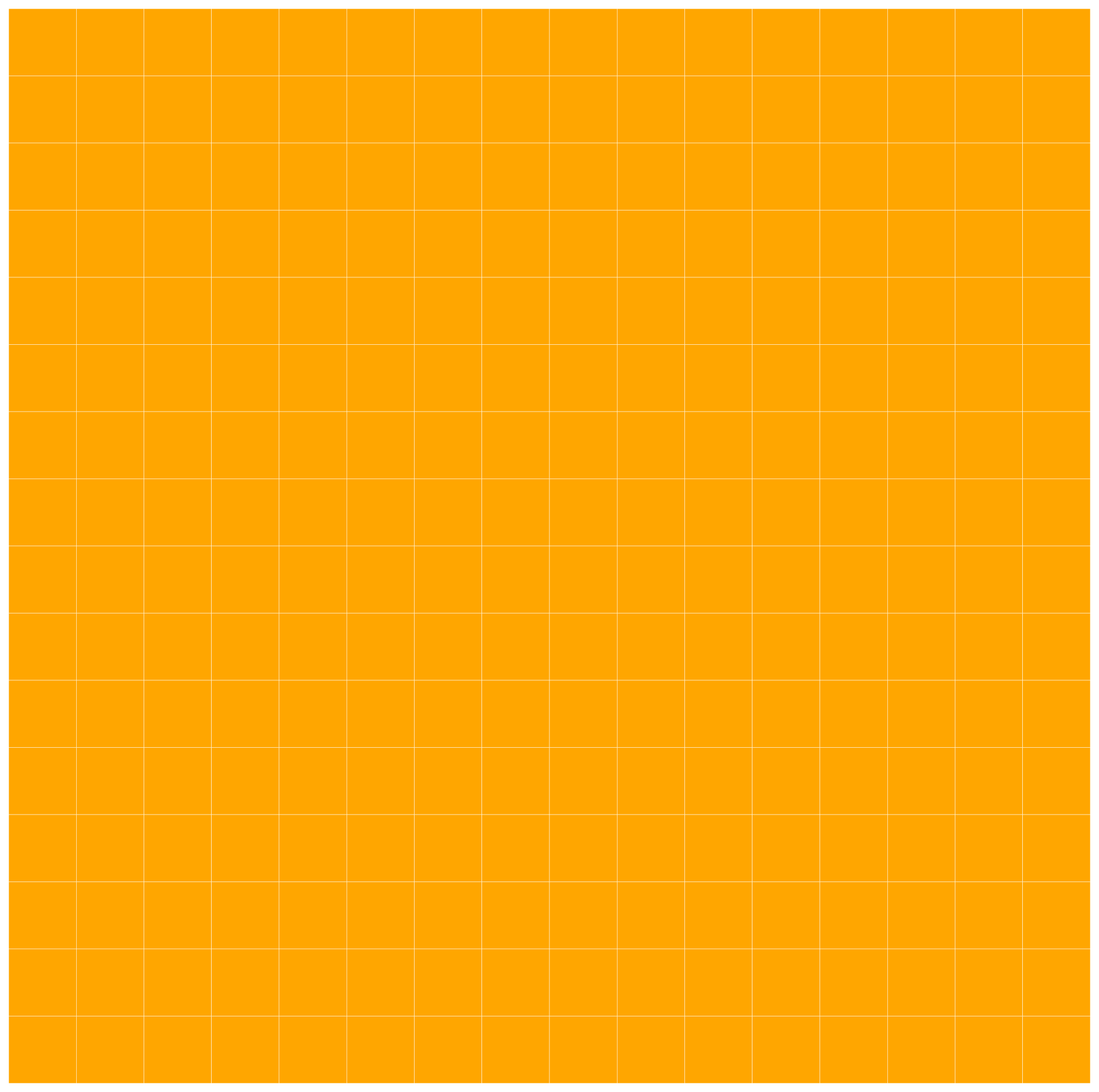}
	}
	\subfloat[$m\!=\!8$, $b\!=\!2$]{
  	  \includegraphics[width=0.18\columnwidth]{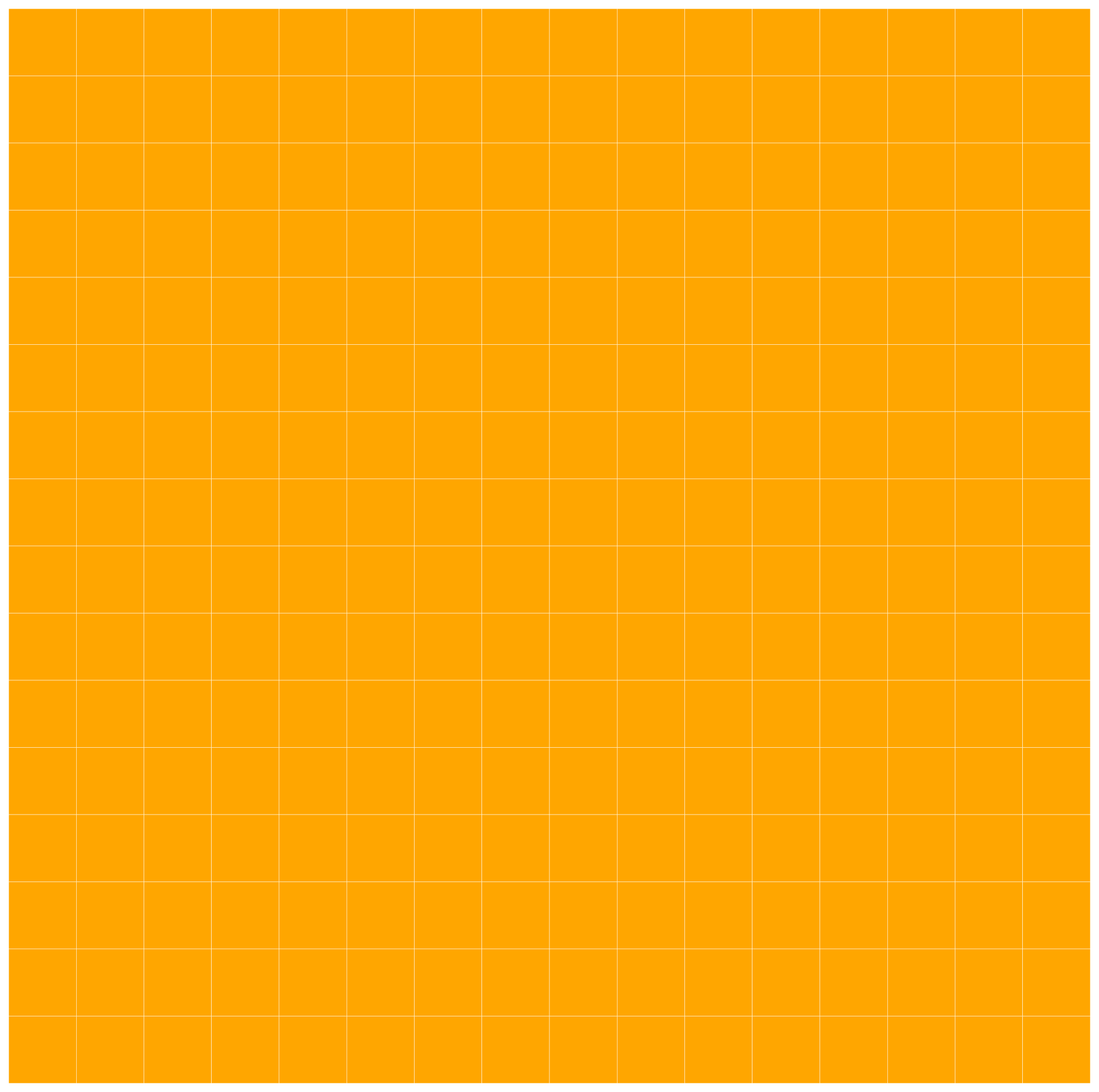}
	}
	\subfloat[$m\!=\!4$, $b\!=\!1$]{
  	  \includegraphics[width=0.18\columnwidth]{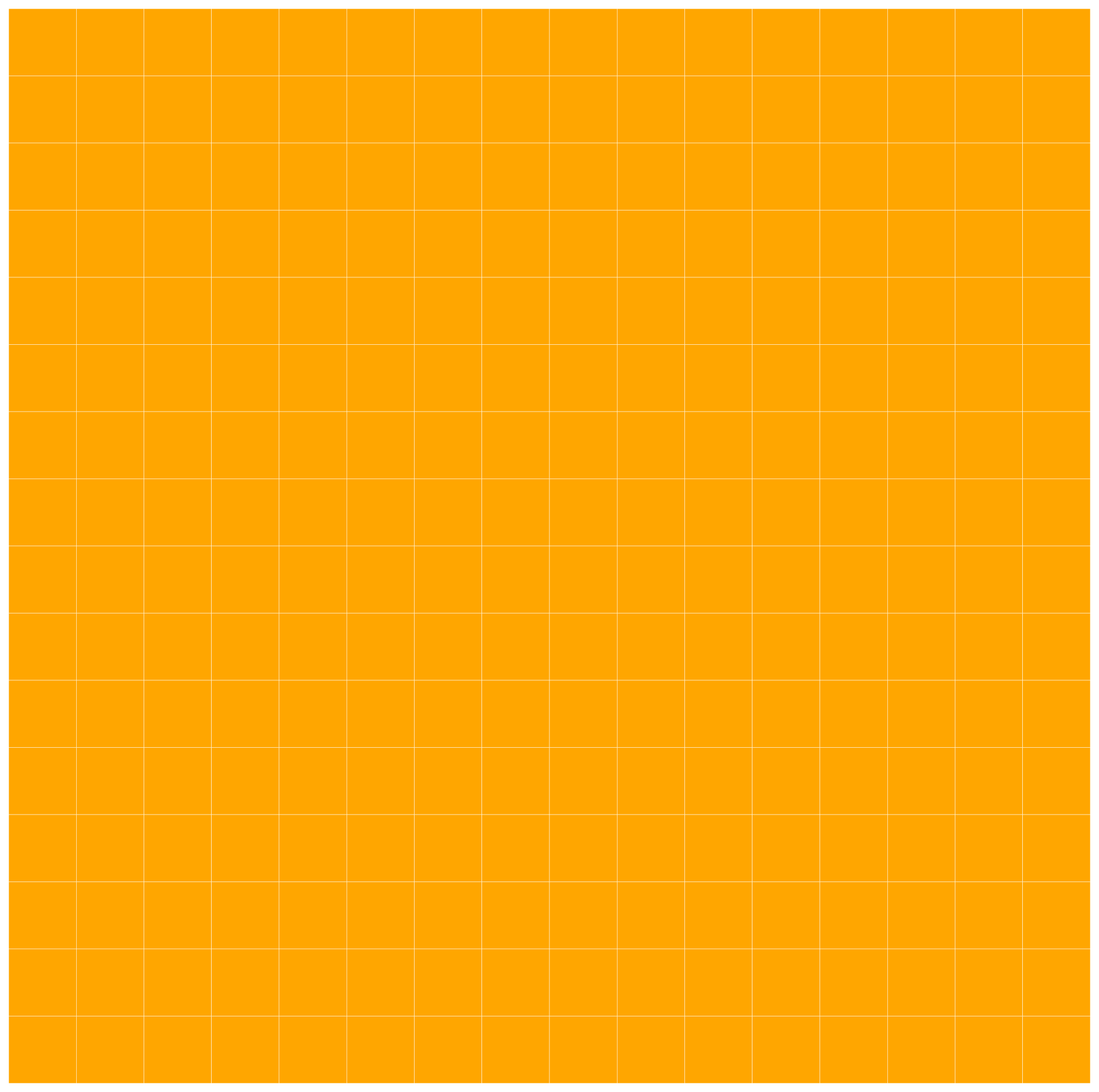}
	}
	\subfloat{
		\psfrag{a}[l][l][0.7]{6.15}
		\psfrag{d}[l][l][0.7]{6.25}
		\psfrag{g}[l][l][0.7]{6.35}
		\psfrag{h}[l][l][0.72]{$\times 10^{-2}$}
  	  \includegraphics[height=0.21\columnwidth]{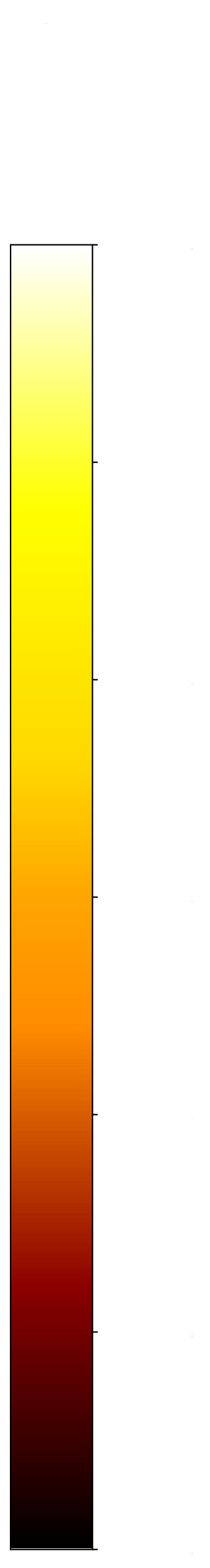}
	}
	\\[-3mm]
	\setcounter{subfigure}{4}
	\subfloat[$m\!=\!16$, $b\!=\!4$]{ 
  	  \includegraphics[width=0.18\columnwidth]{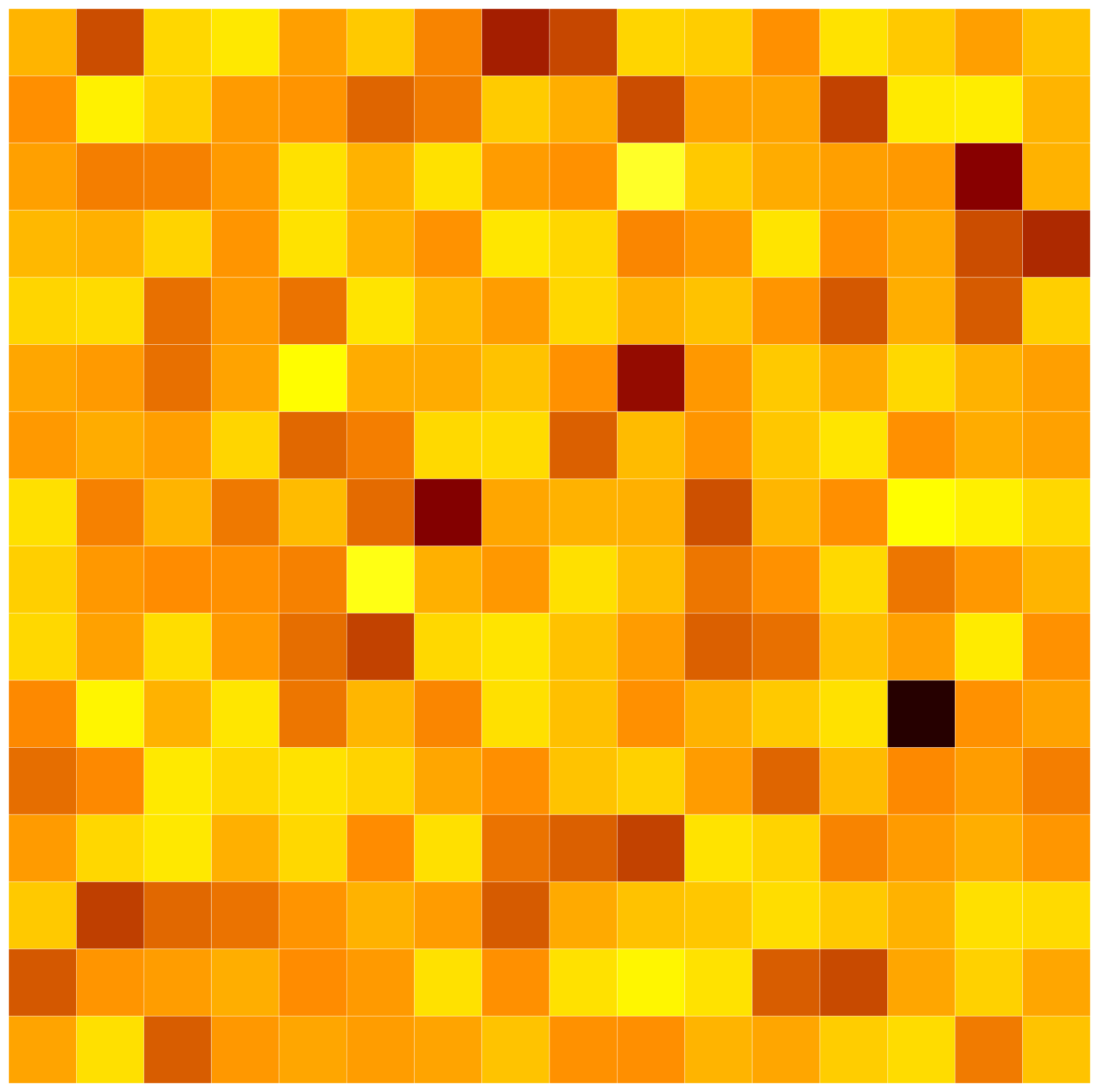}
	}
	\subfloat[$m\!=\!12$, $b\!=\!3$]{
  	  \includegraphics[width=0.18\columnwidth]{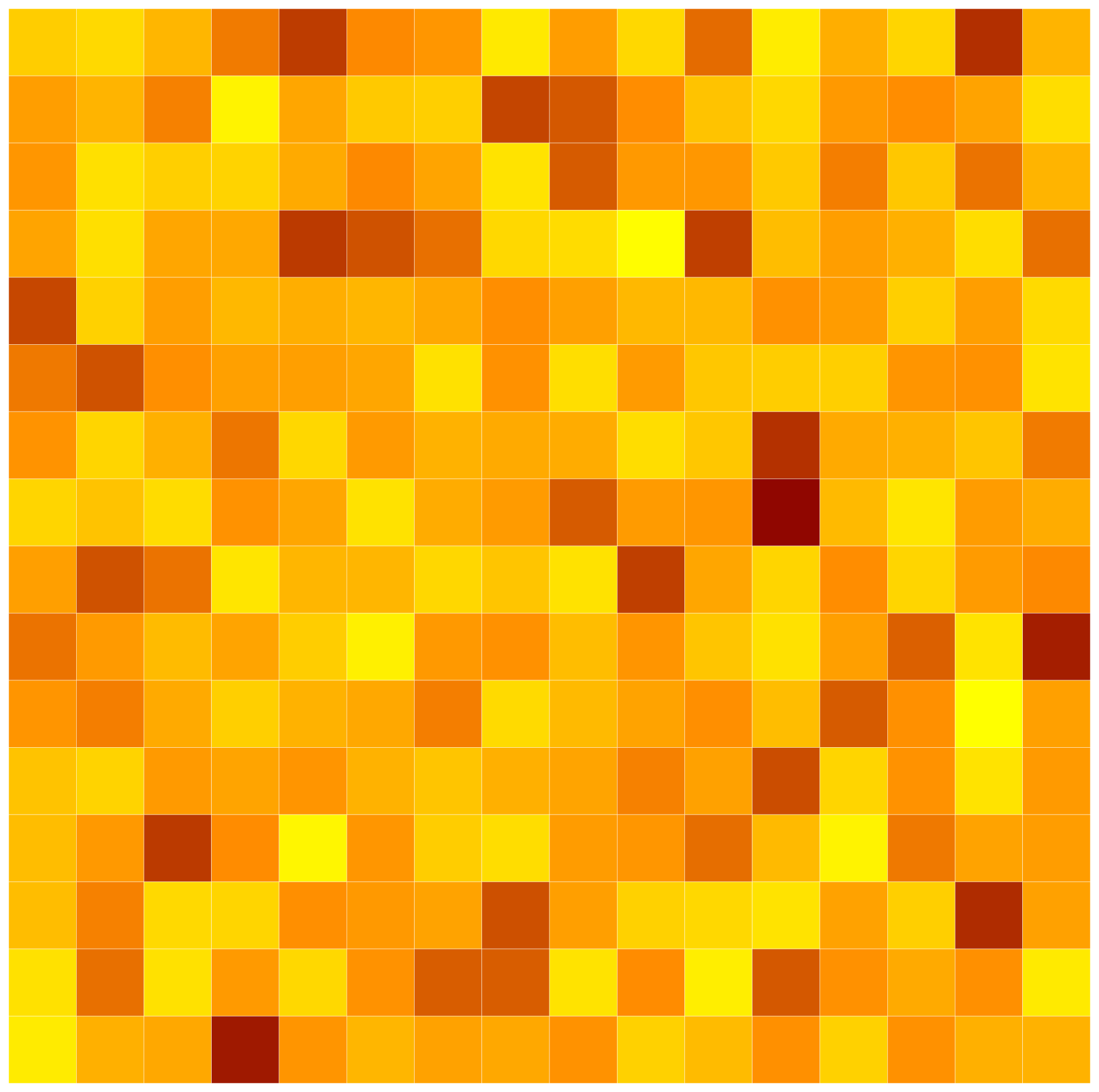}
	}
	\subfloat[$m\!=\!8$, $b\!=\!2$]{
  	  \includegraphics[width=0.18\columnwidth]{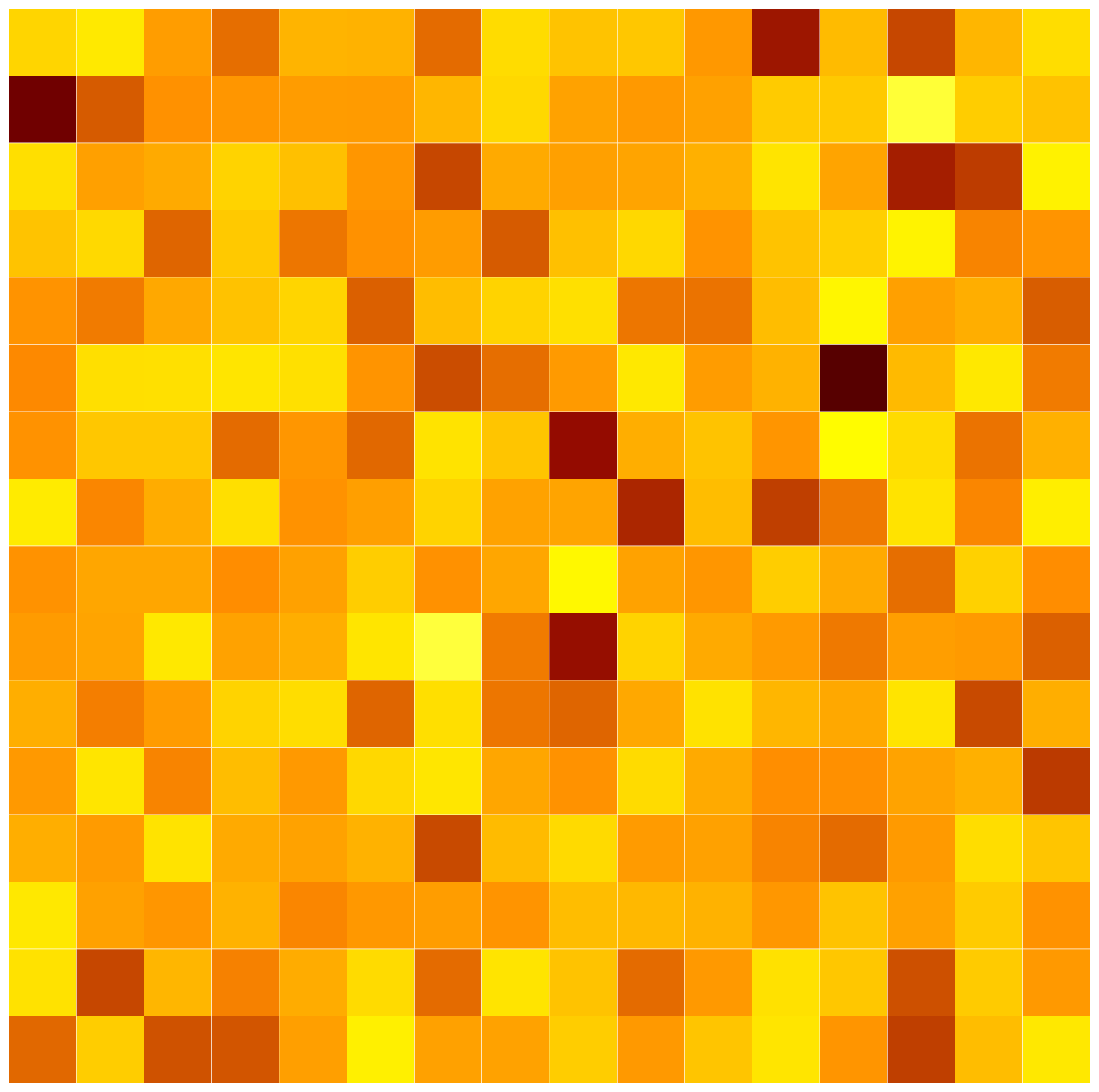}
	}
	\subfloat[$m\!=\!4$, $b\!=\!1$]{
  	  \includegraphics[width=0.18\columnwidth]{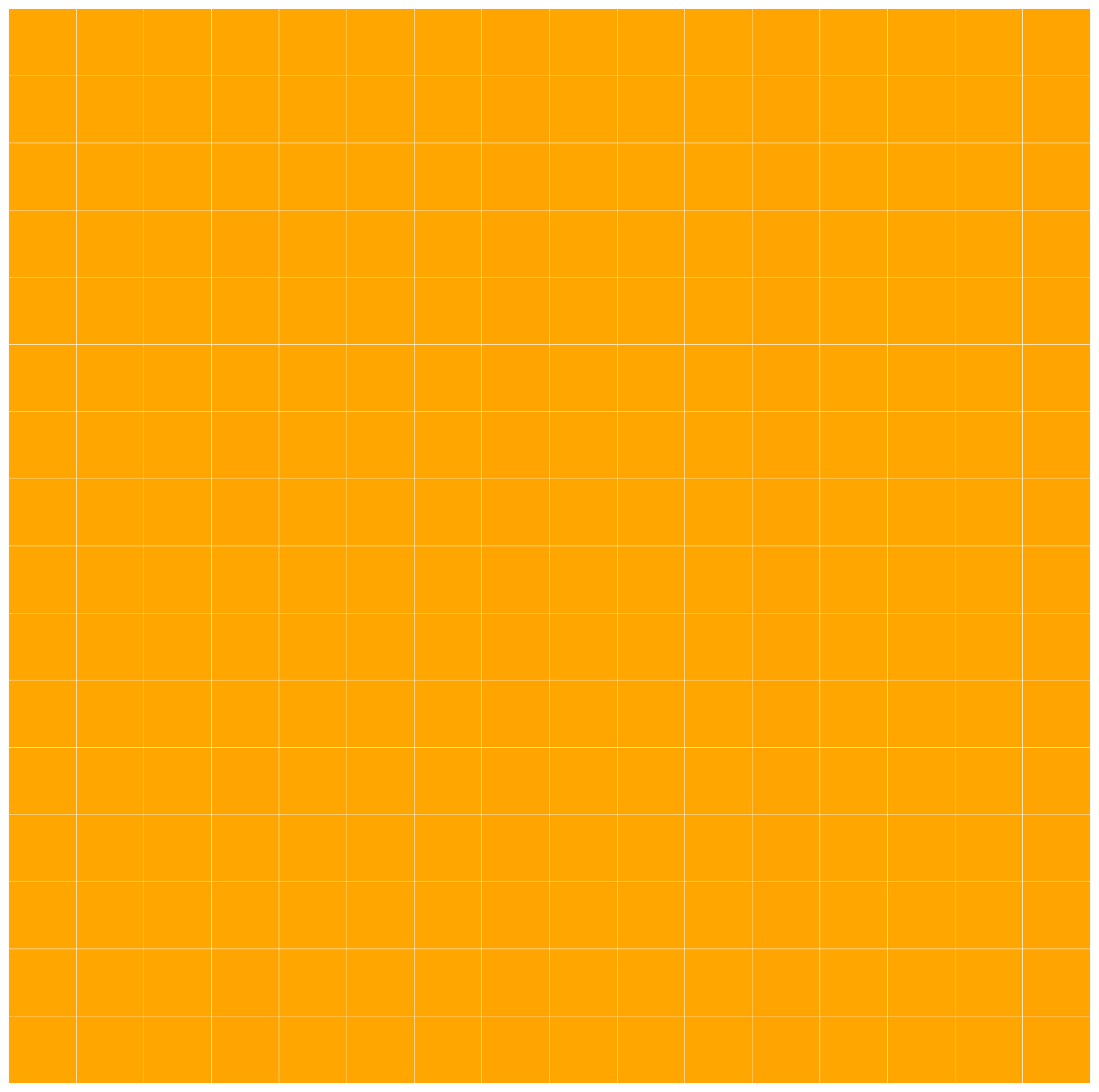}
	}
	\subfloat{
		\psfrag{a}[l][l][0.7]{6.15}
		\psfrag{d}[l][l][0.7]{6.25}
		\psfrag{g}[l][l][0.7]{6.35}
		\psfrag{h}[l][l][0.72]{$\times 10^{-2}$}
  	  \includegraphics[height=0.21\columnwidth]{./Figs/colorbar_n16randbfly.eps}
	}
  \caption{Analytical expectations $\widehat{M}$ (top) and computational averages 
$\overline{M}$ (bottom)
of the squared moduli of the matrix entries
for the FGB architecture with $n\!=\!16$, stage counts $m\!=\!\{16,12,8,4\}$, 
and corresponding butterfly-block counts $b\!=\!\{4,3,2,1\}$.
The computational averages are calculated over $10^{6}$ randomly initialized instances.
The horizontal and vertical axes correspond to input (column) and output (row) indices, 
respectively.
Orange corresponds to $0.0625\!=\!1/n$, and the color scale spans the narrow range
from black (low: $6.15\!\times\!10^{-2}$\,) 
through red and yellow to white (high: $6.35\!\times\!10^{-2}$\,).
}
  \label{fig:heatmap_psbsbfly}
\vspace*{-3mm}
\end{figure}

Figure~\ref{fig:heatmap_psbsbfly} shows 
the analytical expectations $\widehat{M}$ (top) and the computational averages 
$\overline{M}$ (bottom) of the squared moduli of matrix entries
for the FGB architecture with $n\!=\!16$, stage counts $m\!=\!\{16,12,8,4\}$, 
and butterfly-block counts $b\!=\!\{4,3,2,1\}$, respectively.
The expectations were calculated based on Eq.~(\ref{eq:b_randbfly_expected}) and 
the average values were computed over $10^{6}$ instances
whose parameters are initialized uniformly at random.
The RMSE is calculated directly from the $n^2$ entries using
Eq.~(\ref{eq:rmse}); no assumption about their distribution is required.
\begin{table}[t]
\centering
\caption{Root-mean-square error between $\overline{M}$ and $\widehat{M}$}
\label{table:bfly_heatmap_error}
\renewcommand{\arraystretch}{1.4}
\resizebox{\columnwidth}{!}{%
\begin{tabular}{ccccc}\hline
Stage count $m$ &          16 & 12 & 8 & 4\\ \hline 
RMSE $\downarrow$ & $1.52\!\times\!10^{-5}$ & $1.39\!\times\!10^{-5}$ &
$1.53\!\times\!10^{-5}$ & $1.06\!\times\!10^{-10}$\\ \hline
\end{tabular}
}
\end{table}
Table~\ref{table:bfly_heatmap_error} summarizes the RMSE values 
between $\overline{M}$ and $\widehat{M}$ for $m\!=\!16,12,8,4$.
The small RMSE values indicate close agreement between 
the analytical expectations and the computational averages.
These results confirm that the FGB architecture maintains statistical uniformity 
even in parameter-constrained settings.

Thus, the proposed FGB architecture addresses the limitations of the
rectangular mesh architecture while providing a more regular, low-depth design.
In the following section, 
we present computational results
demonstrating its effectiveness
in Haar-random matrix reconstruction and neural-network classification tasks.

\section{Numerical Studies}\label{sec:exp}
We evaluate the performance of the proposed 
FGB architecture through computational experiments.
This evaluation directly tests 
the key architectural requirements 
of statistical uniformity and mixing capability per parameter.
First, we investigate the expressive power of parameterized unitary matrices 
realized by the FGB and conventional mesh architectures 
via Haar-random matrix reconstruction.
To isolate the effects of topology and fundamental building blocks, 
we consider two additional architectures:
an MZI-based butterfly similar to the fast-Fourier-transform (FFT) architecture 
\cite{jing}\footnote{
The FFT architecture belongs to the family of butterfly topologies described
in Appendix~\ref{app_sec:bfly} but it differs from the standard topology in
Fig.~\ref{fig:std_bflyarch}.
}
and a PSBS-based mesh similar to the single-static-beam-splitter architecture
\cite{fldzhyan:20}\footnote{
This architecture uses beam-splitter--phase-shifter (BSPS) pairs 
in a $2n$-stage configuration 
that can implement an arbitrary $n\!\times\!n$ unitary matrix; 
truncated variants were not considered in that work.
}.
Next, we apply these architectures as linear units 
in both feedforward and recurrent neural networks.
We evaluate performance in terms of test accuracy
on the Fashion-MNIST classification task \cite{fmnist},
varying the number of parameters in the neural networks.
As baselines, we consider 
the MZI-based mesh architecture and its truncated variants 
with reduced parameter counts.
We focus on parameter-constrained settings, 
where the number of parameters is insufficient to represent 
arbitrary unitary transformations.
This setting reflects practical constraints in physical implementations 
of ONNs.

\subsection{Matrix Reconstruction Task}\label{ssec:reconst}
In the matrix reconstruction task, we optimize the parameters of an
$n\!\times\!n$ unitary matrix $G$ implemented by a given architecture so that
$G$ approximates a target $n\!\times\!n$ Haar-random unitary matrix $U$
generated using the method of Ref.~\cite{mezzadri}.
To evaluate performance under varying parameter budgets,
we control the number of parameters in each architecture.
For the FGB (PSBS-butterfly), 
this is achieved by varying the number of butterfly blocks, 
each consisting of $\log_2 n$ stages.
For the MZI-based rectangular mesh (MZI-mesh),
the number of parameters is reduced by truncating the number of MZI stages.
As the objective function,
we use the normalized Frobenius error between $G$ and $U$,
defined as
\begin{align}
L_F^{*}(G,U) 
&= \frac{1}{\sqrt{2n}}\,\| G-U\|_F \notag\\
&= \left\{ 1- \frac{1}{n}\operatorname{Re}\!\left[\operatorname{Tr}(G^{\dagger}U)\right] \right\}^{1/2},
\label{eq:objfunct}
\end{align}
where $\|\cdot\|_F$ denotes the Frobenius norm,
$\operatorname{Tr}(\cdot)$ is the trace operator,
and $\operatorname{Re}(\cdot)$ denotes the real part.
The normalization factor $\sqrt{2n}$ ensures that $L_F^{*}(G,U)\!\in\![0,\sqrt{2}\,]$, 
with larger values indicating greater dissimilarity between $G$ and $U$.
When $G$ is sampled independently of $U$,
$L_F^{*}(G,U)\!\approx\!1$ 
while $L_F^{*}(G,U)\!=\!\sqrt{2}$ corresponds to the maximally misaligned case 
 $G\!=\!-U$.
These relationships, including Eq.~(\ref{eq:objfunct}), are derived 
in Appendix~\ref{app_sec:reconst} 
and are interpreted geometrically in Appendix~\ref{app_sec:cossim}.
We quantify the expressive power of the architecture implementing $G$ by the
average normalized Frobenius error over Haar-random targets and their
corresponding optimized matrices:
\begin{equation}
\overline{L_F^{*}}(G) = \frac{1}{|\mathcal{U}_{s}|}\sum_{U\in \mathcal{U}_{s}}L_F^{*}(G,U)
\, ,\label{eq:exp_power}
\end{equation}
where $\mathcal{U}_{s}$ denotes a set of Haar-random target matrices.
A parameterized unitary matrix $G$ is considered to have higher expressive power 
if it achieves a smaller value of $\overline{L_F^{*}}(G)$.
Hereafter, the argument $G$ is omitted unless otherwise stated.

\begin{figure}[t]
\begin{center}
	\psfrag{X}[c][c][1.0]{Normalized number of parameters: NNP}
	\psfrag{Y}[c][c][0.95]{
		\begin{picture}(0,0)
			\put(0,0){\makebox(-6,55)[c]{
					Normalized Frobenius error}}
			\put(0,8){\makebox{$\overline{L_F^{*}}$}}
		\end{picture}
	}
	\psfrag{P}[c][c][0.87]{0}   \psfrag{Q}[c][c][0.87]{0.2} 
	\psfrag{R}[c][c][0.87]{0.4} \psfrag{S}[c][c][0.87]{0.6} 
	\psfrag{T}[c][c][0.87]{0.8} \psfrag{U}[c][c][0.87]{1.0}
	\psfrag{E}[r][r][0.87]{0}   \psfrag{F}[r][r][0.87]{0.2} 
	\psfrag{G}[r][r][0.87]{0.4} \psfrag{H}[r][r][0.87]{0.6} 
	\psfrag{K}[r][r][0.87]{0.8} \psfrag{L}[r][r][0.87]{1.0}
	\psfrag{M}[r][r][0.82]{16-mesh}
	\psfrag{N}[r][r][0.82]{32-mesh}
	\psfrag{O}[r][r][0.82]{64-mesh}
	\psfrag{A}[r][r][0.82]{16-FGB}
	\psfrag{B}[r][r][0.82]{32-FGB}
	\psfrag{C}[r][r][0.82]{64-FGB}
	\includegraphics[width=68mm]{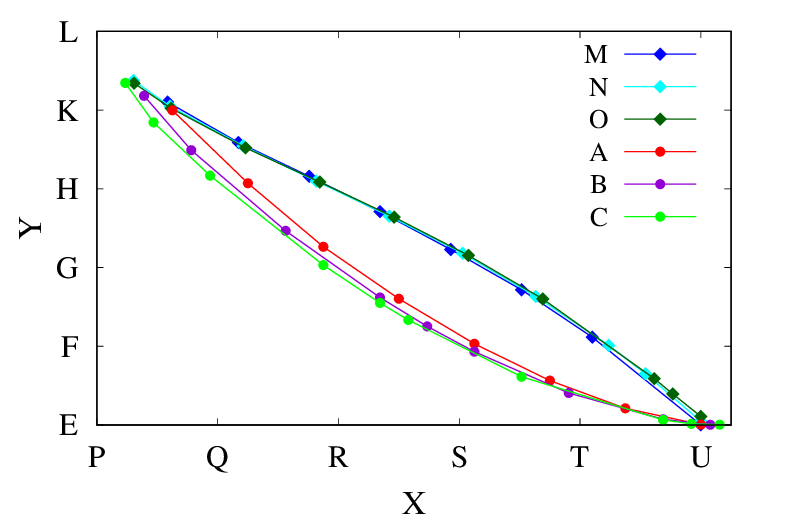}
\end{center}
\caption{Average normalized Frobenius error between target
$n\!\times\!n$ Haar-random unitary matrices ($n\!=\!16,32,64$) and matrices
implemented by the FGB and MZI-based rectangular mesh architectures, plotted
against the normalized number of parameters (NNP). The parameter count is
normalized by $n^2$, and the Frobenius error by $\sqrt{2n}$.
}
\label{fig:reconst_all}
\end{figure}

Figure~\ref{fig:reconst_all} shows the average normalized Frobenius error
$\overline{L_F^{*}}$ as a function of the normalized number of parameters (NNP),
for $n\!\times\!n$ Haar-random targets $U$ with $n\!=\!16,32,64$ and
parameterized matrices $G$ implemented by the FGB and MZI-mesh architectures.
The parameter count is normalized by $n^2$, 
corresponding to the degrees of freedom of an $n\!\times\!n$ unitary matrix.
Each value of $\overline{L_F^{*}}$ is averaged over 100 independent trials
with different target matrices $U$. The MZI-mesh parameters are initialized
using the Haar initialization method of Ref.~\cite{pai}.
\footnote{
Truncated variants are constructed by first initializing a full parameterized mesh
using the Haar initialization method, and then reducing the number of stages 
by removing selected stages while preserving the initialized parameters.
}
We adopt this initialization to provide a favorable, statistically uniform baseline.
By contrast, we employ uniform random initialization for the FGB architecture 
because its structural properties do not require specialized initialization
to achieve statistical uniformity.

The results exhibit two notable features:
(1) two distinct groups of reconstruction-error curves
associated with the two architectures 
and (2) convergence of the curves at both extremes of the parameter range.
The upper group corresponds to the MZI-based rectangular mesh architecture 
and its truncated variants,
whereas the lower group corresponds to the FGB architecture.
Their consistent separation across different $n$ indicates that 
the observed behavior is largely architecture-dependent rather than size-dependent.
The lower error achieved by the FGB architecture indicates
greater expressive power than the mesh architectures,
particularly in parameter-constrained settings.
Furthermore, the curves approach one another at both ends of the evaluated
parameter range: in the high-budget regime (NNP $\gtrsim\!1$), where the
reconstruction errors approach zero, and in the most constrained regime,
where both architectures have similarly limited expressive power.\footnote{
Although no explicit decomposition algorithm is currently available for the FGB architecture,
the observed values of $\overline{L_F^{*}}$ indicate that 
it can approximate the sampled target unitary matrices with high precision,
given sufficient parameters.
}

\begin{figure}[t]
\begin{center}
	\psfrag{X}[c][c][1.0]{Normalized number of parameters: NNP}
	\psfrag{Y}[c][c][0.95]{
		\begin{picture}(0,0)
			\put(0,0){\makebox(-6,55)[c]{
					Normalized Frobenius error}}
			\put(0,8){\makebox{$\overline{L_F^{*}}$}}
		\end{picture}
	}
	\psfrag{P}[c][c][0.87]{0}   \psfrag{Q}[c][c][0.87]{0.2} 
	\psfrag{R}[r][c][0.87]{0.4} \psfrag{S}[l][c][0.87]{0.6} 
	\psfrag{W}[c][c][0.87]{0.5}
	\psfrag{T}[c][c][0.87]{0.8} \psfrag{U}[c][c][0.87]{1.0}
	\psfrag{E}[r][r][0.87]{0}   \psfrag{F}[r][r][0.87]{0.2} 
	\psfrag{G}[r][r][0.87]{0.4} \psfrag{H}[r][r][0.87]{0.6} 
	\psfrag{K}[r][r][0.87]{0.8} \psfrag{L}[r][r][0.87]{1.0}
	\psfrag{A}[r][r][0.82]{MZI-mesh}
	\psfrag{B}[r][r][0.82]{PSBS-mesh}
	\psfrag{C}[r][r][0.82]{MZI-butterfly}
	\psfrag{D}[r][r][0.82]{PSBS-butterfly}
	\includegraphics[width=68mm]{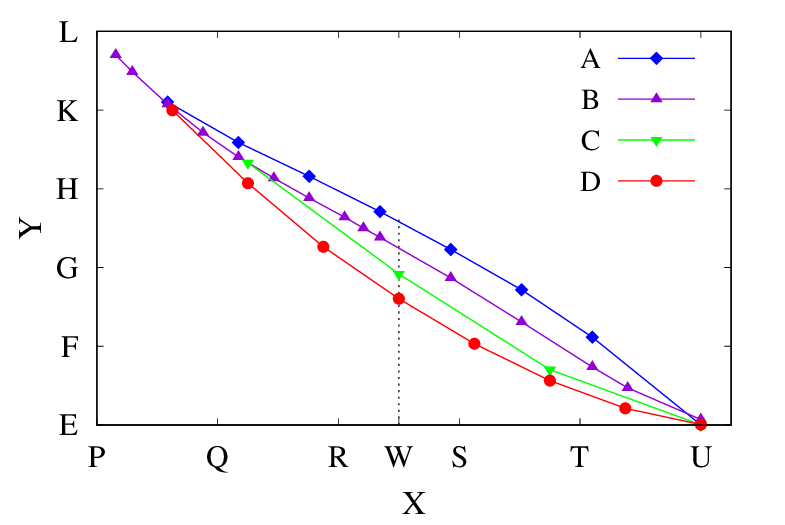}
\end{center}
\caption{Expressive power of four architectures---MZI-mesh, PSBS-mesh,
MZI-butterfly, and PSBS-butterfly---on $16\!\times\!16$ Haar-random unitary targets.
The average normalized Frobenius error $\overline{L_F^{*}}$ is plotted
as a function of the normalized number of parameters (NNP).
In the parameter-constrained setting, the reconstruction error decreases in the order
MZI-mesh, PSBS-mesh, MZI-butterfly, and PSBS-butterfly.
}
\label{fig:reconst_n16}
\end{figure}

To isolate the effects of topology and fundamental building blocks,
we compare four architectures:
PSBS-butterfly (FGB), MZI-butterfly, PSBS-mesh, and MZI-mesh.
The butterfly topology primarily promotes statistical uniformity,
while the PSBS building block enhances mixing capability per parameter.
Figure~\ref{fig:reconst_n16} shows $\overline{L_F^{*}}$ as a function of NNP 
for $16\!\times\!16$ Haar-random unitary targets.
The MZI-mesh parameters are set using Haar initialization, whereas those of
the other architectures are initialized uniformly at random.
In the parameter-constrained setting,
$\overline{L_F^{*}}$ decreases (i.e., expressive power increases) 
in the order MZI-mesh, PSBS-mesh, MZI-butterfly, and PSBS-butterfly.
This ordering indicates that 
the topology and building block each contribute to expressive power, 
and that their combination yields the greatest benefit.

When $\text{NNP}\!<\!1$,
the MZI-mesh exhibits the lowest expressive power,
as it lacks both statistical uniformity and sufficient mixing capability
due to its structural redundancy and, in truncated configurations, 
loss of reachability.
The PSBS-mesh maintains reachability for $\text{NNP}\!\geq\!0.5$ but still
suffers from degraded statistical uniformity because of its mesh topology,
which yields multiple paths between input--output pairs and nonuniform numbers
of PSBS building blocks along those paths, as reflected in Eq.~(\ref{eq:expected}).
The MZI-butterfly improves statistical uniformity relative to the mesh
architectures through its butterfly topology;
however, at fixed parameter count,
it contains fewer stages (or equivalently, fewer mixing operations) than
the PSBS-butterfly, 
resulting in lower mixing capability per parameter.
Consequently,
the proposed FGB (PSBS-butterfly) architecture achieves the highest expressive power 
in the parameter-constrained setting,
owing to the combination of statistical uniformity provided by the butterfly topology 
and efficient mixing enabled by the PSBS building blocks.

\subsection{Classification Using Feedforward Neural Networks}\label{ssec:fnn}
\begin{figure}[t]
\begin{center}
	\psfrag{X}[c][c][1.0]{Number of parameters}
	\psfrag{Y}[c][c][1.0]{
		\begin{picture}(0,0)
			\put(0,0){\makebox(-6,18)[c]{Test accuracy}}
		\end{picture}
	}
	\psfrag{Q}[c][c][0.87]{50} 
	\psfrag{R}[c][c][0.87]{100} \psfrag{S}[c][c][0.87]{150} 
	\psfrag{T}[c][c][0.87]{200} \psfrag{U}[c][c][0.87]{250}
	\psfrag{E}[r][r][0.87]{0.2} \psfrag{F}[r][r][0.87]{0.3} 
	\psfrag{G}[r][r][0.87]{0.4} \psfrag{H}[r][r][0.87]{0.5} 
	\psfrag{J}[r][r][0.87]{0.6}
	\psfrag{K}[r][r][0.87]{0.7} \psfrag{L}[r][r][0.87]{0.8}
	\psfrag{A}[r][r][0.82]{MZI-mesh}
	\psfrag{B}[r][r][0.82]{PSBS-mesh}
	\psfrag{C}[r][r][0.82]{MZI-butterfly}
	\psfrag{D}[r][r][0.82]{PSBS-butterfly}
	\includegraphics[width=65mm]{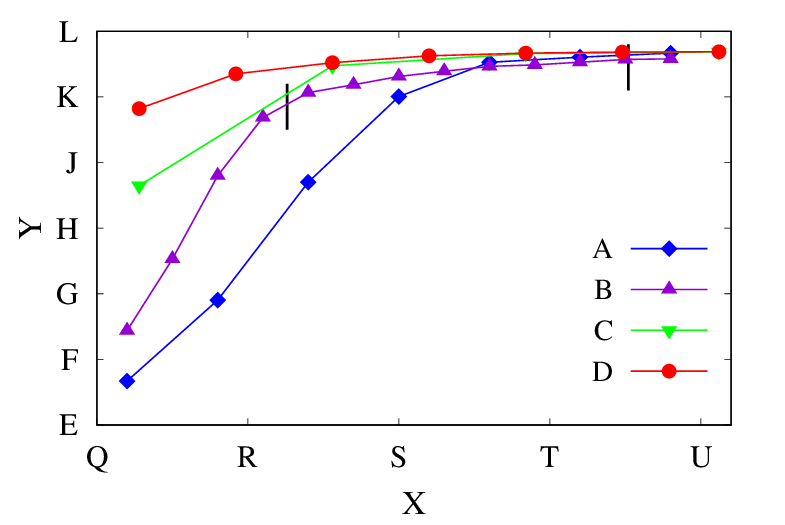}
\end{center}
\caption{Test accuracy as a function of parameter count for four
architectures---MZI-mesh, PSBS-mesh, MZI-butterfly, and PSBS-butterfly
(FGB)---used as the linear unit in a single-layer perceptron model
with $n\!=\!16$ on the Fashion-MNIST classification task.
}
\label{fig:1unit_fnn_testacc}
\end{figure}

We evaluate performance on the Fashion-MNIST classification task \cite{fmnist}
using feedforward neural networks (FNNs) with four architectures:
MZI-mesh, PSBS-mesh, MZI-butterfly, and PSBS-butterfly (FGB).
Each FNN consists of two components: a preprocessing block and a main block.
The main block is a single-layer perceptron comprising a linear unit
implemented by one of these architectures with 16 input and 16 output ports.
A subsequent nonlinear function yields real-valued outputs by measuring the
optical power at each port, thereby modeling photodetection.
A standard classification unit is then applied to these outputs.
The preprocessing block maps each Fashion-MNIST image 
to a 16-dimensional complex-valued vector using a discrete Fourier transform,
ensuring compatibility with the input dimension of the linear unit. 
This system diagram is similar to that illustrated in Fig.~\ref{app_fig:fnn_syst}.

Figure~\ref{fig:1unit_fnn_testacc} shows the resulting test accuracy 
as a function of the parameter count, ranging from $60$ to $256$, 
where 256 corresponds to the number of degrees of freedom of
a $16\!\times\!16$ unitary matrix.\footnote{
The maximum parameter count for the mesh-based architectures is set to 240
by omitting the final 16 phase shifters
because the final phase shifts do not affect output power.
}
Each reported test accuracy is averaged over 16 runs with different
initializations and evaluated after 200 training epochs.
The MZI-mesh parameters are set using the Haar initialization method \cite{pai},
whereas those of the other architectures are initialized uniformly at random.
The parameters are optimized using PyTorch \cite{pytorch} 
with the Adam optimizer \cite{kingma}.

The results show that, at a given parameter count, the PSBS-butterfly
architecture consistently achieves higher test accuracy than the other architectures.
It also exhibits stable performance, whereas the other architectures undergo
sharp degradation at certain parameter counts.
For the mesh-based architectures, performance degradation occurs with 
the loss of reachability, as indicated by the vertical markers in the figure.
In the parameter-constrained setting, this loss substantially reduces
classification accuracy.
By contrast, at the lowest parameter counts (the leftmost point in the figure),
the MZI-butterfly shows reduced performance due to insufficient mixing,
because it uses only one butterfly block, whereas the PSBS-butterfly uses two.
These results indicate that both strong mixing capability per parameter and
the reachability required for statistical uniformity are important for
classification performance.
This behavior is consistent with the trends observed in the matrix reconstruction task.

\begin{figure}[t]
\begin{center}
	\psfrag{X}[c][c][1.0]{Number of parameters (log scale)}
	\psfrag{Y}[c][c][1.0]{
		\begin{picture}(0,0)
			\put(0,0){\makebox(-5,22)[c]{
				Test accuracy}}
		\end{picture}
	}
	\psfrag{A}[c][c][0.88]{$10^2$}
	\psfrag{B}[c][c][0.88]{$10^3$}
	\psfrag{C}[c][c][0.88]{$10^4$}
	\psfrag{D}[r][r][0.88]{0.4}
	\psfrag{E}[r][r][0.88]{0.5}
	\psfrag{F}[r][r][0.88]{0.6}
	\psfrag{G}[r][r][0.88]{0.7}
	\psfrag{H}[r][r][0.88]{0.8}
	\psfrag{J}[r][r][0.88]{0.9}
	\psfrag{W}[r][r][0.87]{MZI-mesh}
	\psfrag{Z}[r][r][0.87]{PSBS-butterfly}
	\psfrag{P}[l][l][0.85]{$n\!=\!16$}
	\psfrag{Q}[l][l][0.85]{$n\!=\!32$}
	\psfrag{R}[l][l][0.85]{$n\!=\!64$}
	\includegraphics[width=65mm]{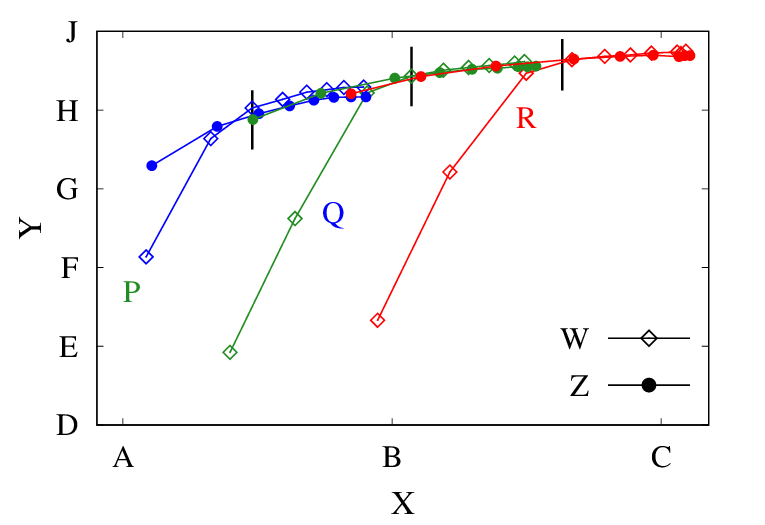}
\end{center}
\caption{Test accuracy as a function of parameter count on the Fashion-MNIST
classification task for the MZI-mesh and PSBS-butterfly (FGB) architectures,
used as the linear units in a three-layer perceptron model
with $n$ input ports ($n\!=\!16,32,64$).
}
\label{fig:fnn_testacc}
\vspace*{-3mm}
\end{figure}

We next consider a more practical FNN comprising a three-layer perceptron.
The first and second layers contain identical linear units implemented using
either the MZI-mesh or PSBS-butterfly (FGB) architecture
with $n$ input ports ($n\!=\!16,32,64$),
followed by a modReLU nonlinear activation function \cite{arjovsky}.
The final layer uses the same type of linear unit, followed by a nonlinear
function that measures optical power,
as illustrated in Fig.~\ref{app_fig:fnn_syst}.
The parameters are initialized and optimized 
using the same procedures as in the single-layer perceptron experiments.

Figure~\ref{fig:fnn_testacc} shows the test accuracy
for the PSBS-butterfly (FGB) and the MZI-mesh architectures,
evaluated for different values of $n$ and parameter counts.
The maximum parameter counts correspond to the case in which
each linear unit realizes a fully parameterized unitary matrix.
The vertical markers indicate the critical parameter counts for
$n\!=\!16,32,64$, below which the sequences of MZI-mesh linear units no longer
satisfy the reachability property.
In the three-layer perceptron, 
reachability corresponds to effective connectivity across the sequence of
linear units.\footnote{
In the butterfly topology, 
the reachability property is satisfied if each linear unit with $n$ input ports
contains at least one butterfly block with $\log_2 n$ stages. 
}
For aligned linear units implemented by an MZI-mesh with $n$ inputs and $m$ stages
per linear unit (with $n$ and $m$ even),
the reachability condition is given by
\begin{equation}
m \geq 2\left\lceil \frac{n-\!1}{2(\text{unit count})}\right\rceil\, ,
\label{eq:fnn_reachability}
\end{equation}
where the unit count is $3$ in this case.
For $n\!=\!16,32,64$, 
the corresponding critical parameter counts are 302, 1180, and 4286, respectively, 
including the parameters used in the modReLU functions.

The results show that the PSBS-butterfly and MZI-mesh architectures achieve
comparable test accuracy when reachability is preserved.
However, the MZI-mesh exhibits a sharp degradation in performance near the
critical parameter count.
These findings indicate that reachability across the sequence of linear units
is a key factor in the performance of feedforward NNs.
Consequently, the PSBS-butterfly (FGB) architecture provides 
high and stable performance relative to the MZI-mesh in this setting.

\subsection{Classification Using Recurrent Neural Networks}\label{ssec:rnn}
\begin{figure}[t]
\begin{center}
	\psfrag{X}[c][c][1.0]{Number of parameters (log scale)}
	\psfrag{Y}[c][c][1.0]{
		\begin{picture}(0,0)
			\put(0,0){\makebox(-5,28)[c]{
				Test accuracy}}
		\end{picture}
	}
	\psfrag{A}[c][c][0.88]{$10^3$}
	\psfrag{B}[c][c][0.88]{$10^4$}
	\psfrag{C}[c][c][0.88]{$10^5$}
	\psfrag{D}[r][r][0.88]{0.82} \psfrag{E}[r][r][0.88]{0.83}
	\psfrag{F}[r][r][0.88]{0.84} \psfrag{G}[r][r][0.88]{0.85}
	\psfrag{H}[r][r][0.88]{0.86} \psfrag{J}[r][r][0.88]{0.87}
	\psfrag{K}[r][r][0.88]{0.88} \psfrag{L}[r][r][0.88]{0.89}
	\psfrag{W}[r][r][0.82]{MZI-mesh} \psfrag{Z}[r][r][0.81]{PSBS-butterfly}
	\psfrag{P}[l][l][0.86]{$n\!=\!64$}
	\psfrag{Q}[r][r][0.86]{$n\!=\!128$}
	\psfrag{R}[r][r][0.86]{$n\!=\!256$}
	\psfrag{U}[r][r][0.86]{$\alpha$} \psfrag{V}[l][l][0.86]{$\beta$}
	\includegraphics[width=65mm]{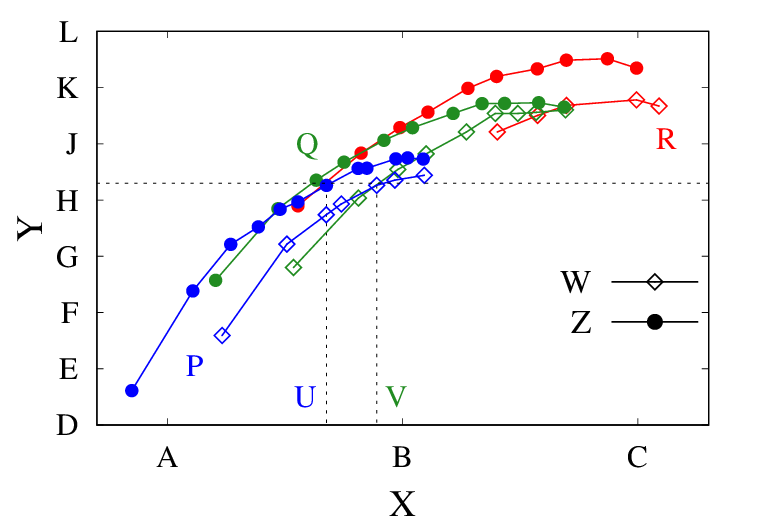}
\end{center}
\caption{Test accuracy of Elman-type RNNs using the MZI-mesh and
PSBS-butterfly (FGB) architectures as a function of parameter count for
$n\!=\!64,128,256$ on the Fashion-MNIST classification task.}
\label{fig:rnn_testacc}
\vspace*{-3mm}
\end{figure}

Recurrent neural networks (RNNs) can be difficult to train because of
vanishing or exploding gradients. To mitigate these difficulties, several RNN
models use unitary weight matrices
\cite{arjovsky,wisdom,maduranga,helfrich}.
Such RNNs are well suited to coherent ONNs, which inherently implement unitary
transformations. Through feedback loops, they can also exploit temporal modes
as additional degrees of freedom \cite{arahata}.
We evaluate the PSBS-butterfly (FGB) architecture in RNNs
by comparing it with the MZI-mesh architecture 
on the Fashion-MNIST classification task.
This setup allows us to examine how mixing capability and structural connectivity 
affect temporal information propagation.

The evaluation system consists of a preprocessing block and a main block.
The main block employs an Elman-type RNN, illustrated in Fig.~\ref{app_fig:rnn_syst}.
The RNN contains three linear units---input, hidden, and output---
each implemented using either the PSBS-butterfly or MZI-mesh architecture.
A modReLU nonlinear activation function is applied to the sum of 
the outputs from the input and hidden units.
The preprocessing block converts each Fashion-MNIST image into 
a sequence of $n$-dimensional complex-valued input vectors, 
where $n$ matches the number of input ports of the linear unit.
The sequence length $l_{(\mathrm{max})}$ is set to $36,18,9$ for
$n\!=\!64,128,256$, respectively. The outputs of the final linear unit are
converted to real-valued signals by photodetection.
Other processing steps follow those used in the FNN computational experiments.
After the $l_{(\mathrm{max})}$th input vector has propagated through the hidden
linear unit, a zero vector is appended to the input sequence so that the final hidden
state is mapped through the output unit. The loss is evaluated from the 
resulting output.

Figure~\ref{fig:rnn_testacc} shows that 
the PSBS-butterfly consistently achieves higher test accuracy than the MZI-mesh
across all values of $n$ and over a wide range of parameter counts.
For both architectures, the $n\!=\!256$ results deviate from the general trend
of improving performance with parameter count because overfitting occurs within
the evaluated range.
For the PSBS-butterfly,
the parameter count varies from $(n/2)\log_2 n$, corresponding to a single
butterfly block, to approximately $n^2$ per linear unit.
The parameter counts are controlled in the same manner as in the FNN computational experiments.
The parameters, which are shared across time steps, are optimized 
using the Adam optimizer in PyTorch.
Each reported test accuracy is averaged over 20 independent runs with different
initializations and evaluated after 200 training epochs.

\begin{figure}[t]
\begin{center}
	\psfrag{X}[c][c][.95]{Epoch}
	\psfrag{Y}[c][c][.95]{
		\begin{picture}(0,0)
			\put(0,0){\makebox(-5,20)[c]{
				Average loss (log scale)}}
		\end{picture}
	}
	\psfrag{A}[c][c][0.88]{$0$} \psfrag{B}[c][c][0.88]{$50$}
	\psfrag{C}[c][c][0.88]{$100$} \psfrag{D}[c][c][0.88]{$150$} 
	\psfrag{E}[c][c][0.88]{$200$}
	\psfrag{F}[r][r][0.87]{0.1} \psfrag{G}[r][r][0.87]{0.2}
	\psfrag{H}[r][r][0.87]{0.3} \psfrag{J}[r][r][0.87]{0.4}
	\psfrag{K}[r][r][0.87]{0.5} \psfrag{L}[r][r][0.87]{0.7}
	\psfrag{M}[r][r][0.87]{1.0} 
	\psfrag{U}[r][r][0.82]{mesh ({\small MZI-mesh})} 
	\psfrag{T}[r][r][0.82]{FGB ({\small PSBS-butterfly})}
	\psfrag{W}[r][r][0.82]{$n\!=\!64$}
	\psfrag{S}[r][r][0.82]{mesh} \psfrag{R}[r][r][0.82]{FGB}
	\psfrag{V}[l][l][0.82]{$n\!=\!128$}
	\psfrag{Q}[r][r][0.82]{mesh} \psfrag{P}[r][r][0.82]{FGB}
	\psfrag{O}[r][r][0.82]{$n\!=\!256$}
	\includegraphics[width=62mm]{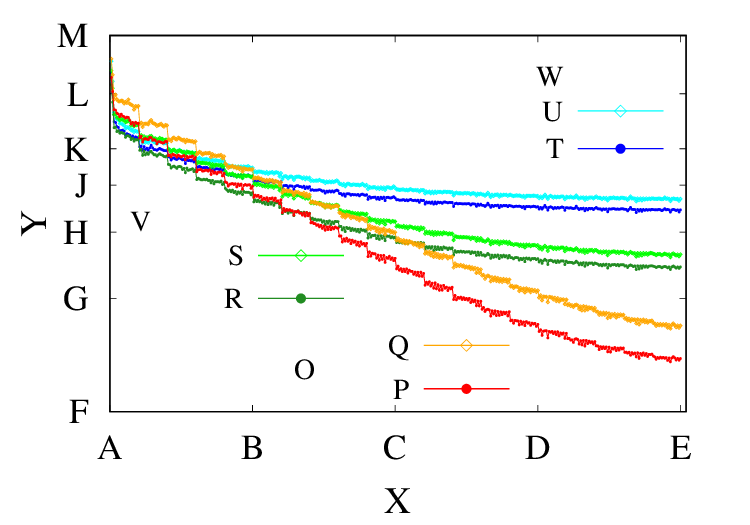}
\end{center}
\vspace*{-3mm}
\caption{Average training loss over 20 runs as a function of epoch for RNNs
using the PSBS-butterfly and MZI-mesh architectures with
$n\!=\!64,128,256$ on the Fashion-MNIST task in Fig.~\ref{fig:rnn_testacc}.
The number of parameters per linear unit is set to approximately $n^2/2$.
The PSBS-butterfly reaches lower loss earlier for $n\!=\!64$ and $128$ and attains
lower final loss in all settings.
}
\label{fig:avgloss}
\end{figure}

Figure~\ref{fig:avgloss} shows the average training-loss curves over 20
independent runs as a function of epoch
for the PSBS-butterfly (FGB) and the MZI-mesh architectures 
with $n$ input ports ($n\!=\!64,128,256$),
where the parameter count per linear unit is set to approximately $n^2/2$.
In contrast to the results in Fig.~\ref{fig:rnn_testacc},
which report the final test accuracy,
these curves reveal the training dynamics of the architectures.
The PSBS-butterfly reaches lower loss earlier for $n\!=\!64$ and $128$ and
attains lower final loss than the MZI-mesh in all settings.
Together with the test accuracy results, 
these curves suggest that the structural properties of the FGB architecture 
support more effective optimization in parameter-constrained settings.

\section{Discussion}\label{sec:disc}
We discuss three topics in light of the computational results:
the two key architectural requirements of statistical uniformity and strong
mixing capability per parameter;
reachability in feedforward NNs with multiple linear units and in recurrent NNs;
and the advantages and remaining challenges of the FGB architecture for
physical implementation.

\subsection{Two Key Requirements and Reachability}\label{ssec:req}
We investigated the design of parameter-efficient coherent ONNs under
practical constraints on the number of tunable components.
Our results show that high expressive power can be achieved in
parameter-constrained settings when the underlying architecture satisfies two
key requirements: statistical uniformity in the implemented unitary
transformations and strong mixing capability per parameter.
The proposed FGB architecture satisfies these requirements through the
combination of a butterfly topology and minimal PSBS building blocks.
The computational results from the matrix-reconstruction and classification
tasks using both FNNs and RNNs consistently demonstrate that, in
parameter-constrained settings, performance is determined not only by the
number of parameters but also by their architectural arrangement.
Reachability is a necessary, but not sufficient, condition for satisfying both
architectural requirements.
When a linear unit lacks reachability, some entries of its implemented unitary
transformation are structurally constrained to zero; these entries are the
\emph{trivial zeros} defined above.
Consequently, the resulting transformation cannot exhibit statistical
uniformity, and the unit cannot mix input signals across all modes.

\subsection{Reachability in Feedforward and Recurrent NNs}
\label{ssec:reach_fnn_rnn}
We first consider FNNs composed of multiple linear units.
Even when the individual units are constructed from the same architecture with
the same number of stages and do not themselves satisfy reachability, their
composition can restore end-to-end, or global, reachability.
For example, when the linear units are implemented using truncated MZI-mesh
architectures, global reachability is recovered if the concatenated units
contain at least $n$ stages and the two stage types occur in the required
alternating order, as shown in Fig.~\ref{fig:mesh}(b).
When considered as a single composite linear transformation, the concatenated
stages may also be initialized using the Haar initialization method, thereby
providing statistical uniformity across the composite transformation.
By contrast, a complete butterfly block provides reachability within each FGB
linear unit using only $\log_{2}n$ stages.
Composition of multiple linear units is therefore unnecessary for recovering
reachability unless a butterfly block is deliberately partitioned among those
units.

We next consider the RNN results.
As shown in Fig.~\ref{fig:rnn_testacc}, the test accuracy of the MZI-mesh RNN
does not degrade sharply at low parameter counts, in contrast to the behavior
of the FNN in Fig.~\ref{fig:fnn_testacc}.
One reason for this difference is that temporal recurrence mitigates the
effects of limited reachability within an individual linear unit.
To explain this mitigation, we introduce the concepts of \emph{partial
reachability} and \emph{complete reachability}.
Consider a sequence of $n$-dimensional input vectors of length
$l_{(\mathrm{max})}$ applied to an RNN comprising three linear units, each
implemented using an MZI-mesh architecture with $m$ stages.
The RNN satisfies partial reachability for the $l$th input vector
($l\!=\!1,2,\ldots,l_{(\mathrm{max})}$) if information from that vector can
propagate to every output terminal of the output unit by the time the loss is
evaluated.
The RNN satisfies complete reachability if partial reachability holds for every
input vector in the sequence.
Because the last input vector undergoes the fewest recurrent transformations,
complete reachability is equivalent to reachability of the last vector,
$l\!=\!l_{(\mathrm{max})}$.

Given $l$, $l_{(\mathrm{max})}$, and even $n$, the reachability condition with
respect to even $m$ is
\begin{equation}
m \geq 2\left\lceil
\frac{n-\!1}{2\left\{2+\left(l_{(\mathrm{max})}-\!l+\!1\right)\right\}}
\right\rceil\, ,
\label{eq:rnn_reachability_each}
\end{equation}
where $l_{(\mathrm{max})}-l+1$ is the number of times that information from the
$l$th input vector passes through the hidden unit, which we refer to as the
loop count.
The constant 2 inside the braces accounts for propagation through the input
and output units.
Unlike Eq.~(\ref{eq:fnn_reachability}), whose denominator contains only the
spatial contribution represented by the number of linear units,
Eq.~(\ref{eq:rnn_reachability_each}) also contains a temporal contribution
through the loop count.
For $l\!=\!l_{(\mathrm{max})}$, the loop count is one, and the complete
reachability condition reduces to Eq.~(\ref{eq:fnn_reachability}) with a unit
count of three, corresponding to the input, hidden, and output units.

Conversely, given $m$, $l_{(\mathrm{max})}$, and $n$, with both $n$ and $m$
even, Eq.~(\ref{eq:rnn_reachability_each}) can be expressed in terms of $l$ as
\begin{equation}
l \leq \left\lfloor
l_{(\mathrm{max})}+3-\frac{n-1}{m}
\right\rfloor\, .
\label{eq:rnn_reachability_lth}
\end{equation}
Thus, the $l$th input vector is reachable whenever it satisfies
Eq.~(\ref{eq:rnn_reachability_lth}).
Let $m_{(\mathrm{cpl})}$ denote the minimum stage count required for complete
reachability.
Even when $m<m_{(\mathrm{cpl})}$, a subset of the input vectors remains
reachable, depending on their positions in the sequence.
Because the upper bound in Eq.~(\ref{eq:rnn_reachability_lth}) changes
stepwise as a function of $1/m$, a substantial prefix of the sequence can
remain reachable even when the individual linear units contain far fewer than
$m_{(\mathrm{cpl})}$ stages.
For example, for $n\!=\!64$ and $l_{(\mathrm{max})}\!=\!36$, the minimum stage
count required for complete reachability is
$m_{(\mathrm{cpl})}\!=\!22$.
Even when $m\!=\!8$, corresponding to approximately
$0.36m_{(\mathrm{cpl})}$, the first 31 input vectors---approximately
$0.86l_{(\mathrm{max})}$---remain reachable.
This partial reachability mitigates the effects of limited connectivity and
helps explain why the MZI-mesh RNN does not exhibit the sharp performance
degradation observed in the FNN.
Thus, temporal recurrence can partially compensate for limited connectivity
within individual linear units.

A key advantage of the FGB architecture in RNNs is that a standard FGB linear
unit contains a complete butterfly block and therefore satisfies reachability
using only $\log_2 n$ stages.
Consequently, every input vector satisfies reachability regardless of its
position in the sequence, and the distinction between partial and complete
reachability need not be considered.
Moreover, the RNN does not rely on temporal recurrence to compensate for
limited connectivity within its linear units.
Partial-reachability analysis becomes relevant to the FGB architecture only in
the optional, nonstandard case in which a butterfly block is deliberately
partitioned into subsets of stages; it is not required for the complete FGB
blocks studied here.

\subsection{Toward Physical Implementation of the FGB Architecture}
\label{ssec:impl}
Although our numerical studies assume ideal devices,
the performance advantage of the PSBS-butterfly (FGB) over the MZI-mesh
architecture also has important implications for physical implementation.
At comparable test accuracies in the FNN and RNN computational experiments
shown in Figs.~\ref{fig:1unit_fnn_testacc}--\ref{fig:rnn_testacc}, 
the FGB architecture consistently requires fewer tunable parameters than 
the MZI-mesh architecture.
For a given target accuracy, this reduction in parameter count also translates
into fewer stages and a smaller optical depth.
For example, in the RNN results with $n\!=\!64$ in
Fig.~\ref{fig:rnn_testacc}, the PSBS-butterfly (FGB) and MZI-mesh
architectures require 4736 parameters (point~$\alpha$) and 7752 parameters
(point~$\beta$), respectively, to achieve an average test accuracy of $0.863$.
These parameter counts correspond to 48 and 80 fine-grained stages,
respectively, in each linear unit.
For this comparison, each MZI stage is decomposed into and counted as two
successive PSBS-equivalent stages; the fine-grained stage count thus equals the
number of beam-splitter stages.
The reduced parameter and stage counts of the FGB architecture
offer three potential implementation advantages:
\begin{enumerate}[(i)]
\item \emph{Reduced latency.}
Optical propagation delay increases approximately linearly with physical path
length, which generally increases with the number of stages \cite{clements}.
A smaller optical depth can therefore reduce processing latency, provided that
the lengths of the interstage connections are appropriately controlled.
\item \emph{Reduced optical loss and laser-power requirement.}
In architectures composed of phase shifters (PSs) and beam splitters (BSs),
component insertion losses accumulate approximately linearly in decibels with
optical depth.
Consequently, optical transmission decreases exponentially with the number of
stages, and the laser power required to maintain a fixed received power
increases accordingly \cite{giam,shafiee}.
By reducing the number of mixing stages, the FGB architecture can reduce the
stage-accumulated component loss and the associated laser-power requirement,
particularly as the network size increases, provided that additional routing
loss does not offset these savings.
\item \emph{Reduced weight-programming energy and control overhead.}
Fewer tunable parameters reduce both the number of weights that must be stored
and the volume of control data transferred to program the optical circuit.
The FGB architecture can therefore reduce the energy and hardware overhead
associated with weight storage, loading, and programming.
This benefit can be particularly important during training, when the weights
are updated repeatedly, although its magnitude depends on the particular
device and control implementation.
\end{enumerate}

The reductions in parameter and stage counts arise from the
combination of the minimal PSBS building block and the butterfly topology.
The latter, however, employs longer interstage connections that may require
many waveguide crossings in an on-chip implementation.
These crossings introduce additional optical loss and phase errors.
Moreover, path-dependent crossing counts and route lengths can produce
nonuniform loss and optical path lengths among input--output pairs, thereby
complicating path-length matching and phase control.
These challenges may be mitigated using existing technologies, including
ultralow-loss waveguide crossings \cite{liu_crossing,wu_crossing}, multilayer
waveguide routing based on three-dimensional integration \cite{xia}, and dummy
crossings that equalize crossing counts among optical paths.
From a topological perspective, architectures isomorphic to the standard
butterfly topology may also reduce the number of crossings and the lengths of
interstage connections while preserving the relevant connectivity properties
\cite{wu}.
Thus, the FGB architecture offers a promising route toward parameter-efficient
physical implementation, although its practical advantage will depend on the
balance between the reduced optical depth and the additional routing overhead.

The FGB architecture also offers a potential advantage in parameter
initialization.
Under the ideal-device assumptions considered here, independently sampling its
phase parameters from a uniform distribution produces the entrywise statistical
uniformity examined in this work.
The FGB architecture therefore does not require a specialized procedure such as
the Haar initialization method used for MZI-based meshes to obtain this
first-moment property \cite{pai}.
Avoiding such a procedure may simplify physical configuration, particularly
when device imperfections and crosstalk hinder the precise realization of
prescribed MZI settings.
Nevertheless, this initialization advantage should be evaluated under realistic
component imperfections.

The proposed FGB architecture also has two principal limitations:
\begin{enumerate}[(a)]
\item
The current configuration restricts the number of input ports to a power of
two, $n\!=\!2^t$ with $t\!\in\!\mathbb{N}$; for example,
$n\!=\!16,32,64,128,\ldots$.
Because the gap between successive supported input-port counts increases with
$n$, an implementation may need either to restrict the input dimension to a
smaller supported value or to allocate unused ports at the next larger
supported value.
An important direction for future work is therefore to extend the architecture
to arbitrary numbers of input ports while preserving its underlying design
principles.
\item
No general decomposition algorithm is currently available for the FGB
architecture.
Consequently, it has not been established whether or how a given FGB
configuration can realize an arbitrary unitary transformation, even when its
parameter count exceeds the number of degrees of freedom of the unitary group.
A constructive decomposition framework would identify sufficient
configurations, provide deterministic mappings from target unitary matrices to
device parameters, and establish corresponding universality guarantees.
\end{enumerate}

Addressing these limitations, together with reducing the physical overhead
associated with crossings and long interstage connections, will be essential
for scalable photonic integration.
More broadly, the architectural principles identified in this
work---statistical uniformity and strong mixing capability per parameter---may
apply to a wider class of optical and hybrid computing systems.
Evaluating these principles under realistic conditions, including optical
loss, phase noise, crosstalk, and fabrication variability, will be important
for bridging the gap between theoretical architectural design and practical
deployment.
These efforts will help establish a principled framework for designing
parameter-efficient ONNs under realistic physical constraints.

\section{Conclusions}\label{sec:conc}
We identified two key architectural requirements for linear units
in parameter-constrained settings:
statistical uniformity in the implemented transformations
and strong mixing capability per parameter.
Through computational experiments,
we demonstrated that the proposed fine-grained butterfly (FGB) architecture
achieves lower reconstruction error and higher classification accuracy than
conventional rectangular mesh architectures based on Mach-Zehnder
interferometers (MZIs) at restricted parameter counts.
This advantage arises from the combination of a
butterfly topology and minimal phase-shifter--beam-splitter (PSBS) building blocks,
which together realize parameterized unitary transformations 
satisfying the proposed architectural principles.

Several directions for future work remain.
From a practical perspective,
addressing the physical implementation challenges of the FGB architecture
is essential for scalable deployment.
From a theoretical standpoint,
a deeper understanding of how parameterization influences expressive power 
remains an open problem.
Furthermore, developing a principled framework for identifying 
optimal parameterizations under physical constraints, 
given a fixed parameter budget, is an important direction for future research.

These results show that architectures that jointly provide statistical
uniformity and strong mixing capability per parameter can substantially improve
performance in parameter-constrained ONNs.
These principles provide a foundation for a general design framework under
realistic physical constraints
and may guide the development of future photonic computing systems.

\begin{acknowledgments}
The authors thank Shuji Horinaga for helpful discussions on the mathematical
aspects of this work.
\end{acknowledgments}

\newpage
\appendix
\section{Matrix Reconstruction Task}\label{app_sec:reconst}
The matrix-reconstruction task in Section~\ref{ssec:reconst} uses the objective
function in Eq.~(\ref{eq:objfunct}):
\begin{align*}
L_F^{*}(T,U)
&= \frac{1}{\sqrt{2n}}\|T-U\|_F \\
&= \frac{1}{\sqrt{2n}}
\left\{\operatorname{Tr}\!\left[(T-U)^{\dagger}(T-U)\right]\right\}^{1/2}\, ,
\end{align*}
where $n$ denotes the dimensionality of the parameterized unitary matrix 
$T\!\in\!\mathbb{C}^{n\times n}$.
The expression inside braces on the right-hand side can be rewritten as
\begin{equation}
\operatorname{Tr}\!\left[(T-U)^{\dagger}(T-U)\right] = 2n\left\{
1-\frac{1}{n}\operatorname{Re}\!\left[\operatorname{Tr}(T^{\dagger}U)\right]\right\} \, ,
\label{app_eq:trace}
\end{equation}
where $\operatorname{Re}(z)$ denotes the real part of $z\!\in\!\mathbb{C}$.
The objective function $L_F^{*}(T,U)$ can therefore be rewritten as
\begin{equation}
L_F^{*}(T,U) = \left\{
1-\frac{1}{n}\operatorname{Re}\!\left[\operatorname{Tr}(T^{\dagger}U)\right]\right\}^{1/2} \, ,
\label{app_eq:mat_inner_prod}
\end{equation}
where $\operatorname{Re}[\operatorname{Tr}(T^{\dagger}U)]$ is the real
Hilbert--Schmidt (Frobenius)
inner product of matrices $T$ and $U$ \cite{meckes}.

Moreover, $\operatorname{Re}[\operatorname{Tr}(T^{\dagger}U)]$ can be
expressed using the corresponding vectorizations as follows:
\begin{align}
\operatorname{Re}\!\left[\operatorname{Tr}(T^{\dagger}U)\right]
&= \operatorname{Re}\!\left[\operatorname{vec}(T^{*})\cdot
\operatorname{vec}(U)\right] \nonumber\\
&\coloneqq \|\operatorname{vec}(T^{*})\|_2\,
\|\operatorname{vec}(U)\|_2 \cos\theta \nonumber\\
&= n\cos\theta\,,
\label{app_eq:cossim}
\end{align}
where $\operatorname{vec}(U)$ denotes the vectorization of $U$ in
$\mathbb{C}^{n^2}$,
$T^{*}$ is the conjugate of $T$,
and $\theta\!\in\![0,\pi]$ is the angle between the corresponding real
representations of $T$ and $U$.
Substituting Eq.~(\ref{app_eq:cossim}) into Eq.~(\ref{app_eq:mat_inner_prod})
simplifies $L_F^{*}$ to
\begin{equation}
L_F^{*}(T,U) = \frac{1}{\sqrt{2n}}\| T-U\|_F = \sqrt{1-\cos\theta}\, ,
\label{app_eq:objfunct_cossim}
\end{equation}
where $L_F^{*}(T,U)\!\in\![0,\sqrt{2}]$.
For $T\!=\!U$, $\theta\!=\!0$ and $L_F^{*}\!=\!0$; for $T\!=\!-U$,
$\theta\!=\!\pi$ and $L_F^{*}\!=\!\sqrt{2}$.
$L_F^{*}(T,U)\!=\!1$ if and only if $\theta\!=\!\pi/2$.
In Figs.~\ref{fig:reconst_all} and \ref{fig:reconst_n16}, $L_F^{*}(T,U)$
approaches 1 in the low-parameter limit. This value is expected when $T$ is
effectively uncorrelated with the target $U$.
This relationship is analogous to that between Euclidean distance and cosine
similarity for two real-valued vectors on a unit hypersphere.

Alternatively, $\theta(T,U)$ itself can serve as the objective function:
\begin{align}
\theta(T,U) &= \arccos\!\left(1-[L_F^{*}(T,U)]^2\right)\,.
\label{app_eq:theta}
\end{align}
When $T$ and $U$ are far apart, this measure emphasizes their dissimilarity
more strongly in the objective-function value.

\section{Geometric Interpretation of Cosine Similarity}\label{app_sec:cossim}
Given two $n\times n$ unitary matrices $A$ and $B$,
their real Hilbert--Schmidt (Euclidean Frobenius) inner product
$\langle A,B\rangle$
is defined by
\begin{equation}
\langle A,B\rangle := \text{Re}(\text{Tr}(A^{\dagger}B))\,.
\label{app_eq:HS}
\end{equation}
Let $\bm{a}\!=\!\text{vec}(A)$ and $\bm{b}\!=\!\text{vec}(B)$ denote
their flattened $k$-dimensional real-vector representations. Decomposing each
complex entry into its real and imaginary components gives
\begin{align}
\bm{a} = &(\text{Re}(A_{11}),\text{Re}(A_{12}),\ldots,\text{Re}(A_{nn}),\nonumber\\
&\, \text{Im}(A_{11}),\text{Im}(A_{12}),\ldots,\text{Im}(A_{nn}))^\intercal\, ,
\end{align}
where $A_{jh}$ is the entry in row $j$ and column $h$ of $A$, and
$\operatorname{Re}(A_{jh})$ and $\operatorname{Im}(A_{jh})$ denote, respectively,
the real and imaginary components of $A_{jh}$.
Vector $\bm{b}$ is defined analogously. Then, $k\!=\!2n^2$ and
$\|\bm{a}\|_2\!=\!\|\bm{b}\|_2\!=\!\sqrt{n}$, so both vectors lie on the
surface of a $2n^2$-dimensional ball of radius $r\!=\!\sqrt{n}$, that is, on a
$(2n^2\!-\!1)$-dimensional hypersphere.

\begin{figure}[t]
\begin{center}
	\psfrag{A}[c][c][1.0]{$\bm{a}$} \psfrag{B}[c][c][1.0]{$\bm{b}$}
	\psfrag{P}[c][c][1.0]{$r\sin\theta$}
	\psfrag{Q}[c][c][1.0]{$\theta$}
	\psfrag{R}[c][c][1.0]{$r\!=\!\sqrt{n}$}
	\includegraphics[width=35mm]{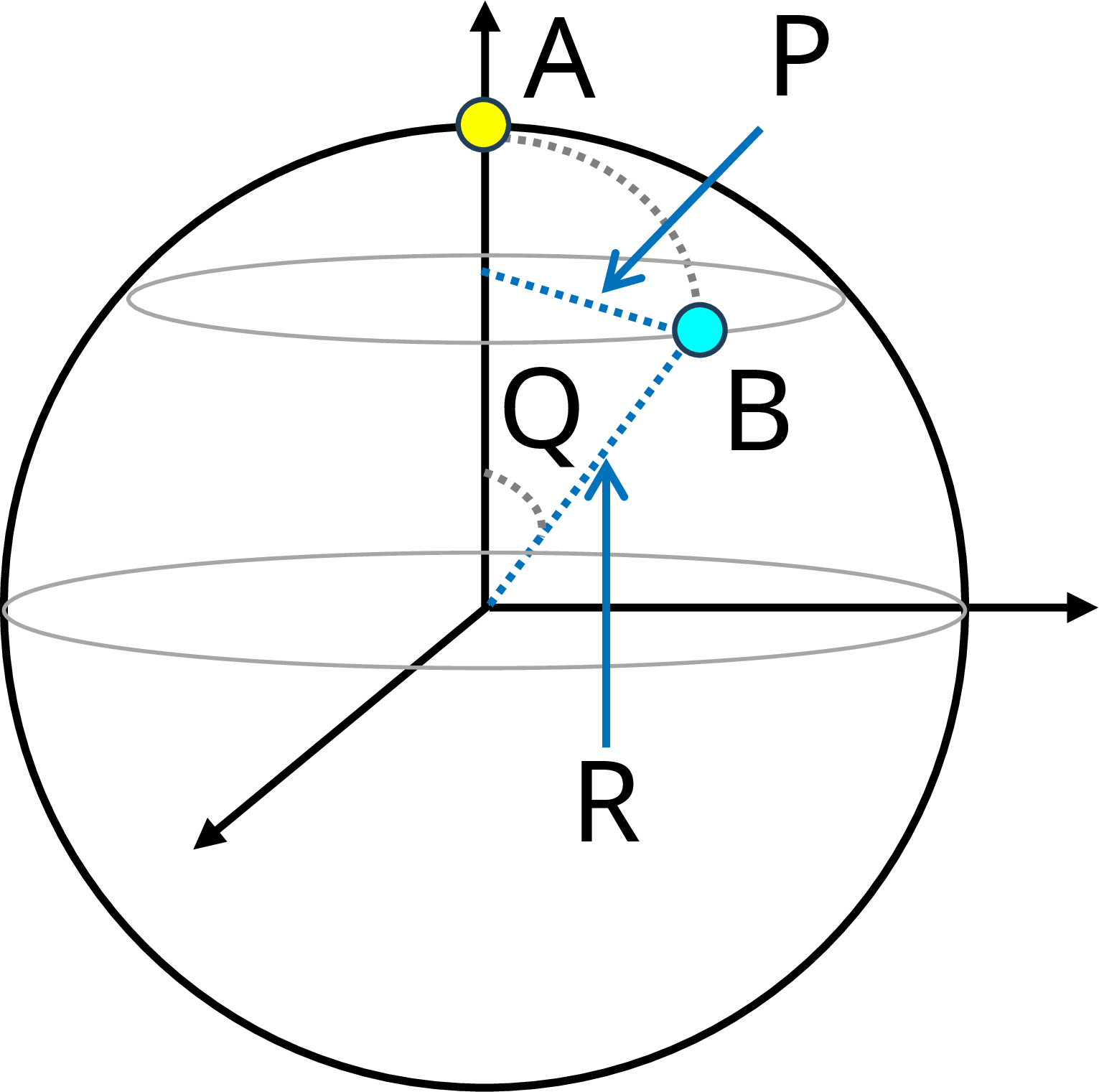}
\end{center}
\caption{Conceptual diagram of vectors $\bm{a}$ and $\bm{b}$ on
the hypersphere, which have an angular separation of $\theta$.
}\label{app_fig:sphere}
\end{figure}

Without loss of generality,
we can assume that vector $\bm{a}$ is located at the north pole 
of the hypersphere, 
while vector $\bm{b}$ is situated at colatitude angle $\theta$ from $\bm{a}$,
as shown in Fig.~\ref{app_fig:sphere}.
Equation~(\ref{app_eq:HS}) can be written in terms of $\bm{a}$ and $\bm{b}$ as
\begin{equation}
\text{Re}(\text{Tr}(A^{\dagger}B)) = \bm{a}^\intercal \bm{b} = n\cos\theta \, .
\label{app_eq:cossim_ab}
\end{equation}

\begin{proposition}\label{app_prop:pdf}
Assume that vector $\bm{b}$ is uniformly distributed on the surface of a
$k$-dimensional ball, that is, on a $(k\!-\!1)$-dimensional hypersphere.
For a given $k$, the probability density function (pdf) of $\theta$ is
\begin{align}
\text{pdf}(\theta;k) &= 
\frac{1}{\sqrt{\pi}}\, \frac{\Gamma(k/2)}{\Gamma((k\!-\!1)/2)}\,
(\sin\theta)^{k-2}, \label{app_eq:pdf} \\
\int_{0}^{\pi}\text{pdf}(\theta;k)\,d\theta &= 1 
\label{app_eq:prob1}\, ,
\end{align}
where $\Gamma(\ast)$ denotes the Gamma function.
\end{proposition}

\begin{proof}
The surface area of the $(k\!-\!1)$-dimensional hypersphere of radius $r$ is
\begin{equation}
\text{Area}(r;k) = \frac{2\pi^{k/2}}{\Gamma(k/2)}\, r^{k-1} \, .
\label{app_eq:area}
\end{equation}
When a hypersphere is divided by a hyperplane at colatitude $\theta$, the
portion with $\theta\!<\!\pi/2$ is a smaller hyperspherical cap \cite{li}.
The surface area of the hyperspherical cap is obtained by integrating its
infinitesimal area elements:
\begin{equation}
\text{Area}_{(cap)}(\theta, r ;k) = \frac{2\pi^{(k-1)/2}}{\Gamma((k-1)/2)} r^{k-1}
\int_{0}^{\theta}(\sin\phi)^{k-2}d\phi \, .
\label{app_eq:area_cap}
\end{equation}
The derivative of this area with respect to $\theta$ is given by
\begin{equation}
\frac{d\text{Area}_{(cap)}}{d\theta} = \frac{2\pi^{(k-1)/2}}{\Gamma((k-1)/2)} r^{k-1}
(\sin\theta)^{k-2}\, .
\label{app_eq:deriv_area}
\end{equation}
If $\bm{b}$ is uniformly distributed on the hypersphere, the density at angle
$\theta$ is
the ratio of the derivative of the hyperspherical cap to the surface area:
\begin{equation}
\text{pdf}(\theta;k) =\! \frac{d\text{Area}_{(cap)}/d\theta}{\text{Area}}
=\! \frac{1}{\sqrt{\pi}}\, \frac{\Gamma(k/2)}{\Gamma((k\!-\!1)/2)}\,
(\sin\theta)^{k-2} .
\end{equation}
This is consistent with the formula in Eq.~(\ref{app_eq:pdf}).

We next consider the value of the integral in Eq.~(\ref{app_eq:prob1}).
\begin{equation}
\int_{0}^{\pi}\text{pdf}(\theta;k) d\theta 
= \frac{2}{\sqrt{\pi}}\, \frac{\Gamma(k/2)}{\Gamma((k\!-\!1)/2)}
\int_{0}^{\pi/2} (\sin\theta)^{k-2}d\theta\, .
\label{app_eq:integral_pdf}
\end{equation}
By using the Beta and Gamma functions, the integral in Eq.~(\ref{app_eq:integral_pdf})
is represented as 
\begin{equation}
2\int_{0}^{\pi/2}\, (\sin\theta)^{k-2}d\theta
= \mathrm{B}\!\left(\frac{k-1}{2},\frac{1}{2}\right)\!=\frac{\Gamma((k-1)/2)\sqrt{\pi}}{\Gamma(k/2)}.
\label{app_eq:beta-gamma}
\end{equation}
Substituting Eq.~(\ref{app_eq:beta-gamma}) into Eq.~(\ref{app_eq:integral_pdf})
yields Eq.~(\ref{app_eq:prob1}).
\end{proof}

\begin{figure}[t]
\begin{center}
	\psfrag{X}[c][c][1.0]{Colatitude angle $\theta$ ($\times \pi$)}
	\psfrag{Y}[c][c][1.0]{Probability ($\times 10^{-2}$)}
	\psfrag{A}[c][c][0.9]{0.4} \psfrag{B}[c][c][0.9]{0.45}
	\psfrag{C}[c][c][0.9]{0.5} \psfrag{D}[c][c][0.9]{0.55}
	\psfrag{E}[c][c][0.9]{0.6}
	\psfrag{F}[c][c][0.9]{0} \psfrag{G}[c][c][0.9]{2}
	\psfrag{H}[c][c][0.9]{4} \psfrag{J}[c][c][0.9]{6}
	\psfrag{K}[c][c][0.9]{8} \psfrag{L}[c][c][0.9]{10}
	\psfrag{P}[r][r][0.8]{Haar 16}
	\psfrag{Q}[r][r][0.8]{$k\!=\!512$}
	\psfrag{R}[r][r][0.8]{Haar 4}
	\psfrag{S}[r][r][0.8]{$k\!=\!32$}
	\includegraphics[width=70mm]{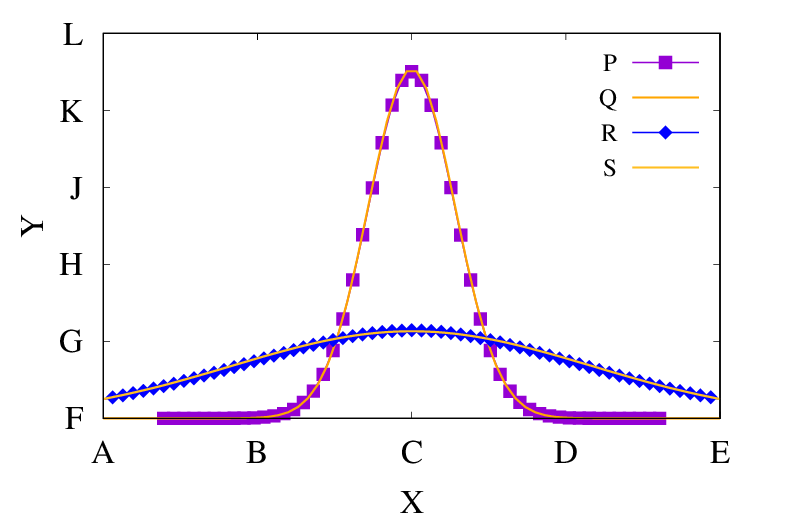}
\end{center}
\caption{Comparison of selection probabilities for vectorized Haar-random unitaries
and points uniformly distributed on a hypersphere at colatitude $\theta$.
}\label{app_fig:prob_angle}
\end{figure}

\begin{table}[t]
\centering
\caption{Two-sample Kolmogorov--Smirnov test. Here, n.s. denotes a result that
is not statistically significant at the 0.05 level.}
\label{app_table:kstest}
\renewcommand{\arraystretch}{1.4}
\begin{tabular}{cccc}\hline
$n$ & $2n^2$ & $p$ value & $D$ statistic (\%) $\downarrow$\\ \hline
16 & 512 & 0.764 (n.s.)& 0.666 \\
4 & 32 & 0.376 (n.s.)& 0.911 \\ \hline
\end{tabular}
\end{table}

We compare the angular distributions of vectorized Haar-random unitary matrices
and points uniformly distributed on the hypersphere.
The former is obtained from the cosine similarities between pairs of
$n\!\times\!n$ unitary matrices $A$ and $B$ sampled from the Haar measure
\cite{mezzadri}, whereas the latter is calculated using
$\text{pdf}(\theta;k\!=\!2n^2)$ in Eq.~(\ref{app_eq:pdf}).
Figure~\ref{app_fig:prob_angle} shows the comparison results.
The plots labeled ``Haar 16'' and ``Haar 4'' represent the probabilities for
the Haar-random matrices with $n\!=\!16$ and $4$, respectively,
obtained from the cosine similarities of $10^{8}$ pairs.
Similarly, the plots labeled with the dimensionality $k$ 
(e.g., ``$k\!=\!512$") 
correspond to the probabilities for points on the surface of the $k$-dimensional ball,
where $k\!=\!2n^2$.
The probability distributions for Haar-random unitary matrices show
excellent agreement with those for the points on the hypersphere surface.

We assess this agreement using a two-sample Kolmogorov--Smirnov (KS) test.
An empirical cumulative distribution function is constructed from the
colatitude angles derived from the cosine similarities between Haar-random
unitary matrices and compared with the cumulative distribution corresponding
to Eq.~(\ref{app_eq:pdf}).
For both matrix sizes, the KS test does not reject the null hypothesis that the
observed angles follow the analytical distribution ($p\!>\!0.05$).
The $p$ values and small $D$ statistics in Table~\ref{app_table:kstest}
therefore indicate no statistically significant discrepancy for the cases
tested.

Thus, for the matrix sizes tested, the angular distribution of vectorized
Haar-random unitary matrices is statistically consistent with that of points
uniformly distributed on the surface of the $2n^2$-dimensional ball.

\section{Path Counts in Mesh Architectures}\label{app_sec:path}
\subsection{Total Path Count to Each Output Terminal}
\label{app_ssec:app_path_total}
\begin{figure}[t]
\begin{center}
	\psfrag{a}[c][c][1.0]{$x_0$} \psfrag{b}[c][c][1.0]{$x_1$}
	\psfrag{c}[c][c][1.0]{$x_2$} \psfrag{d}[c][c][1.0]{$x_3$}
	\psfrag{e}[c][c][1.0]{$x_4$} \psfrag{f}[c][c][1.0]{$x_5$}
	\psfrag{g}[c][c][1.0]{$x_6$} \psfrag{h}[c][c][1.0]{$x_7$}
	\psfrag{A}[c][c][1.0]{$y_0$} \psfrag{B}[c][c][1.0]{$y_1$}
	\psfrag{C}[c][c][1.0]{$y_2$} \psfrag{D}[c][c][1.0]{$y_3$}
	\psfrag{E}[c][c][1.0]{$y_4$} \psfrag{F}[c][c][1.0]{$y_5$}
	\psfrag{G}[c][c][1.0]{$y_6$} \psfrag{H}[c][c][1.0]{$y_7$}
	\psfrag{X}[c][c][.9]{A} \psfrag{Y}[c][c][.9]{B}
	\psfrag{P}[c][c][.9]{Stage type}
	\includegraphics[width=50mm]{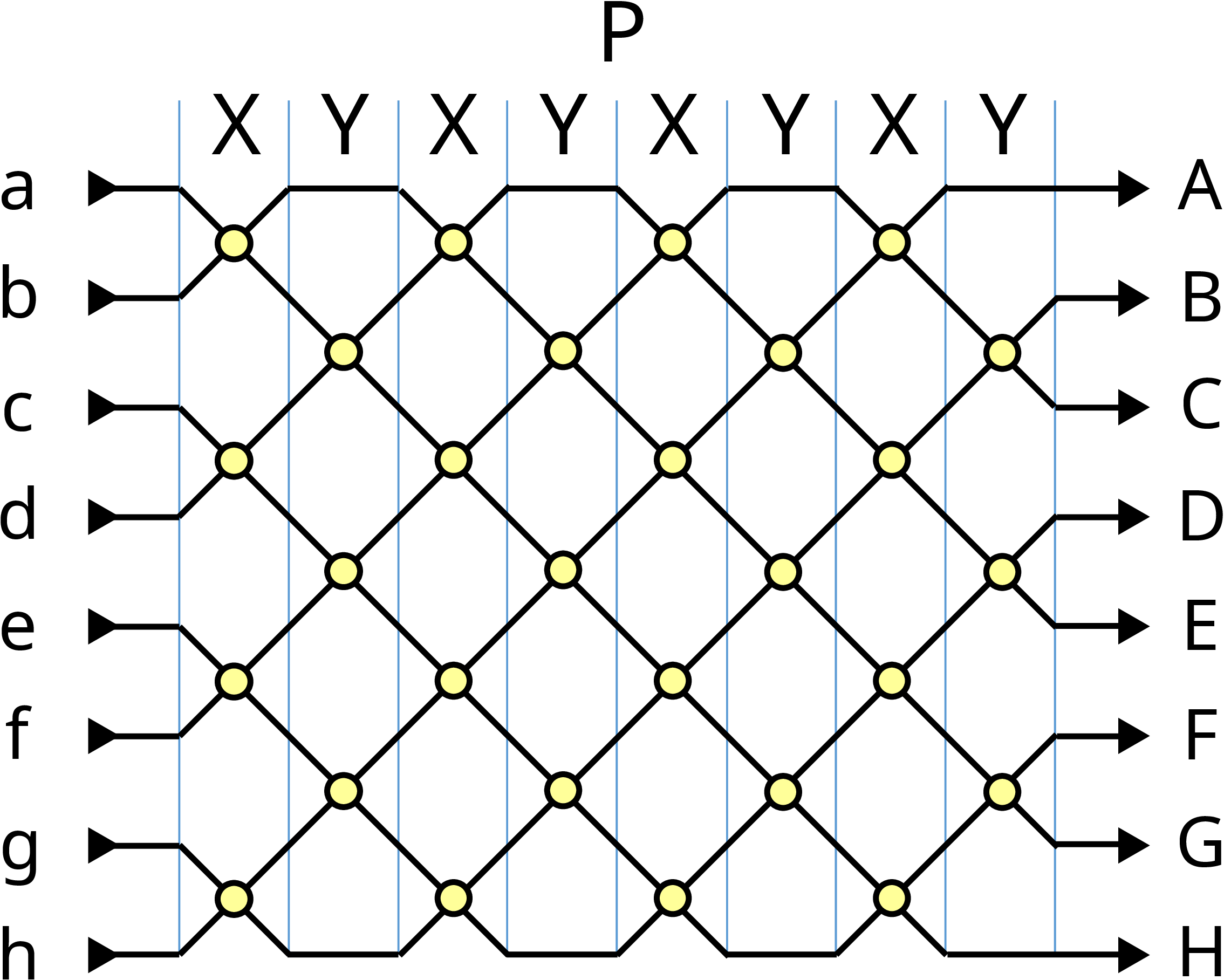}
\end{center}
\vspace*{-2mm}
\caption{Mesh architecture with eight input and eight output terminals.
}\label{app_fig:mesh}
\end{figure}

We analyze input-output path counts in a mesh architecture to quantify their
imbalance. Figure~\ref{app_fig:mesh} shows an example with eight inputs and
eight outputs ($n\!=\!8$) and eight stages ($m\!=\!8$).
As illustrated in this figure,
the mesh architecture alternates between two stage types, denoted A and B.
We represent the mesh as a graph in which MZI
terminals are vertices, waveguides are edges, and 
each MZI acts as a \emph{connector}
between each of its inputs and both of its outputs. 
Then, adjacency matrices representing type-A and type-B stages 
with $n$ inputs and $n$ outputs
are explicitly expressed as block-diagonal matrices, 
$M_{A}$ and $M_{B}$, as 
\begin{equation}
\hspace*{-5pt}
M_A =\! \left[\begin{smallmatrix}
1 & 1 & & & \\
1 & 1 & & & \\
 & & \ddots & & \\
 & &  & 1 & 1\\
 & &  & 1 & 1
\end{smallmatrix}\right]_{n\times n},
\end{equation}
\begin{equation}
M_B =\! \left[\begin{smallmatrix}
1& &  &      & &  &\\
 &1&1 &      & &  &\\
 &1&1 &      & &  &\\
 & &  &\ddots& &  &\\
 & &  &      &1&1 &\\
 & &  &      &1&1 &\\
 & &  &      & &  &1
\end{smallmatrix}\right]_{n\times n},
\end{equation}
where zero entries are left blank for clarity.
For simplicity, assume that the numbers of inputs and stages are \emph{even}.
The adjacency matrix representing the mesh, $\mathcal{M}$, is
\begin{align}
\mathcal{M} &= (M_B M_A)^{\widehat{m}},
\quad \widehat{m}=\frac{m}{2}\in \mathbb{N}\, .
\label{app_eq:mesh_matrix}
\end{align}
where $\widehat{m}$ is half the stage count $m$.

Using $\mathcal{M}$, we obtain the total number of paths reaching output
$y_j$, for $j\!=\!0,1,2,\ldots,n\!-\!1$, as
\begin{equation}
\bm{y} = \mathcal{M} \bm{1}_n\, ,
\label{app_eq:paths}
\end{equation}
where $\bm{1}_n$ is the $n$-dimensional all-ones input vector
$\bm{1}_n\!=\!(1,1,\dots,1)^\intercal$, and $\bm{y}$ is the vector whose
$j$th entry, $y_j$, is the total path count to the $j$th output.
For example, when Eq.~(\ref{app_eq:paths}) is applied to the mesh 
in Fig.~\ref{app_fig:mesh}, 
where $n\!=\!8$ and $\widehat{m}\!=\!m/2\!=4$, 
$\bm{y}$ is shown as 
\begin{equation}
\bm{y} = (70,180,180,220,220,180,180,70)^\intercal\, .
\label{app_eq:y_8}
\end{equation}

\begin{figure}[t]
\begin{center}
	\psfrag{X}[c][c][.95]{Normalized output index}
	\psfrag{Y}[c][c][0.9]{
		\begin{picture}(0,0)
			\put(0,0){\makebox(-6,38)[c]{
				Normalized path count}}
			\put(-15,2){\makebox{($\times 10^{-2}$)}}
		\end{picture}
	}
	\psfrag{A}[c][c][.88]{0} \psfrag{B}[c][c][.88]{0.2}
	\psfrag{C}[c][c][.88]{0.4} \psfrag{D}[c][c][.88]{0.6}
	\psfrag{E}[c][c][.88]{0.8} \psfrag{F}[c][c][.88]{1.0}
	\psfrag{G}[r][r][.88]{0} \psfrag{H}[r][r][.88]{2}
	\psfrag{J}[r][r][.88]{4} \psfrag{K}[r][r][.88]{6}
	\psfrag{L}[r][r][.88]{8} \psfrag{M}[r][r][.88]{10}
	\psfrag{P}[r][r][.75]{$n$: 16} \psfrag{Q}[r][r][.75]{32} 
	\psfrag{R}[r][r][.75]{64}
	\includegraphics[width=60mm]{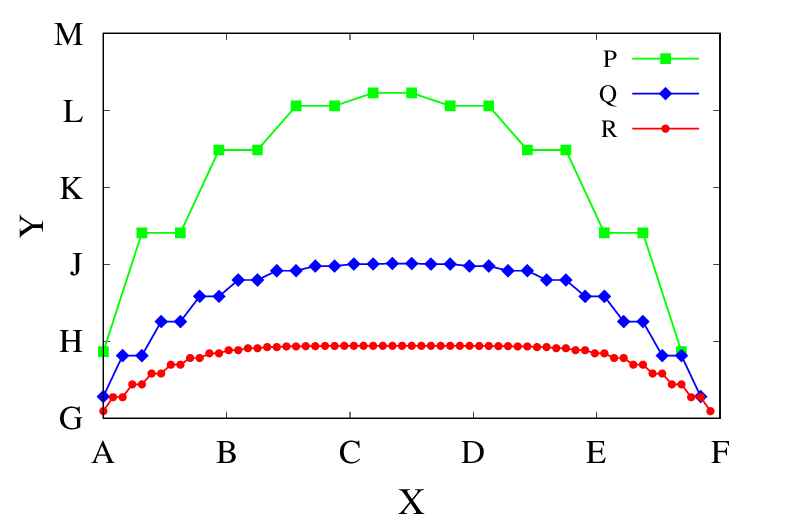}
\end{center}
\vspace*{-2mm}
\caption{Normalized path counts 
along normalized output index in the mesh architecture.
}\label{app_fig:pathprob}
\end{figure}

\begin{figure}[t]
\begin{center}
	\psfrag{X}[c][c][.95]{Number of inputs ($n$)}
	\psfrag{Y}[c][c][0.95]{
		\begin{picture}(0,0)
			\put(0,0){\makebox(-6,55)[c]{
				Number of paths}}
			\put(-25,12){\makebox{(log scale)}}
		\end{picture}
	}
	\psfrag{A}[c][c][.88]{0} \psfrag{B}[c][c][.88]{10}
	\psfrag{C}[c][c][.88]{20} \psfrag{D}[c][c][.88]{30}
	\psfrag{E}[c][c][.88]{40} \psfrag{F}[c][c][.88]{50}
	\psfrag{G}[c][c][.88]{60} \psfrag{H}[c][c][.88]{70}
	\psfrag{J}[r][r][.88]{$10^2$} \psfrag{K}[r][r][.88]{$10^6$}
	\psfrag{L}[r][r][.88]{$10^{10}$} \psfrag{M}[r][r][.88]{$10^{14}$}
	\psfrag{N}[r][r][.88]{$10^{18}$}
	\psfrag{P}[r][r][.78]{Max} \psfrag{Q}[r][r][.78]{Min} 
	\includegraphics[width=58mm]{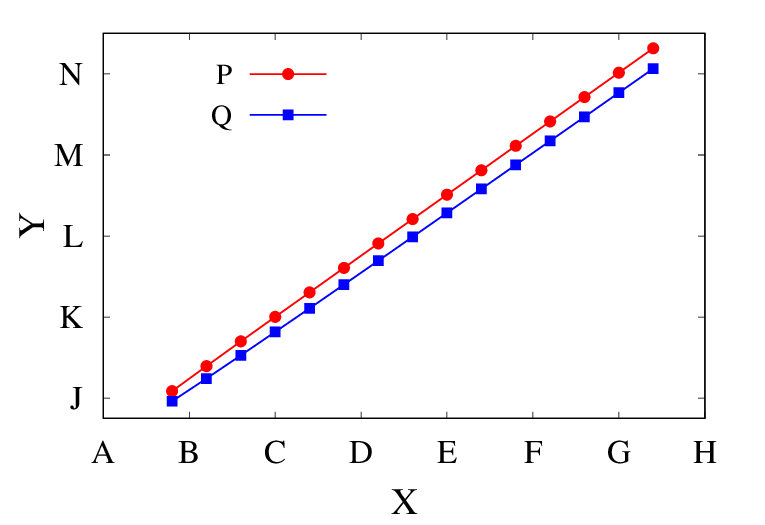}
\end{center}
\vspace*{-2mm}
\caption{Maximum and minimum path counts as functions of the number of inputs
$n$. Points on the upper and lower curves are $\max_j(y_j)$ and $\min_j(y_j)$,
respectively, with $n\!=\!m$.
}\label{app_fig:pathcount}
\end{figure}

Figure~\ref{app_fig:pathprob} shows the normalized path-count distribution
$y_j/\|\bm{y}\|_1$ as a function of normalized output index $j/n$ for
$n\!=\!m\!=\!16,32,64$. Paths are concentrated near the central output
terminals, and the width of this central region increases with $n$.
Figure~\ref{app_fig:pathcount} shows the maximum and minimum path counts, i.e.,
$\max_j(y_j)$ and $\min_j(y_j)$, as functions of $n\!=\!m$. Both quantities
grow approximately exponentially over the evaluated range and therefore appear
nearly linear on the logarithmic scale. The maximum has a slightly larger
slope, so the ratio of the maximum to the minimum increases with $n$.

\subsection{Path Count from Each Input Terminal}
\label{app_ssec:app_path_indi}
Having analyzed the total path count over all input terminals, we next consider
the number of paths from a specified input terminal.
Let $\bm{y}_h$ denote the path-count vector whose $j$th entry represents 
the number of paths from the $h$th input terminal to the $j$th output.
In place of Eq.~(\ref{app_eq:paths}), we then have
\begin{align}
\bm{y}_h &= \mathcal{M}\bm{1}_{(n,h)},
\label{app_eq:path_h} \\
\bm{1}_{(n,h)} &= (0,\ldots,0,1,0,\ldots,0)^\intercal,
\label{app_eq:1h}
\end{align}
where $\bm{1}_{(n,h)}$ denotes the $n$-dimensional vector whose $h$th entry is
one and all other entries are zero, with $h\!=\!0,1,\ldots,n\!-\!1$.
\begin{figure}[t]
\begin{center}
	\psfrag{X}[c][c][.95]{Normalized output index}
	\psfrag{Y}[c][c][.90]{
		\begin{picture}(0,0)
			\put(0,0){\makebox(-6,38)[c]{
				Normalized path count}}
			\put(-15,2){\makebox{($\times 10^{-2}$)}}
		\end{picture}
	}
	\psfrag{A}[c][c][0.9]{0} \psfrag{B}[c][c][0.9]{0.2}
	\psfrag{C}[c][c][0.9]{0.4} \psfrag{D}[c][c][0.9]{0.6}
	\psfrag{E}[c][c][0.9]{0.8} \psfrag{F}[c][c][0.9]{1}
	\psfrag{G}[r][r][0.9]{0} 
	\psfrag{H}[r][r][0.9]{2} \psfrag{J}[r][r][0.9]{4}
	\psfrag{K}[r][r][0.9]{6} \psfrag{L}[r][r][0.9]{8}
	\psfrag{M}[r][r][0.9]{10} \psfrag{N}[r][r][0.9]{12}
	\psfrag{O}[r][r][0.9]{12}
	\psfrag{P}[r][r][.8]{n: 16}
	\psfrag{Q}[r][r][.8]{32}
	\psfrag{R}[r][r][.8]{64}
	\includegraphics[width=60mm]{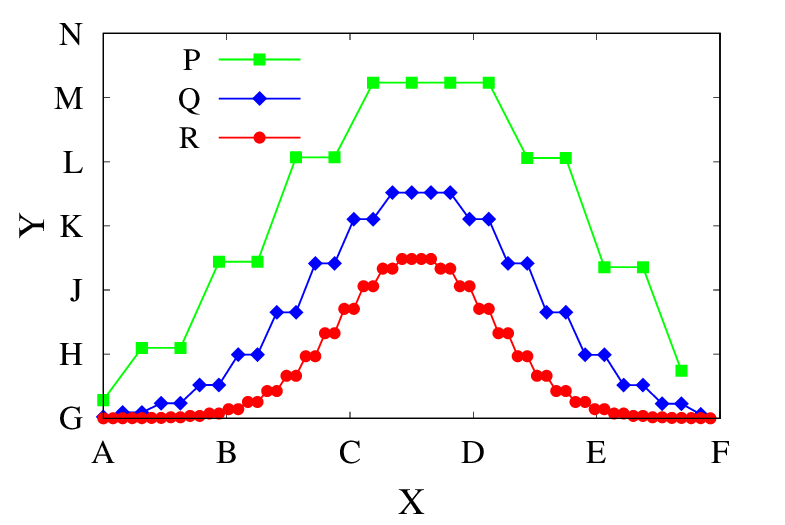}
\end{center}
\vspace*{-2mm}
\caption{Normalized path counts from the specified input terminal whose index is $n/2$ 
to the output terminals of a mesh architecture with $n\!=\!m\!=\!\{16,32,64\}$.
The path counts and output indices are normalized by the sum of path counts and $n$,
respectively.
}\label{app_fig:dim_onein}
\end{figure}

Figure~\ref{fig:pathcount} is obtained by applying
Eq.~(\ref{app_eq:path_h}) to the specified input indices.
It shows that the path-count distribution depends on the input position and is
concentrated within a limited range of output indices.
Figure~\ref{app_fig:dim_onein} shows the path counts from the input terminal
with index $h\!=\!n/2$ as functions of output index for mesh architectures
with $n\!=\!16,32,64$ and $m\!=\!n$.
The path counts are normalized by $\|\bm{y}_h\|_{1}$, and the output indices
by $n$. The similar profiles across the tested sizes show that the
concentration persists as the mesh dimension increases.

Thus, the mesh architecture contains numerous input-output paths, as shown in
Fig.~\ref{app_fig:pathcount}, and their multiplicities are strongly
nonuniform across output terminals, as shown in
Figs.~\ref{app_fig:pathprob} and \ref{app_fig:dim_onein}.

\section{Reachability}\label{app_sec:reach}
A rectangular mesh architecture with $n$ inputs needs at least $n\!-\!1$
MZI stages to satisfy the reachability property in Definition~\ref{dfn:reachability}.
To reduce the number of parameters in the mesh architecture,
we consider the truncated mesh architecture in Section~\ref{sec:limit}, whose
stage count is less than $n\!-\!1$. A unitary matrix implemented by such a mesh
contains \emph{trivial zero} entries.
These trivial zero entries force a banded structure on the unitary matrix.
We describe their positions in the matrix and their count,
relating them to paths from the inputs to the outputs. 

\begin{proposition}\label{prop:nonzero_pos}
Consider entries $g_{jh}$ in an $n\!\times\!n$ unitary matrix implemented by
the truncated mesh architecture with $m$ stages,
$m\!\in\!\mathbb{N}$ and $m\!<\!n$,
where $j$ and $h$ denote, respectively, the row and column indices,
$j,h\!\in\!\{0,1,\ldots,n\!-\!1\}$.
Entry $g_{jh}$ is nonzero (i.e., not trivially zero) if the following
inequality holds:
\begin{align}
\max( 2\hat{h}\!+\!1 -\!m, 0 ) &\leq j \leq 
\min( 2\hat{h}\!+\!m, n\!-\!1 ) \label{app_eq:trunc_nonzero}\\
\hat{h} &= \lfloor h/2 \rfloor\, . \nonumber
\end{align}
The number of nonzero entries ($\zeta_{\lnot 0}$) is
\begin{equation}
\zeta_{\lnot 0}\! =\! \sum_{h=0}^{n-1}\left\{
\min(2\hat{h}\!+\!m,n\!-\!1) -\!\max(2\hat{h}\!+\!1\!-\!m,0) +\!1 \right\} \, .
\label{app_eq:nnz}
\end{equation}
\end{proposition}
\begin{proof}
Entry $g_{jh}$ is structurally constrained to zero if no path exists from the
$h$th input terminal to the $j$th output terminal.
For a given input terminal $x_h$ ($h\!=\!0,1,\ldots,n\!-\!1$) and $m$ stages,
let the top and bottom horizons be the lines passing through the reachable
terminals with the minimum and maximum indices, respectively,
at each stage up to $m$.
The two horizons are
\begin{align}
\text{Bottom-horizon:}\quad & j=\min\left( 2\left\lfloor h/2 \right\rfloor +m, n\!-\!1\right) \\
\text{Top-horizon:}\quad & j=\max\left( 2\left\lfloor h/2\right\rfloor +1\!-\! m, 0\right)\, .
\label{app_eq:horizons}
\end{align}

\begin{figure}[t]
\begin{center}
	\psfrag{a}[c][c][1.0]{$x_0$} \psfrag{b}[c][c][1.0]{$x_1$}
	\psfrag{c}[c][c][1.0]{$x_2$} \psfrag{d}[c][c][1.0]{$x_3$}
	\psfrag{e}[c][c][1.0]{$x_4$} \psfrag{f}[c][c][1.0]{$x_5$}
	\psfrag{g}[c][c][1.0]{$x_6$} \psfrag{h}[c][c][1.0]{$x_7$}
	\psfrag{A}[c][c][1.0]{$y_0$} \psfrag{B}[c][c][1.0]{$y_1$}
	\psfrag{C}[c][c][1.0]{$y_2$} \psfrag{D}[c][c][1.0]{$y_3$}
	\psfrag{E}[c][c][1.0]{$y_4$} \psfrag{F}[c][c][1.0]{$y_5$}
	\psfrag{G}[c][c][1.0]{$y_6$} \psfrag{H}[c][c][1.0]{$y_7$}
	\psfrag{0}[c][c][.9]{0} \psfrag{1}[c][c][.9]{1}
	\psfrag{2}[c][c][.9]{2} \psfrag{3}[c][c][.9]{3}
	\psfrag{P}[c][c][.9]{Stage number}
	\psfrag{Q}[r][r][.9]{Top} \psfrag{R}[r][r][.9]{Bottom}
	\psfrag{S}[r][r][.9]{Top-horizon} \psfrag{T}[r][r][.9]{Bottom-horizon}
	\includegraphics[width=39mm]{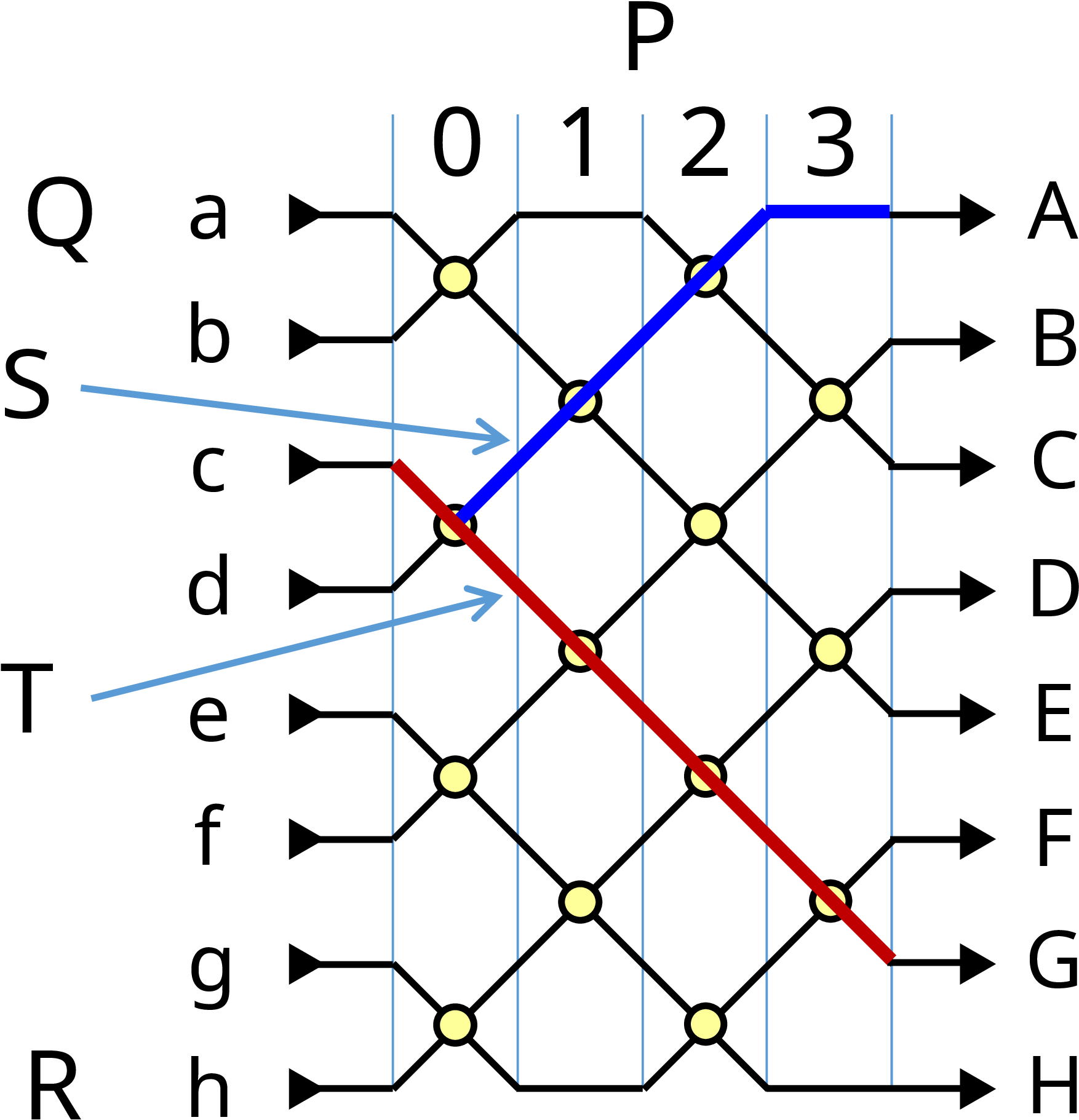}
\end{center}
\vspace*{-2mm}
\caption{Truncated mesh architecture with eight input terminals, eight output
terminals, and four stages. The top and bottom horizons of $x_{2}$ are shown
by thick solid lines.
}\label{app_fig:truncPath}
\end{figure}
Figure~\ref{app_fig:truncPath} shows an example of the top and bottom horizons
in the truncated mesh architecture with $n\!=\!8$ and $m\!=\!4 < n$.
The thick solid lines represent the two horizons starting from $x_2$.

Because every output terminal between the two horizons, inclusive, has at
least one path,
Eq.~(\ref{app_eq:trunc_nonzero}) holds.
Equation~(\ref{app_eq:nnz}) then follows from
Eq.~(\ref{app_eq:trunc_nonzero}).
\end{proof}

\section{Expected Squared Moduli of Matrix Entries}\label{app_sec:heatmap}
\begin{figure}[t]
\begin{center}
	\psfrag{A}[c][c][1.0]{$w_{00}$} \psfrag{B}[c][c][1.0]{$w_{10}$}
	\psfrag{C}[c][c][1.0]{$w_{01}$} \psfrag{D}[c][c][1.0]{$w_{11}$}
	\psfrag{T}[c][c][1.0]{$T_{(\text{MZI})}$}
	\includegraphics[width=48mm]{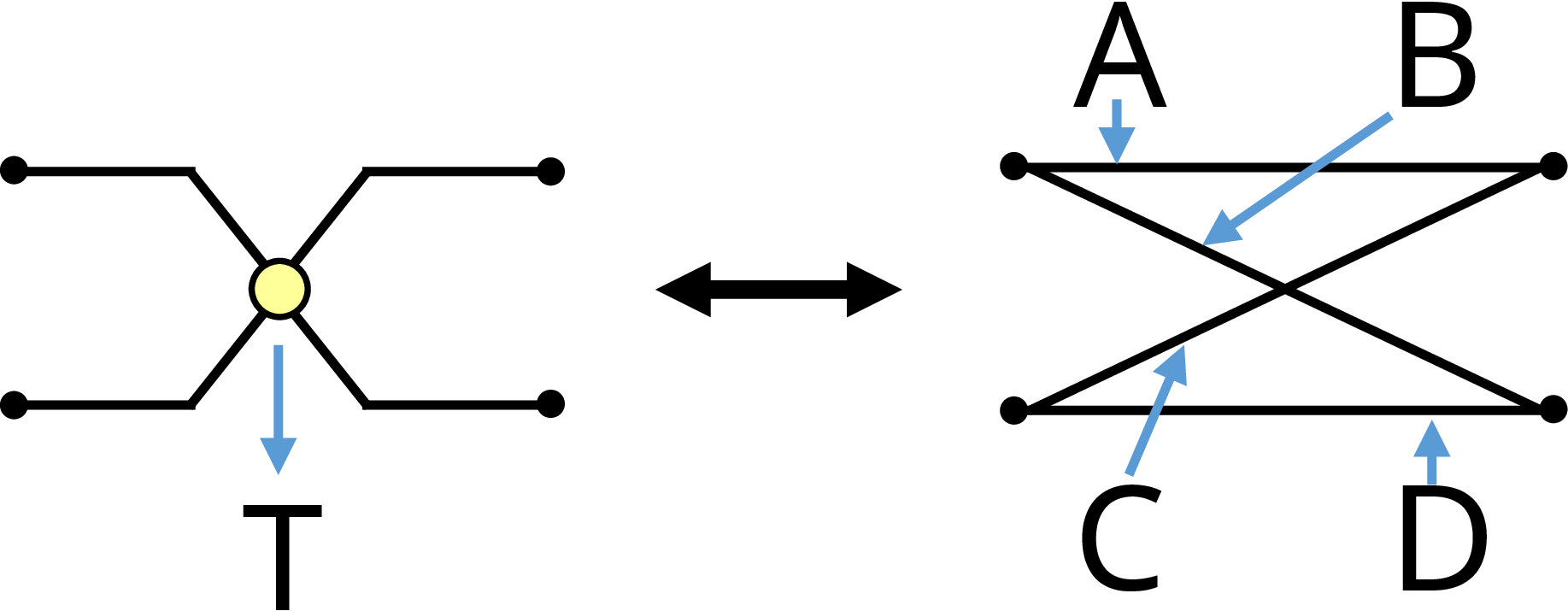}
\end{center}
\vspace*{-2mm}
\caption{Graph representation of an MZI: circuit with transfer matrix
$T_{(\text{MZI})}$ (left) and weighted graph whose edge weights correspond to
the matrix entries (right).
}\label{app_fig:2x2graph}
\end{figure}

It is known that a parameterized unitary matrix implemented by 
an MZI-based rectangular mesh architecture exhibits a banded structure 
when its parameters are drawn from a uniform random distribution \cite{pai}.
We first formulate the expected squared moduli of the matrix entries in terms
of the architectural structure.
We then explain the mechanism that produces the banded structure.

Next, we formulate the corresponding expectations 
for a PSBS-based butterfly architecture 
and show that the implemented unitary matrices exhibit entrywise statistical
uniformity when the parameters are sampled uniformly at random.

Finally, we identify common conditions under which the two formulations for
the expected squared moduli apply.

Both formulations rely on the following algebraic path lemma.

\begin{lemma}[Algebraic path problem \cite{baras}]\label{app_lemma:path-sum}
Consider a multistage bipartite graph in which each stage consists of $n$ vertices
and the edges between contiguous stages are weighted.
Let its $n\times n$ adjacency matrix $T$ be the product of the adjacency
matrices representing the successive bipartite stages. Then, entry $T_{jh}$ is
\begin{equation}
T_{jh} = \sum_{\substack{p(j,h,k)\\\in P_{(j,h)}}}\quad
\prod_{\substack{\text{edge}\\\in p_{(j,h,k)}}}w(\text{edge})\, ,
\label{app_eq:path-sum}
\end{equation}
where $P_{(j,h)}$ denotes the set of all paths from the $h$th vertex in the first stage
to the $j$th vertex in the final stage, $p_{(j,h,k)}$ the $k$th path in $P_{(j,h)}$,
and $w(\text{edge})$ denotes the edge weight.
\end{lemma}

\subsection{MZI-based rectangular mesh architecture}
	\label{app_ssec:heatmap_mzimesh}
\begin{figure}[t]
\begin{center}
	\psfrag{a}[c][c][1.0]{$x_0$} \psfrag{b}[c][c][1.0]{$x_1$}
	\psfrag{c}[c][c][1.0]{$x_2$} \psfrag{d}[c][c][1.0]{$x_3$}
	\psfrag{e}[c][c][1.0]{$x_4$} 
	\psfrag{A}[c][c][1.0]{$y_0$} \psfrag{B}[c][c][1.0]{$y_1$}
	\psfrag{C}[c][c][1.0]{$y_2$} \psfrag{D}[c][c][1.0]{$y_3$}
	\psfrag{E}[c][c][1.0]{$y_4$}
	\psfrag{X}[c][c][.9]{A} \psfrag{Y}[c][c][.9]{B}
	\psfrag{P}[c][c][.95]{Stage type}
	\psfrag{Q}[c][c][.95]{Stage number}
	\psfrag{0}[c][c][.9]{0} \psfrag{1}[c][c][.9]{1}
	\psfrag{2}[c][c][.9]{2} \psfrag{3}[c][c][.9]{3}
	\psfrag{R}[r][r][.9]{Top} \psfrag{S}[r][r][.9]{Bottom}
	\psfrag{u}[c][c][.8]{0} \psfrag{v}[c][c][.9]{2}
	\psfrag{w}[c][c][.8]{1}
	\psfrag{E}[l][l][1]{$p_{(0,0,0)}$}
	\psfrag{F}[l][l][1]{$p_{(0,0,1)}$}
	\psfrag{G}[r][r][.9]{MZI index}
	\includegraphics[width=48mm]{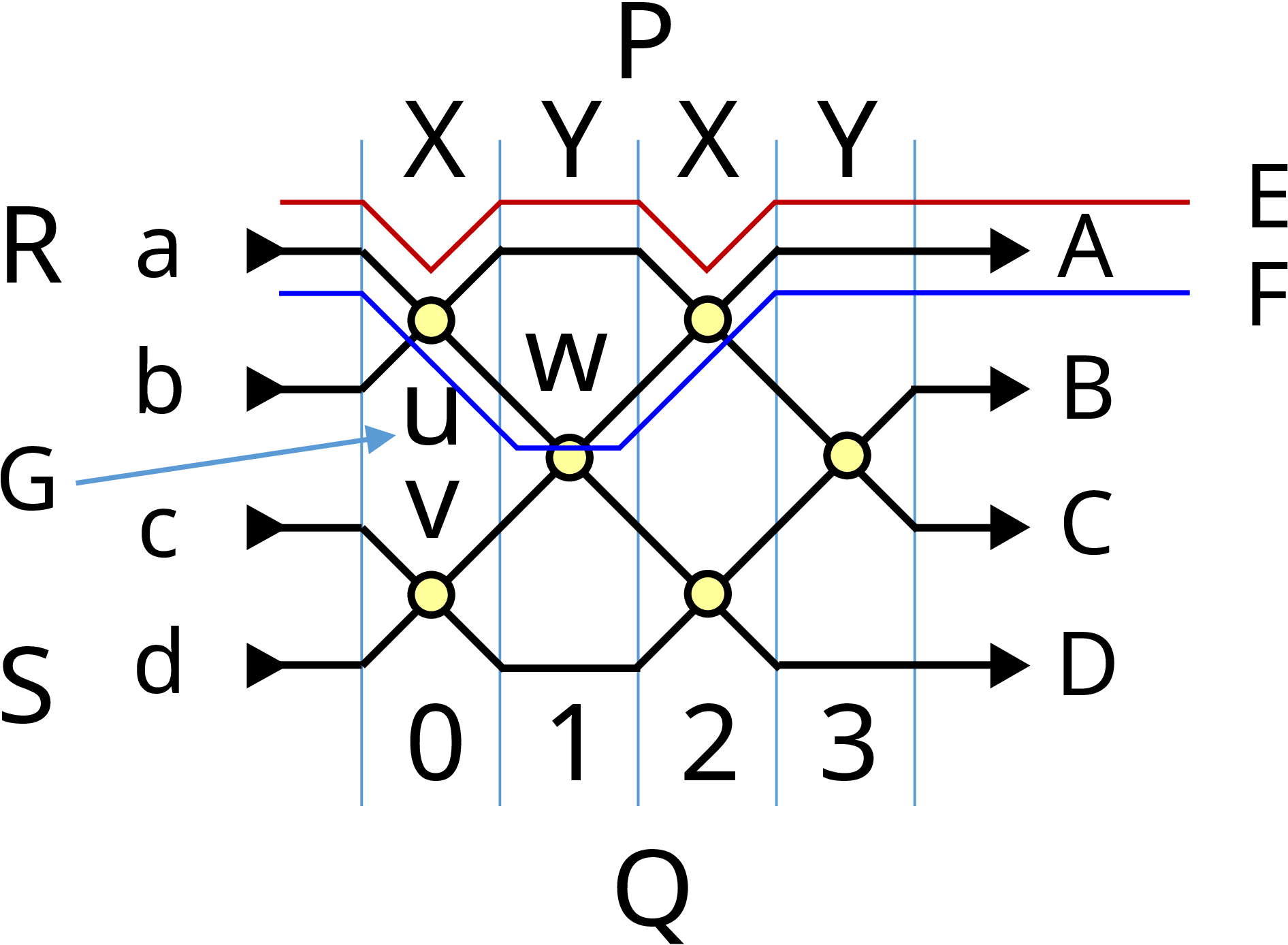}
\end{center}
\vspace*{-2mm}
\caption{Mesh architecture with four input and four output terminals.
}\label{app_fig:4inMesh_heatmap}
\end{figure}

\begin{proposition}\label{app_prop:expected}
Consider a rectangular mesh architecture whose fundamental building blocks are MZIs.
Let $P_{(j,h)}$ denote the set of paths from the $h$th input terminal to 
the $j$th output terminal in the mesh
and $p_{(j,h,k)}$ denote the $k$th path in $P_{(j,h)}$,
where $k$ is just a path index. 
Let $F(p_{(j,h,k)})$ denote the number of MZIs on path $p_{(j,h,k)}$.
Let $G$ denote a parameterized unitary matrix 
implemented by the MZI-mesh architecture 
and $g_{jh}$ the entry at the $j$th row and $h$th column in $G$.
If the phase-shifter angles, which parameterize $G$, are sampled uniformly at
random, the expected squared moduli are
\begin{equation}
\mathbb{E}\left[\,|g_{jh}|^2\,\right] = 
\sum_{p_{(j,h,k)}\in P_{(j,h)}} \frac{1}{2^{\,F(p_{(j,h,k)})}}\, .
\label{app_eq:expected}
\end{equation}
Note that the expectation depends only on the structural properties 
of the mesh and not on the phase-shifter parameters.
\end{proposition}

\begin{proof}\label{app_proof:expected}
A parameterized unitary matrix $G$ can be regarded as an adjacency matrix.
The fundamental building block, the $2\!\times\!2$ parameterized unitary matrix
$T_{\text{(MZI)}}(\phi,\theta)$ in Eq.~(\ref{eq:mzi}) is
\begin{equation*}
T_{(\text{MZI})}(\phi,\theta) 
= ie^{i\frac{\theta}{2}}
\begin{pmatrix}e^{i\phi}\sin{\frac{\theta}{2}}& \cos{\frac{\theta}{2}}\\
e^{i\phi}\cos{\frac{\theta}{2}}& -\sin{\frac{\theta}{2}}
\end{pmatrix} \, . 
\end{equation*}
The entry $w_{\alpha\beta}$ at the $\alpha$th row and the $\beta$th column 
of $T_{\text{(MZI)}}$
corresponds to the edge weight from the $\beta$th terminal to the $\alpha$th terminal 
in the bipartite graph in Fig.~\ref{app_fig:2x2graph}.
The weight $w_{\alpha\beta}$ can be written generally as
\begin{equation}
w_{\alpha\beta} = ie^{i\frac{\theta}{2}}\, e^{iz_{\alpha\beta}}\, 
\psi_{\alpha\beta}(\theta/2)\, ,
\label{app_eq:general_w}
\end{equation}
where $\psi_{\alpha\beta}(\theta/2)$ is either the sine or cosine of $\theta/2$,
depending on $\alpha$ and $\beta$, $z_{\alpha\beta}\!\in\!\{\phi,0,\pi\}$, 
and $\theta\!\in\![0,\pi]$.

By Lemma~\ref{app_lemma:path-sum}, the entry $g_{jh}$ is expressed by
\begin{equation}
g_{jh} = \sum_{\substack{p(j,h,k)\\\in P_{(j,h)}}}\quad
\prod_{\substack{\text{edge}\\\in p_{(j,h,k)}}}w(\text{edge})\, ,
\label{app_eq:gjh}
\end{equation}
where $w(\text{edge})\!\in\!\{w_{00},w_{01},w_{10},w_{11}\}$ if the two terminals 
for inputs and outputs are connected via an MZI as in Fig.~\ref{app_fig:2x2graph}, 
and $w(\text{edge})\!\in\!\{w_{00},w_{11},1\}$ otherwise.
We illustrate these cases using Fig.~\ref{app_fig:4inMesh_heatmap}.
Consider the paths from $x_0$ to $y_0$.
There exist two paths $p_{(j,h,k)}\!\in\!\{p_{(0,0,0)},\,p_{(0,0,1)}\}$.
Assume that the MZIs at the even- and odd-numbered stages
are respectively given the labels of even ($0,2,\ldots$) and odd ($1,3,\ldots$)
in ascending order from top to bottom.
The $u$th MZI in stage $s$ is identified by the tuple $(s,u)$ and denoted by
$\text{MZI}_{(s,u)}$.
Using the tuple, let $w_{\alpha\beta(s,u)}$ denote the edge weight 
$w_{\alpha\beta}$ of $\text{MZI}_{(s,u)}$.
The path $p_{(0,0,0)}$ has the edge weights, $w_{00(0,0)}$, 1, $w_{00(2,0)}$, and 1 
in ascending order of the stage number from 0 to 3.
By contrast, $p_{(0,0,1)}$ has $w_{10(0,0)}$, $w_{00(1,1)}$, $w_{01(2,0)}$, and 1.
Note that a path has a unique edge weight for each stage.

We focus on a single path $p_{(j,h,k)}$.
Based on Eqs.~(\ref{app_eq:general_w}) and (\ref{app_eq:gjh}),
the product of edge weights $\Omega$ is expressed by
\begin{align}
\Omega(p_{(j,h,k)}) &=
\prod_{\substack{\text{edge}\\\in p_{(j,h,k)}}}w(\text{edge}) \nonumber \\
&=
\exp\left( \frac{i}{2}\sum_{s=0}^{m-1}\theta_{(s,u)} \right)
\exp\left( i \sum_{s=0}^{m-1} z_{\alpha\beta(s,u)} \right) \nonumber\\
&\quad\times\prod_{s=0}^{m-1}i\psi_{\alpha\beta(s,u)}
\left( \frac{\theta_{(s,u)}}{2} \right)\, ,
\label{app_eq:Omega}
\end{align}
where $m$ denotes the number of stages and 
$\theta_{(s,u)}$, $z_{\alpha\beta(s,u)}$, and $\psi_{\alpha\beta(s,u)}$ 
denote, respectively, $\theta$, $z_{\alpha\beta}$, and $\psi_{\alpha\beta}$
identified by $(s,u)$ on $p_{(j,h,k)}$. 
A stage index $s$ in Eq.~(\ref{app_eq:Omega}) is omitted 
when no MZI exists in that stage on the path, 
and $u$ is determined by the path and stage.
By using $\Omega({p_{(j,h,k)}})$, the squared modulus of $g_{jh}$ is
\begin{equation}
|g_{jh}|^2 =\! 
\left( \sum_{\substack{p(j,h,k')\\\in P_{(j,h)}}} \Omega^{*}(p_{(j,h,k')}) \right)\!
\left( \sum_{\substack{p(j,h,k)\\\in P_{(j,h)}}} \Omega(p_{(j,h,k)}) \right)\, ,
\label{app_eq:sq_modulus}
\end{equation}
where $\Omega^{*}(p_{(j,h,k')})$ denotes the complex conjugate of
$\Omega(p_{(j,h,k')})$.

Assume that the physical phase parameters
$\phi_{(s,u)}\!\in\![0,2\pi)$ and
$\theta_{(s,u)}\!\in\![0,\pi]$ of distinct MZIs are mutually independent.
When a parameter is associated with a particular path, we append the path
index $k$ to its subscript while retaining the same random variable for paths
that traverse the same MZI. Under this convention,
we consider the expectation of
$\Omega^{*}(p_{(j,h,k')})\,\Omega(p_{(j,h,k)})$ 
on the right-hand side of Eq.~(\ref{app_eq:sq_modulus}) by separating two
cases: a conjugate-pair product along the same path
($k'\!=\!k$) and a cross product involving two distinct paths ($k'\!\neq\!k$).
\begin{enumerate}[(a)]
\item Case of $p_{(j,h,k')} = p_{(j,h,k)}$:\\
The product is expressed as 
\begin{equation}
\Omega^{*}(p_{(j,h,k)})\,\Omega(p_{(j,h,k)}) =
\prod_{s=0}^{m-1}\left( \psi_{\alpha\beta(s,u)}
\left( \frac{\theta_{(s,u)}}{2} \right)\right)^2\, .
\label{app_eq:absOmega}
\end{equation}
The conjugate-pair products associated with the first two factors on the
right-hand side of Eq.~(\ref{app_eq:Omega}) are both unity.
Since $\psi_{\alpha\beta(s,u)}(\theta_{(s,u)}/2)\!\in\!
\{\sin(\theta_{(s,u)}/2),\cos(\theta_{(s,u)}/2)\}$, 
Eq.~(\ref{app_eq:absOmega}) is rewritten as
\begin{equation}
\Omega^{*}(p_{(j,h,k)})\Omega(p_{(j,h,k)}) =
\prod_{s=0}^{m-1} \left\{\frac{1}{2} \left( 1\pm \cos(\theta_{(s,u)}) \right) \right\}\, ,
\label{app_eq:absOmega_cos}
\end{equation}
where the upper ($+$) and lower ($-$) signs correspond to the cases of  
$\psi_{\alpha\beta(s,u)}(\theta_{(s,u)}/2)\!=\!\cos(\theta_{(s,u)}/2)$ 
and $\sin(\theta_{(s,u)}/2)$,
respectively.

The expectation of $\Omega^{*}(p_{(j,h,k)})\Omega(p_{(j,h,k)})$ is the product of
individual expectations of $(1\pm \cos(\theta_{(s,u)}))/2$ 
for $\theta_{(s,u)}\!\in\![0,\pi]$.
Each individual expectation is
\begin{equation}
\frac{1}{2\pi}\int_0^{\pi} \left( 1\pm \cos(\theta_{(s,u)})\right)\, d\theta_{(s,u)}
= \frac{1}{2}\, .
\label{app_eq:indi_same_path}
\end{equation}
Since the number of MZIs on the path $p_{(j,h,k)}$ is $F(p_{(j,h,k)})$,
the expectation of the conjugate-pair product is
\begin{equation}
\mathbb{E}\left[\,\Omega^{*}(p_{(j,h,k)})\,\Omega(p_{(j,h,k)})\,\right] 
= \frac{1}{2^{F(p_{(j,h,k)})}}\, .
\label{app_eq:expected_indi_same_path}
\end{equation}

\item Case of $p_{(j,h,k')} \neq p_{(j,h,k)}$:\\
We focus on the second term 
$\exp(i\sum_{s=0}^{m-1} z_{\alpha\beta(s,u)})$ in Eq.~(\ref{app_eq:Omega}).
To distinguish the parameters $z_{\alpha\beta(s,u)}$ on two distinct paths,
we introduce the path indices $k$ and $k'$ 
and denote them by $z_{\alpha\beta(k,s,u)}$ and $z_{\alpha\beta(k',s,u')}$, respectively.
The term related to $\exp(i\sum_{s=0}^{m-1} z_{\alpha\beta(s,u)})$ in 
$\Omega^{*}(p_{(j,h,k')})\,\Omega(p_{(j,h,k)})$ is represented by
\begin{equation}
\exp\left( i \sum_{s=0}^{m-1}(z_{\alpha\beta(k,s,u)} -z_{\alpha\beta(k',s,u')}) \right)\, ,
\label{app_eq:diffpaths}
\end{equation}
where $z_{\alpha\beta(k,s,u)}$ and $z_{\alpha\beta(k',s,u')}$ can be parameters 
$\phi_{(k,s,u)}$ and $\phi_{(k',s,u')}$ identified by the path and MZI indices.

Recall that all parameters are sampled uniformly from
\begin{equation}
\phi_{(k,s,u)} \in [0, 2\pi)\quad\text{and}\quad
\theta_{(k,s,u)} \in [0, \pi]\, . \label{app_eq:param_range}
\end{equation}

At least one term in the exponent in Eq.~(\ref{app_eq:diffpaths}) remains with the form of 
$\exp(i z_{\alpha\beta(k,s,u)})\!=\!\exp(i\phi_{(k,s,u)})$.
When the expectation is calculated,
this term is integrated from 0 to $2\pi$ as
\begin{equation}
\int_{0}^{2\pi} \exp(i\phi_{(k,s,u)})\, d\phi_{(k,s,u)} = 0\, . 
\label{app_eq:diffpaths_integral}
\end{equation}
Since this zero term enters the expression multiplicatively,
it causes the entire expectation in this case, $k'\!\neq\!k$, to vanish.
\end{enumerate}

From these two cases,
the expected squared modulus of $g_{jh}$ is the sum of the same-path contributions
in Eq.~(\ref{app_eq:expected_indi_same_path}):
\begin{equation}
\mathbb{E}\left[\, |g_{jh}|^2\,\right] =
\sum_{p(j,h,k)\in P_{(j,h)}} \frac{1}{2^{F(p_{(j,h,k)})}}\, .
\nonumber
\end{equation}
\end{proof}

\begin{ex}
We calculate the expected squared modulus of $g_{00}$ in Fig.~\ref{app_fig:4inMesh_heatmap}. 
There are two paths: $p_{(0,0,0)}$, containing two MZIs, and
$p_{(0,0,1)}$, containing three. Equation~(\ref{app_eq:expected}) gives
\begin{equation}
\mathbb{E}\left[\,|g_{00}|^2\,\right] = \frac{1}{2^2}+\frac{1}{2^3} =\frac{3}{8} \,.
\end{equation}

For comparison, we also derive the expectation directly
based on Eq.~(\ref{app_eq:Omega})-(\ref{app_eq:diffpaths_integral}).
$\Omega(p_{(0,0,0)})$ and $\Omega(p_{(0,0,1)})$ are
\begin{align}
\Omega(p_{(0,0,0)}) &= ie^{(i/2)(\theta_{(0,0)}+\theta_{(2,0)})}
e^{i(\phi_{(0,0)}+\phi_{(2,0)})}\nonumber\\
&\quad \times\sin\frac{\theta_{(0,0)}}{2} \sin\frac{\theta_{(2,0)}}{2}
\label{app_eq:exOmega_p000} \\
\Omega(p_{(0,0,1)}) &= ie^{(i/2)(\theta_{(0,0)}+\theta_{(1,1)}+\theta_{(2,0)})}
e^{i(\phi_{(0,0)}+\phi_{(1,1)}+\phi_{(2,0)})}\nonumber\\
&\quad \times\cos\frac{\theta_{(0,0)}}{2} \sin\frac{\theta_{(1,1)}}{2}
\cos\frac{\theta_{(2,0)}}{2}\, ,
\label{app_eq:exOmega_p001}
\end{align}
where the path index $k$ in the subscripts of $\theta_{(k,s,u)}$ and $\phi_{(k,s,u)}$
are omitted for simplicity. The conjugate-pair products are
\begin{align}
& \Omega^{*}(p_{(0,0,0)})\Omega(p_{(0,0,0)})\nonumber\\
&\quad\quad= 
\frac{1}{2^{2}}\left( 1-\cos\theta_{(0,0)} \right)\left( 1-\cos\theta_{(2,0)} \right)
\\
& \Omega^{*}(p_{(0,0,1)})\Omega(p_{(0,0,1)})\nonumber\\
&\quad\quad= 
\frac{1}{2^{3}}\left( 1+\cos\theta_{(0,0)} \right)
\left( 1-\cos\theta_{(1,1)} \right) \left( 1+\cos\theta_{(2,0)} \right)
\end{align}
By contrast, 
for the two distinct paths,
represented by $\Omega^{*}(p_{(0,0,0)})\Omega(p_{(0,0,1)})$,
the term proportional to $e^{i\phi_{(1,1)}}$ remains.
The same holds true for $\Omega^{*}(p_{(0,0,1)})\Omega(p_{(0,0,0)})$.
These expectations vanish by Eq.~(\ref{app_eq:diffpaths_integral}).
The squared modulus of $g_{00}$ is therefore
\begin{equation}
|g_{00}|^2 =\! 
\left( \sum_{k\in\{0,1\}} \Omega^{*}(p_{(0,0,k)}) \right)\!
\left( \sum_{k\in\{0,1\}} \Omega(p_{(0,0,k)}) \right)\, .
\label{app_eq:sq_modulus_00}
\end{equation}
Because the expectation of $|g_{00}|^2$ is the sum of individual expectations
over $\phi_{(s,u)}\!\in\![0,2\pi)$ and $\theta_{(s,u)}\!\in\![0,\pi]$, 
and the parameters are separable,
each term in Eq.~(\ref{app_eq:sq_modulus_00}) can be considered in isolation
as follows:
\begin{align}
&\mathbb{E}\left[ \Omega^{*}(p_{(0,0,0)})\Omega(p_{(0,0,0)})\right]\nonumber\\
&\quad =\frac{1}{2^2}\times \frac{1}{\pi}
\int_0^{\pi}\!\!\left( 1\!-\!\cos\theta_{(0,0)} \right)d\theta_{(0,0)}
\nonumber\\
&\quad\quad\quad\,\times \frac{1}{\pi}
\int_0^{\pi}\!\!\left( 1\!-\!\cos\theta_{(2,0)} \right)d\theta_{(2,0)}
\nonumber\\
&\quad =\frac{1}{2^2} \label{app_eq:ex1}\\
&\mathbb{E}\left[ \Omega^{*}(p_{(0,0,1)})\Omega(p_{(0,0,1)})\right]
=\frac{1}{2^3} \label{app_eq:ex2}\\
&\mathbb{E}\left[ \Omega^{*}(p_{(0,0,0)})\Omega(p_{(0,0,1)})\right]\nonumber\\
&\quad =\frac{1}{2\pi}\int_0^{2\pi}e^{i\phi_{(1,1)}}d\phi_{(1,1)}
\times [\text{other terms}] = 0 \label{app_eq:ex3}\\
&\mathbb{E}\left[ \Omega^{*}(p_{(0,0,1)})\Omega(p_{(0,0,0)})\right]
= 0\, . \label{app_eq:ex4}
\end{align}
From Eqs.~(\ref{app_eq:ex1})--(\ref{app_eq:ex4}),
the expected squared modulus of $g_{00}$ is represented by
\begin{equation}
\mathbb{E}\left[\,|g_{00}|^2\,\right] = \frac{1}{2^2}+\frac{1}{2^3} =\frac{3}{8} \,.
\end{equation}
\end{ex}

\begin{corollary}\label{app_coroll:impl}
Consider a rectangular mesh architecture with $n$ inputs and $n$ outputs, 
consisting of $m$ alternating type-A (even-numbered) and type-B (odd-numbered) stages,
starting with the A-type as shown in Fig.~\ref{app_fig:mesh}.
Let $\widehat{M}_A$ and $\widehat{M}_B$ denote matrices representing 
the A-type and B-type stages, respectively.
The expected squared moduli given by the right-hand side of Eq.~(\ref{app_eq:expected}) 
form the product $\widehat{M}$ of $m$ matrices $\widehat{M}_A$ and
$\widehat{M}_B$, where

\begin{align}
\widehat{M}_A &=\! \frac{1}{2}\left[ \begin{matrix}
1 & 1 & & & \\
1 & 1 & & & \\
 & & \ddots & & \\
 & &  & 1 & 1\\
 & &  & 1 & 1
\end{matrix}\right]_{n\times n} \\
\widehat{M}_B &=\! \frac{1}{2}\left[ \begin{matrix}
2& &  &      & &  &\\
 &1&1 &      & &  &\\
 &1&1 &      & &  &\\
 & &  &\ddots& &  &\\
 & &  &      &1&1 &\\
 & &  &      &1&1 &\\
 & &  &      & &  &2
\end{matrix}\right]_{n\times n}
\end{align}
\begin{equation}
\widehat{M} = \underbrace{
\cdots \widehat{M}_B\,\widehat{M}_A\,\widehat{M}_B\,\widehat{M}_A}_{m}
\end{equation}
\begin{equation}
\mathbb{E}\left[|g_{jh}|^2\right] = \widehat{M}_{jh}\, ,
\label{app_eq:expect_M}
\end{equation}
where $\widehat{M}_{jh}$ denotes the entry in row $j$ and column $h$ of
$\widehat{M}$, and blank entries denote zero.
This formulation enables a direct calculation and is well suited to implementation.
\end{corollary}

\subsection{PSBS-based butterfly architecture}
	\label{app_ssec:heatmap_psbsbfly}
We consider the $n$-input PSBS-butterfly architecture described in
Section~\ref{ssec:bfly}.
Provided that each butterfly block remains intact, uniformly random settings
of the phase shifters in the PSBS-butterfly architecture yield
$\mathbb{E}[\,|g_{jh}|^2]\!=\!1/n$ for every entry of the implemented
$n\!\times\!n$ unitary matrix. The expected squared moduli follow
Eq.~(\ref{app_eq:expected}), as in the MZI-mesh architecture, but the
butterfly topology makes them equal across all entries.

\begin{proposition}\label{app_prop:psbsbfly_expected}
Consider a butterfly architecture whose fundamental building blocks are PSBS
units and that has $n$ inputs ($n\!=\!2^t$, $t\!\in\!\mathbb{N}$) and
$b$ butterfly blocks ($b\!\in\!\mathbb{N}$).
Let $G$ denote a parameterized unitary matrix 
implemented by the PSBS-butterfly architecture 
and $g_{jh}$ the entry at the $j$th row and $h$th column in $G$.
If the phase-shifter angles, which parameterize $G$, are sampled uniformly at
random, the expected squared modulus of every entry $g_{jh}$ is $1/n$.
\end{proposition}

\begin{proof}
We first derive the squared modulus of $g_{jh}$ for a topology containing a
single butterfly block, i.e., for $b\!=\!1$. We then extend the derivation to
an arbitrary number of blocks.

The fundamental building block, the $2\!\times\!2$ parameterized unitary matrix
$T_{\text{(PSBS)}}(\phi)$ in Eq.~(\ref{eq:psbs}) is
\begin{equation*}
T_{(\text{PSBS})}(\phi) 
= \frac{1}{\sqrt{2}}
\begin{pmatrix}e^{i\phi}& i\\
ie^{i\phi}& 1
\end{pmatrix} \, . 
\end{equation*}
The entry $w_{\alpha\beta}$ at the $\alpha$th row and the $\beta$th column 
of $T_{\text{(PSBS)}}$ can be regarded as 
the edge weight from the $\beta$th terminal to the $\alpha$th terminal 
in the bipartite graph in Fig.~\ref{app_fig:2x2graph}(right).
Then, $w_{\alpha\beta}$ is generalized as
\begin{equation}
w_{\alpha\beta} = \frac{1}{\sqrt{2}}\, e^{iz_{\alpha\beta}}\,,
\label{app_eq:psbs_w}
\end{equation}
where $z_{\alpha\beta}\!\in\!\{\phi,\pi/2,\phi\!+\!\pi/2,0\}$, 
and $\phi\!\in\! [0,2\pi)$.

Similarly to the MZI-mesh architecture,
the entry $g_{jh}$ is expressed by Eq.~(\ref{app_eq:gjh}).
The product of edge weights $\Omega$ on path $p_{(j,h,k)}$ is expressed by
\begin{align}
\Omega(p_{(j,h,k)}) &=
\prod_{\substack{\text{edge}\\\in p_{(j,h,k)}}}w(\text{edge}) \nonumber \\
&=
\left(\frac{1}{\sqrt{2}}\right)^{F(p_{(j,h,k)})}
\exp\left( i \sum_{s=0}^{m-1} z_{\alpha\beta(s,u)} \right) \, ,
\label{app_eq:psbs_Omega}
\end{align}
where $F(p_{(j,h,k)})$ is the number of PSBS units on $p_{(j,h,k)}$, 
$z_{\alpha\beta(s,u)}$ is $z_{\alpha\beta}$ 
at the $u$th PSBS in the $s$th stage on $p_{(j,h,k)}$, 
and $m$ denotes the number of stages.

For $b\!=\!1$, every input-output pair has a unique path
($|P_{(j,h)}|\!=\!1$), and every path contains $\log_2 n$ PSBS units
($F(p_{(j,h)})\!=\!\log_2 n$, with $k$ omitted). Therefore,
\begin{align}
|g_{jh}|^2 &= \Omega^{*}(p_{(j,h)})\,\Omega(p_{(j,h)})\nonumber \\
&= \left( \frac{1}{\sqrt{2}} \right)^{F(p_{(j,h)})} 
\exp\left( -i \sum_{s=0}^{m-1} z_{\alpha\beta(s,u)} \right)\nonumber \\
&\quad\times \left( \frac{1}{\sqrt{2}} \right)^{F(p_{(j,h)})} 
\exp\left( i \sum_{s=0}^{m-1} z_{\alpha\beta(s,u)} \right) \nonumber \\
&= \frac{1}{2^{F(p_{(j,h)})}} = \frac{1}{n}\, .
\label{app_eq:psbs_absOmega}
\end{align}

Next, we find the expected squared modulus of entry $g_{jh}$ in 
an $n\!\times\! n$ unitary matrix $G$
implemented by the PSBS-butterfly architecture consisting of $b$ butterfly blocks 
($b\!\in\!\mathbb{N}$),
where $n^{b-1}$ paths exist between every input-output pair
and $b\log_2 n$ PSBS units exist on each path.
Since multiple paths exist from the $h$th input to the $j$th output,
$|g_{jh}|^2$ is represented by Eq.~(\ref{app_eq:sq_modulus}) as
\begin{equation}
|g_{jh}|^2 =\! 
\left( \sum_{\substack{p(j,h,k')\\\in P_{(j,h)}}} \Omega^{*}(p_{(j,h,k')}) \right)\!
\left( \sum_{\substack{p(j,h,k)\\\in P_{(j,h)}}} \Omega(p_{(j,h,k)}) \right)\, .
\nonumber
\end{equation}
We consider the expectation of
$\Omega^{*}(p_{(j,h,k')})\,\Omega(p_{(j,h,k)})$ under the assumption that the
phase parameters of distinct PSBS units are mutually independent and uniformly
distributed over $[0,2\pi)$. Parameters associated with a PSBS unit shared by
the two paths represent the same random variable. We use the following two
cases.
\begin{enumerate}[(a)]
\item Case of $p_{(j,h,k')} = p_{(j,h,k)}$:\\
Analogously to Eq.~(\ref{app_eq:psbs_absOmega}),
\begin{equation}
\Omega^{*}(p_{(j,h,k)})\,\Omega(p_{(j,h,k)})
= \frac{1}{2^{F(p_{(j,h,k)})}}\, .
\label{app_eq:psbs_absOmega_same}
\end{equation}
The expectation is given by the same expression.
\item Case of $p_{(j,h,k')} \neq p_{(j,h,k)}$:\\
Two distinct paths sharing the same endpoints have at least one merge point,
corresponding to a PSBS unit, 
at which two distinct incoming edges merge into a single outgoing edge.
Based on Fig.~\ref{app_fig:2x2graph} with $T_{\text{(PSBS)}}$ 
instead of $T_{\text{(MZI)}}$ and Eq.~(\ref{app_eq:psbs_w}),
a tuple of two edge weights at the merge point $(w_{\alpha\beta},w_{\alpha,\beta'})$
is either $(e^{i\phi}/\sqrt{2},i/\sqrt{2})$ or $(ie^{i\phi}/\sqrt{2},1/\sqrt{2})$,
where $\phi$ denotes the phase parameter of the merge-point PSBS unit.
This means that at least one exponential function with an exponent of $\pm i\phi$
remains in $\Omega^{*}(p_{(j,h,k')})\,\Omega(p_{(j,h,k)})$.
In this case, the expectation contains the following term:
\begin{equation}
\frac{1}{4\pi}\int_0^{2\pi} e^{\pm i\phi}\,d\phi = 0\,.
\label{app_eq:psbs_expected_zero}
\end{equation}
Since the phase parameters are independent, 
the expectation of $\Omega^{*}(p_{(j,h,k')})\,\Omega(p_{(j,h,k)})$ 
factorizes into expectation terms associated with
$e^{iz_{\alpha\beta(s,u)}}$ and $e^{iz_{\alpha\beta(s,u')}}$ in Eq.~(\ref{app_eq:psbs_w}),
and therefore vanishes.
\end{enumerate}

From the results in the two cases,
the expected squared modulus of $g_{jh}$ contains only the same-path terms:
\begin{align}
\mathbb{E}\left[ |g_{jh}|^2 \right] 
&= \sum_{p(j,h,k)\in P_{(j,h)}} \Omega^{*}(p_{(j,h,k)})\,\Omega(p_{(j,h,k)})\nonumber\\
&= \sum_{p(j,h,k)\in P_{(j,h)}} \frac{1}{2^{\,F(p_{(j,h,k)})}} \, .\nonumber
\end{align} 
This equation is identical to Eq.~(\ref{app_eq:expected}).
In the butterfly architecture with $b$ blocks, 
every path $p_{(j,h,k)}$ contains $b\log_{2}n$ PSBS units, independently of $(j,h,k)$,
i.e., $F(p_{(j,h,k)})\!=\!b\log_{2}n$,
and the number of paths between any input-output pair is $n^{b-1}$, i.e., $|P_{(j,h)}|\!=\!n^{b-1}$.
Then, $\mathbb{E}\left[ |g_{jh}|^2 \right]$ is expressed by
\begin{equation}
\mathbb{E}\left[ |g_{jh}|^2 \right] = n^{b-1}\frac{1}{2^{\,(b\log_2 n)}} = \frac{1}{n}\,.
\label{app_eq:psbsbfly_expected}
\end{equation}
\end{proof}

\begin{corollary}\label{app_psbsbfly_coroll:impl}
Consider a PSBS-butterfly architecture with $n$ inputs and $n$ outputs, 
which consists of $b$ butterfly blocks.
Let $\widehat{M}_{F_r}$ denote the matrix representing the $r$th butterfly block,
where $r\!=\!0,1,2,\ldots,b\!-\!1$.
This matrix is defined by
\begin{equation}
\widehat{M}_{F_r} = \frac{1}{n}
\left[ \begin{matrix}
1 &\cdots& &\cdots& 1\\
\vdots & 1 & & & \vdots\\
 & & \ddots & & \\
\vdots& &  & 1 &\vdots\\
1 &\cdots&  &\cdots& 1
\end{matrix}\right]_{n\times n} =
\frac{1}{n}\, \bm{1}_{n\times n}\,.
\label{app_eq:heatmap_bfly_matrix}
\end{equation}
Matrix $\widehat{M}_{F}$, whose entries are  
the expected squared moduli in Eq.~(\ref{app_eq:expected}),
is expressed by 
\begin{equation}
\widehat{M}_{F} = \left(\widehat{M}_{F_r}\right)^{b}
=\left(\frac{1}{n}\,\bm{1}_{n\times n}\right)^{b}
=\frac{1}{n}\,\bm{1}_{n\times n}\, .
\end{equation}
Therefore, each expected squared modulus is
\begin{equation}
\mathbb{E}\left[|g_{jh}|^2\right] = \widehat{M}_{F,jh} = \frac{1}{n}\,.
\end{equation}
\end{corollary}

\subsection{Generalization}\label{app_ssec:heatmap_general}
For both the MZI-mesh and PSBS-butterfly architectures,
Eq.~(\ref{app_eq:expected}) gives the expected squared modulus of entry
$g_{jh}$ of the implemented unitary matrix $G$:
\[
\mathbb{E}\left[ |g_{jh}|^2 \right] 
= \sum_{p_{(j,h,k)}\in P_{(j,h)}} \frac{1}{2^{\,F(p_{(j,h,k)})}} \, .
\]
We now characterize the $2\!\times\!2$ unitary building blocks that yield this form.

\begin{proposition}\label{app_prop:expected_form}
Let $G\!\in\!\mathbb{C}^{n\times n}$ be an $n\!\times\!n$ 
parameterized unitary matrix 
implemented by an architecture with $n$ inputs and $n$ outputs and composed of
$m$ stages indexed by $s\!=\!0,1,\ldots,m\!-\!1$.
Each stage contains fundamental building blocks such as MZIs and PSBSs,
each of which implements a $2\times 2$ unitary matrix 
represented by 
\begin{equation*}
T_{(s,u)} = \begin{pmatrix}w_{00(s,u)}&w_{01(s,u)}\\
w_{10(s,u)}&w_{11(s,u)} \end{pmatrix} \, ,
\end{equation*}
where $(s,u)$ identifies the $u$th building block in stage $s$
and $w_{\alpha\beta(s,u)}$, $\alpha,\beta\!\in\!\{0,1\}$, is a weight.
Let $P_{(j,h)}$ be the set of paths from input port $h$ to output port $j$,
let $p_{(j,h,k)}\!\in\!P_{(j,h)}$ denote its $k$th path, and let
$F(p_{(j,h,k)})$ denote the number of building blocks along that path.
Assume that the parameters in each building block are 
independently and uniformly distributed.

If the $2\!\times\!2$ building block satisfies the following properties:
\begin{enumerate}[(i)]
\item $\forall \alpha,\beta,(s,u)$,\quad 
$\mathbb{E}[\,|w_{\alpha\beta(s,u)}|^2\,] = 1/2$,
\item for every $(s,u)$, either
\[
\mathbb{E}[\,w^{*}_{\alpha\beta'(s,u)}w_{\alpha\beta(s,u)}\,]=0
\quad\text{for all }\alpha\text{ and }\beta'\!\neq\!\beta,
\]
or
\[
\mathbb{E}[\,w^{*}_{\alpha'\beta(s,u)}w_{\alpha\beta(s,u)}\,]=0
\quad\text{for all }\beta\text{ and }\alpha'\!\neq\!\alpha.
\]
\end{enumerate}
then the expected squared modulus of each entry $g_{jh}$ of $G$ is given by
\begin{equation*}
\mathbb{E}\left[\,|g_{jh}|^2\,\right]
= \sum_{p_{(j,h,k)} \in P_{(j,h)}} \frac{1}{2^{\,F(p_{(j,h,k)})}}.
\end{equation*}
\end{proposition}

\begin{proof}
The entry $g_{jh}$ of $G$ is expressed as a sum over all paths 
from the $h$th input to the $j$th output:
\[
g_{jh} = \sum_{p_{(j,h,k)} \in P_{(j,h)}} \Omega(p_{(j,h,k)}),
\]
where $\Omega(p_{(j,h,k)})$ denotes the product of weights along the path
$p_{(j,h,k)}$.
The squared modulus is
\[
|g_{jh}|^2 = 
\sum_{k}\sum_{k'} \Omega^{*}(p_{(j,h,k')})\,\Omega(p_{(j,h,k)}).
\]
Taking expectation, we obtain
\[
\mathbb{E}\left[\,|g_{jh}|^2\,\right] =
\sum_{k}\sum_{k'}
\mathbb{E}\!\left[\Omega^{*}(p_{(j,h,k')})\,\Omega(p_{(j,h,k)})\right].
\]
We evaluate the expectation by separating two cases.

\begin{enumerate}[(a)]
\item Case of $p_{(j,h,k')}\!=\!p_{(j,h,k)}$:\\
For identical paths, $\Omega^{*}(p_{(j,h,k)})\Omega(p_{(j,h,k)})$ 
is a product of squared magnitudes of weights along the path.
By assumption (i), each factor contributes $1/2$, and hence
\[
\mathbb{E}\!\left[\,\Omega^{*}(p_{(j,h,k)})\Omega(p_{(j,h,k)})\,\right]
= \frac{1}{2^{F(p_{(j,h,k)})}}.
\]

\item Case of $p_{(j,h,k')}\!\neq\!p_{(j,h,k)}$:\\
Two distinct paths with the same endpoints $(j,h)$ must diverge at a
\emph{split point} and later reconverge at a \emph{merge point}. At these
points, the paths share the same building-block parameters but traverse
different edges. Depending on which alternative in assumption (ii) the
building block satisfies, the split or merge point therefore contributes a
factor with zero expectation, implying
\[
\mathbb{E}\!\left[\,\Omega^{*}(p_{(j,h,k')})\,\Omega(p_{(j,h,k)})\,\right] 
= 0.
\]
\end{enumerate}

Thus, only the identical-path terms ($k'\!=\!k$) remain, and
\[
\mathbb{E}[\,|g_{jh}|^2\,] =
\sum_{p_{(j,h,k)} \in P_{(j,h)}} \frac{1}{2^{F(p_{(j,h,k)})}}.
\]
\end{proof}

\begin{corollary}\label{app_cor:uniform_case}
If, in addition, the number of paths $|P_{(j,h)}|$ and 
the number of building blocks along each path $F(p_{(j,h,k)})$ 
are identical for all $(j,h)$, then
\[
\mathbb{E}\left[\,|g_{jh}|^2\,\right] = \frac{1}{n}\,.
\]
\end{corollary}

\begin{proof}
Equation~(\ref{app_eq:expected}) implies that all entries have the same expected
squared modulus. Because every row of the $n\!\times\!n$ unitary matrix $G$ has
unit squared norm, summing these equal expectations across a row gives 1;
therefore, each expectation is $1/n$.
\end{proof}

We verify the conditions of Proposition~\ref{app_prop:expected_form} for three
commonly used fundamental building blocks.
\begin{ex}\label{app_ex:psbs} PSBS
\begin{equation*}
T_{(\text{PSBS})}(\phi) = \frac{1}{\sqrt{2}}
\begin{pmatrix}
e^{i\phi} & i\\
ie^{i\phi} &1 \end{pmatrix} \, .
\end{equation*}
Each entry has squared magnitude $1/2$, 
and thus property (i) is satisfied.
For property (ii), the same calculation applies to either row at a merge point;
for example,
\[
\mathbb{E}\left[w^*_{01}\,w_{00}\right] 
= \frac{1}{2}\,\mathbb{E}\left[-i e^{i\phi}\right]
= \frac{-i}{2}\times\frac{1}{2\pi}\int_{0}^{2\pi} e^{i\phi}\, d\phi 
= 0\, .
\]
\end{ex}

\begin{ex}\label{app_ex:2p_mzi} 2-parameter MZI
\begin{equation*}
T_{(\text{MZI})}(\phi,\theta) = i e^{i\frac{\theta}{2}}
\begin{pmatrix}
e^{i\phi}\sin\frac{\theta}{2} &\cos\frac{\theta}{2}\\
e^{i\phi}\cos\frac{\theta}{2} &-\sin\frac{\theta}{2}
\end{pmatrix} \, .
\end{equation*}
Since 
$|w_{\alpha\beta}|^2\!=\!\sin^{2}(\theta/2)$ or $\cos^{2}(\theta/2)$
and $\theta\!\in\![0,\pi]$,
we obtain 
\begin{equation*}
\mathbb{E}\left[\,|w_{\alpha\beta}|^2 \,\right]=
\frac{1}{\pi}\int_{0}^{\pi} \frac{1}{2}(1\pm \cos\theta) d\theta =\frac{1}{2}\, ,
\end{equation*}
where $+$ and $-$ correspond to 
$\cos^{2}(\theta/2)$ and $\sin^{2}(\theta/2)$, respectively.
Thus, property (i) is satisfied.

At a merge point, the same calculation applies to either row; for example,
\[
w^{*}_{01}w_{00} = e^{i\phi}\sin\frac{\theta}{2}\cos\frac{\theta}{2}\, .
\]
Since $\phi\!\in\![0,2\pi)$ and $\theta\!\in\![0,\pi]$,
we have
\begin{align*}
\mathbb{E}\left[ w^{*}_{01}w_{00}\right] 
&= 
\left(\frac{1}{\pi}\int_0^{\pi}\sin\frac{\theta}{2}\cos\frac{\theta}{2}\,d\theta\right)
\left(\frac{1}{2\pi}\int_0^{2\pi}e^{i\phi}\,d\phi\right)\\
&= \frac{1}{\pi}\cdot 0 = 0\, .
\end{align*}
Thus, property (ii) is satisfied.
\end{ex}

\begin{ex}\label{app_ex:1p_mzi} 1-parameter MZI (BS-PS-BS)
\begin{equation*}
T_{(\text{1p-MZI})}(\phi) = i e^{i\frac{\phi}{2}}
\begin{pmatrix}
\sin\frac{\phi}{2} &\cos\frac{\phi}{2}\\
\cos\frac{\phi}{2} &-\sin\frac{\phi}{2}
\end{pmatrix} \, .
\end{equation*}
As in the two-parameter MZI case, property (i) is satisfied.

At a split or merge point, 
\[
w^{*}_{10}w_{00} = w^{*}_{01}w_{00} = \sin\frac{\phi}{2}\cos\frac{\phi}{2}\,.
\]
Taking expectation,
\[
\mathbb{E}\left[w^{*}_{10}\,w_{00}\right] = 
\mathbb{E}\left[w^{*}_{01}\,w_{00}\right] = 
\frac{1}{2\pi}\int_{0}^{2\pi}\frac{1}{2}\sin\phi\, d\phi = 0\,.
\]
Thus, property (ii) is satisfied.
\end{ex}

Reverse-order configurations of the fundamental building blocks
in Examples~\ref{app_ex:psbs}--\ref{app_ex:1p_mzi}
implement the transposes of the corresponding $2\!\times\!2$ unitary matrices.
Since the transpose preserves the magnitudes and pairwise correlations of entries,
it also satisfies properties (i) and (ii).

\subsection{Relationship to Haar Initialization}\label{app_ssec:haar_init}
We relate the graph-theoretic approach to the Haar initialization method
of Ref.~\cite{pai}.

\begin{proposition}\label{app_prop:relate_haar_init}
Consider an $n\!\times\!n$ unitary matrix $G$ implemented by an MZI-based
rectangular mesh architecture. Assume that $\theta_{(s,u)}\!\in\![0,\pi]$ has
the following probability density:
\begin{align}
P_{\alpha_{(s,u)}}\!\left( \frac{\theta_{(s,u)}}{2} \right) 
&=\alpha_{(s,u)}
\sin\!\left( \frac{\theta_{(s,u)}}{2}\right)
\cos^{2\alpha_{(s,u)}-1}\!\left( \frac{\theta_{(s,u)}}{2}\right)\, ,
\label{app_eq:pai_pdf} \\
\alpha_{(s,u)} &= |I_{(s,u)}|+|O_{(s,u)}|-n-1\, , \nonumber
\end{align}
where $\theta_{(s,u)}$ is the phase angle of the internal phase shifter in
MZI$_{(s,u)}$ at stage $s$ and position $u$, and $\alpha_{(s,u)}$ is the
sensitivity index defined in Ref.~\cite{pai}. 
Following Ref.~\cite{pai}, $P_{\alpha_{(s,u)}}(\theta_{(s,u)}/2)$ denotes
the probability density of $\theta_{(s,u)}$, expressed in terms of its half-angle.
The quantities $|I_{(s,u)}|$ and $|O_{(s,u)}|$ denote, respectively, 
the numbers of input modes that can reach the MZI and output modes reachable from it. 
Then, the expected squared modulus of the entry $g_{jh}$ 
at the $j$th row and the $h$th column of $G$ is expressed by
\begin{equation}
\mathbb{E}\left[\,|g_{jh}|^2\,\right] = 
\sum_{\substack{p_{(j,h,k)}\\\in P_{(j,h)}}} \quad
\prod_{\substack{\text{MZI}_{(s,u)}\\\in p_{(j,h,k)}}}
\! v_{(s,u)}
\label{app_eq:expected_haarinit}
\end{equation}
\begin{equation}
v_{(s,u)}=
\begin{dcases}
	\frac{\alpha_{(s,u)}}{\alpha_{(s,u)}+1} & \text{for transmission}\\
	\frac{1}{\alpha_{(s,u)}+1} & \text{for reflection} \, .
\end{dcases}
\end{equation}
\end{proposition}

\begin{proof}
Following the proof of Proposition~\ref{app_prop:expected_form}, if the
fundamental building block is an MZI, the following equation holds
independently of the probability density of $\theta$ and is therefore not
restricted to a uniform distribution:
\begin{equation}
\mathbb{E}\left[\, |g_{jh}|^2\,\right] 
= \sum_{k}\mathbb{E}\left[\, \Omega^*(p_{(j,h,k)})\Omega(p_{(j,h,k)})\,\right]\,.
\label{app_eq:expected_iden_path}
\end{equation}
From Eq.~(\ref{app_eq:absOmega}),
\begin{equation*}
\Omega^{*}(p_{(j,h,k)})\,\Omega(p_{(j,h,k)}) =
\prod_{s=0}^{m-1}\left( \psi_{(s,u)}
\left( \frac{\theta_{(s,u)}}{2} \right)\right)^2\, ,
\end{equation*}
where $m$ is the stage count on $p_{(j,h,k)}$ 
and $\psi(\cdot)$ denotes $\cos(\cdot)$ for transmission through
MZI$_{(s,u)}$ and $\sin(\cdot)$ for reflection at that MZI.
Because the parameters $\theta_{(s,u)}$ are mutually independent,
\begin{align}
&\mathbb{E}\left[\,\Omega^{*}(p_{(j,h,k)})\,\Omega(p_{(j,h,k)})\,\right]
\nonumber\\
&\quad =\prod_{s=0}^{m-1}\mathbb{E}\left[\left( \psi_{(s,u)}
\left( \frac{\theta_{(s,u)}}{2} \right)\right)^2\right]\, .
\label{app_eq:expect_prod}
\end{align}
First, we consider the case of $\psi_{(s,u)}(\cdot)\!=\!\cos(\cdot)$.
\begin{align}
\mathbb{E}\left[\,\cos^{2}(\theta/2)\,\right]
&=\alpha\int_0^{\pi}\sin\frac{\theta}{2}
\left( \cos\frac{\theta}{2} \right)^{2\alpha-1}
\cos^{2}\frac{\theta}{2}\, d\theta \nonumber\\
&=\frac{\alpha}{2^{\alpha+1}}
\int_0^{\pi}(1+\cos\theta)^{\alpha}\sin\theta\, d\theta \nonumber\\
&=\frac{\alpha}{2^{\alpha+1}}\int_0^2 u^{\alpha}\, du\,
\quad(\leftarrow\, u=1\!+\cos\theta)\nonumber\\
&=\frac{\alpha}{\alpha+1} \, ,
\label{app_eq:expect_cos}
\end{align}
where the subscript $(s,u)$ is omitted for simplicity.
We similarly obtain $\mathbb{E}[\,\sin^{2}(\theta/2)\,]$ as
\begin{align}
\mathbb{E} \left[\,\sin^{2}(\theta/2)\,\right]
&=1-\mathbb{E} \left[\,\cos^{2}(\theta/2)\,\right]\nonumber\\
&=\frac{1}{\alpha+1}\, .
\label{app_eq:expect_sin}
\end{align}
Substituting Eqs.~(\ref{app_eq:expect_cos}) and (\ref{app_eq:expect_sin})
into Eqs.~(\ref{app_eq:expect_prod}) and (\ref{app_eq:expected_iden_path}) yields 
Eq.~(\ref{app_eq:expected_haarinit}).
\end{proof}

\section{Butterfly Topologies}\label{app_sec:bfly}
\begin{figure}[t]
\begin{center}
	\psfrag{0}[c][c][.8]{0} \psfrag{1}[c][c][.8]{1} \psfrag{2}[c][c][.8]{2}
	\psfrag{3}[c][c][.8]{3} \psfrag{4}[c][c][.8]{4} \psfrag{5}[c][c][.8]{5}
	\psfrag{6}[c][c][.8]{6} \psfrag{7}[c][c][.8]{7} \psfrag{8}[c][c][.8]{8}
	\psfrag{9}[c][c][.8]{9} \psfrag{10}[c][c][.8]{10} \psfrag{11}[c][c][.8]{11}
	\psfrag{12}[c][c][.8]{12} \psfrag{13}[c][c][.8]{13} \psfrag{14}[c][c][.8]{14}
	\psfrag{15}[c][c][.8]{15} 
	\psfrag{A}[c][c][.72]{000} \psfrag{B}[c][c][.72]{001}
	\psfrag{C}[c][c][.72]{010} \psfrag{D}[c][c][.72]{011}
	\psfrag{E}[c][c][.72]{100} \psfrag{F}[c][c][.72]{101}
	\psfrag{G}[c][c][.72]{110} \psfrag{H}[c][c][.72]{111}
	\psfrag{a}[c][c][.6]{0000} \psfrag{b}[c][c][.6]{0001}
	\psfrag{c}[c][c][.6]{0010} \psfrag{d}[c][c][.6]{0011}
	\psfrag{e}[c][c][.6]{0100} \psfrag{k}[c][c][.6]{0101}
	\psfrag{m}[c][c][.6]{0110} \psfrag{n}[c][c][.6]{0111}
	\psfrag{o}[c][c][.6]{1000} \psfrag{p}[c][c][.6]{1001}
	\psfrag{q}[c][c][.6]{1010} \psfrag{r}[c][c][.6]{1011}
	\psfrag{s}[c][c][.6]{1100} \psfrag{t}[c][c][.6]{1101}
	\psfrag{u}[c][c][.6]{1110} \psfrag{v}[c][c][.6]{1111}
	\psfrag{K}[c][c][.80]{Stage~0} \psfrag{L}[c][c][.80]{Stage~1}
	\psfrag{M}[c][c][.80]{Stage~2} \psfrag{N}[c][c][.80]{Stage~3}
	\includegraphics[width=68mm]{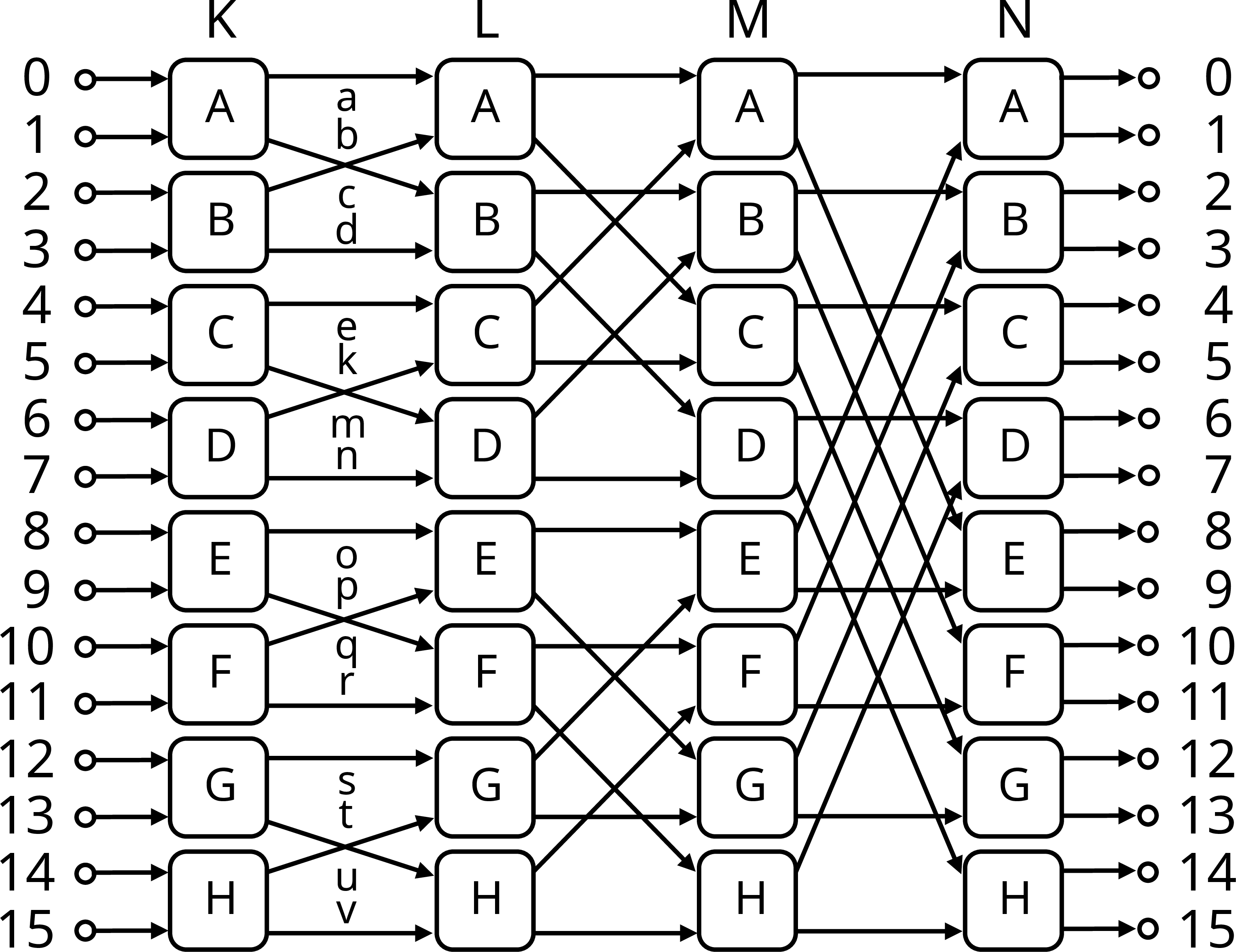}
\end{center}
\vspace*{-2mm}
\caption{Standard butterfly topology with 16 input terminals and
four stages.
Rounded squares represent functional nodes, and directed edges connect output
terminals in one stage to input terminals in the next.
}\label{app_fig:std_bfly}
\end{figure}
We first introduce the components in the butterfly topology and 
establish the notation.
We then define the standard butterfly topology 
in a manner consistent with its usage in interconnection networks, 
as described in \cite{wu}.
\begin{dfn}[Butterfly topology components]
A butterfly topology consists of \emph{functional nodes}, each with two input
and two output terminals, and directed edges that connect
nodes from one stage to the next stage.
Each functional node performs a specific function, such as 
a switch, interferometer, or mixer.
Given $n$ input terminals, where $n\!=\!2^t$ for some $t\in \mathbb{N}$,
each stage contains $2^{t-1}$ functional nodes.
Each node in the $s$th stage is assigned 
a {\em bit-representation identification number} (BID) of the form 
\[(b_{t-1},\ldots,b_k,\ldots,b_1)_s,\]
where $b_k\!\in\!\{0,\,1\}$ and $k\!=\!1,2,\ldots,t\!-\!1$
and 
\[ s\in\{0,1,\ldots,\log_2 n-1\}. \]
To label the two input and output terminals of a node, a least-significant bit
$b_0\!\in\!\{0,1\}$ is appended to the BID together with a superscript that
indicates the terminal type, either \textsuperscript{[in]} or
\textsuperscript{[out]}.
The resulting extended label takes the form 
\[(b_{t-1},\ldots,b_k,\ldots,b_1,b_0)_s^{[\ast]},\]
where $b_0\!=\!0$ and $b_0\!=\!1$ correspond to the upper and lower terminals,
respectively, and $\ast\!\in\!\{\text{in},\text{out}\}$ specifies the terminal type.
\label{app_dfn:component}
\end{dfn}
\begin{dfn}[Standard butterfly topology]
A {\em standard butterfly topology} is defined by connecting pairs of output
and input terminals between adjacent stages $s$ and $s+1$, for
$s\!\in\!\{0,1,\ldots,t\!-\!2\}$, according to the following rule:
\begin{align*}
&(b_{t-1},\ldots,b_{s+2},b_{s+1},b_s,\ldots,b_1,b_0)_s^{\text{[out]}}\quad\text{and}\\
&(b_{t-1},\ldots,b_{s+2},b_0,b_s,\ldots,b_1,b_{s+1})_{s+1}^{\text{[in]}}\,.
\end{align*}
The topology connects each output terminal at stage $s$ to an input terminal 
at stage ($s\!+\!1$) by swapping the terminal bit $b_{0}$ and the stride bit $b_{s+1}$ 
in the BID. 
In this notation, 
$b_{0}$ and $b_{s+1}$ are referred to as the terminal bit and {\em stride bit}, 
respectively. 
The stride at the $s$th stage is defined as $2^s$ and 
the stride bit indicates that the current node is connected to a node 
in the next stage located $2^s$ positions away if $b_{s+1}\!\neq\!b_0$.
This structured swapping defines the recursive interstage wiring pattern characteristic
of the standard butterfly topology.
\label{app_dfn:std_bfly}
\end{dfn}

Figure~\ref{app_fig:std_bfly} illustrates a standard butterfly topology 
with 16 input terminals 0 through 15.
Each rounded square represents a functional node, 
equipped with two input and two output terminals. 
The BIDs of the nodes range from (000) to (111), 
arranged from top to bottom.
Each BID encodes the position of a node within a stage using $(t\!-\!1)$ bits,
where $t\!=\!\log_2 n$ for $n$ input terminals.
For example, with $n\!=\!16$, we have $t\!=\!4$,
and nodes at each stage are labeled with 3-bit BIDs representing binary indices from 0 to 7. 
The BID reflects the vertical position of a node, 
with the most significant bit at the leftmost position.
The directed edges depict connections from nodes at one stage to those at the
next.
In stage 0, the outgoing edges are labeled using the same BIDs as their corresponding 
output terminals.
The interstage connections follow the rule specified
in Definition~\ref{app_dfn:std_bfly}. 
For example, the output terminals (0010) and (0011) at stage 0 
are connected to the input terminals (0001) and (0011) at stage 1, 
respectively. 
This corresponds to swapping the terminal bit $b_0$ with the stride bit $b_{s+1}$ 
as outlined in the standard butterfly topology.

\begin{dfn}[Butterfly topology family]
A butterfly-topology family contains the topologies that are equivalent to the
standard butterfly topology under the transformations described in Ref.~\cite{wu}.
\label{app_dfn:family}
\end{dfn}

\section{Preprocessing}\label{app_sec:preproc}
\begin{figure*}[t]
\begin{center}
	\includegraphics[width=144mm]{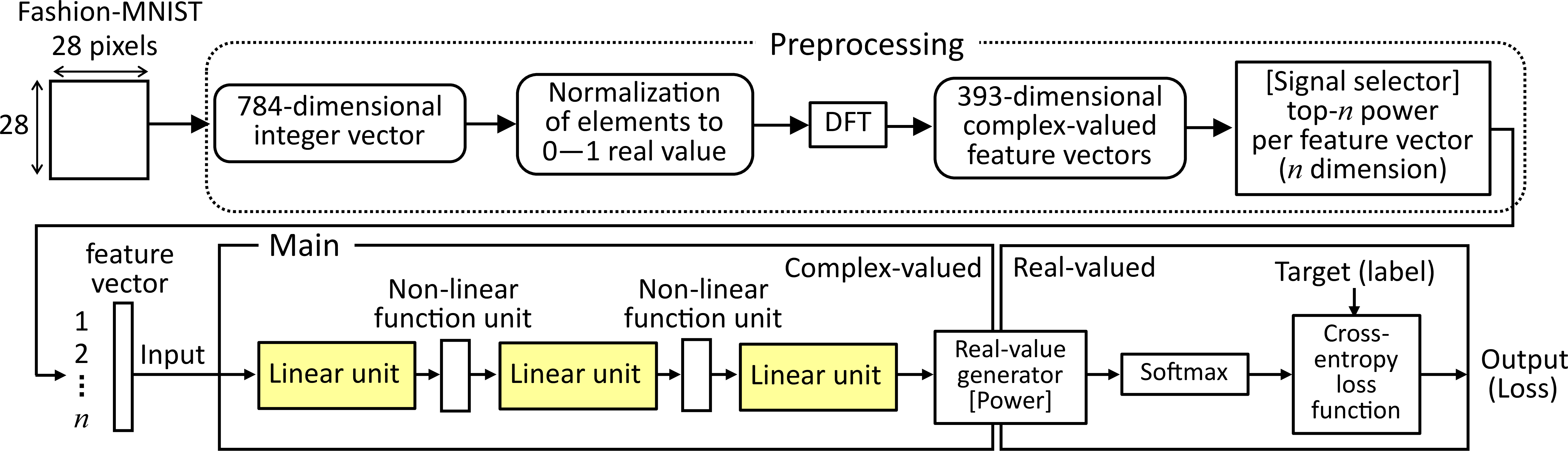}
\end{center}
\caption{Evaluation system for three-layer feedforward neural networks.}
\label{app_fig:fnn_syst}
\end{figure*}

\begin{figure*}[t]
\begin{center}
    \includegraphics[width=145mm]{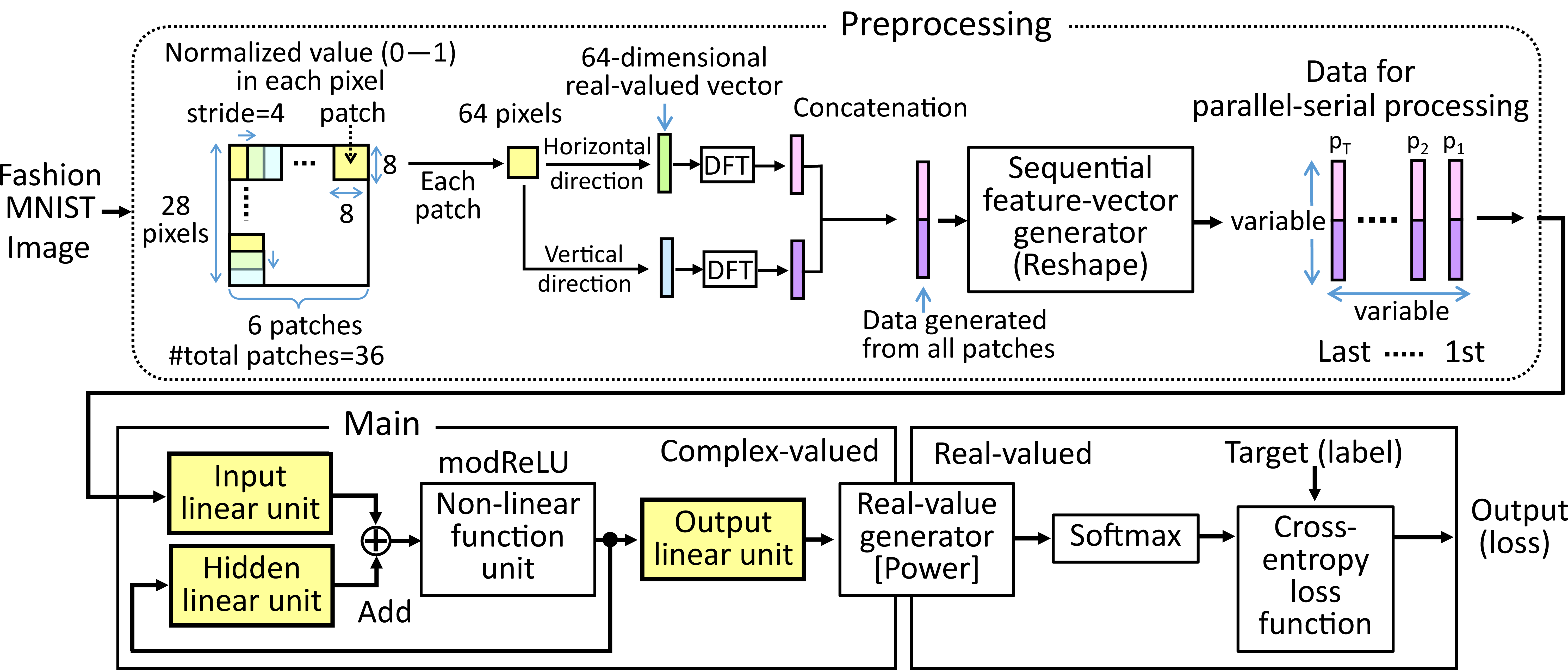}
\end{center}
\caption{Evaluation system for recurrent neural networks.}
\label{app_fig:rnn_syst}
\end{figure*}

We describe the preprocessing schemes used for 
the feedforward and recurrent neural networks (FNNs and RNNs).
Although different preprocessing methods are employed for these two settings,
they are designed based on a common set of principles 
to ensure fair and physically meaningful evaluation of the architectures.
First, the preprocessing 
is selected to be compatible with near-term optical implementations,
primarily relying on Fourier transforms that can be realized efficiently 
in photonic circuits.
Second, the preprocessing 
avoids excessive feature engineering so that the performance of
the classification task reflects the capability of the main linear architectures
rather than the preprocessing itself.
Third, the preprocessing produces lower-dimensional input representations whose
dimensionality matches the number of input ports of the linear units in the main
blocks.
Based on these principles,
we employ distinct preprocessing methods for FNNs and RNNs,
which are described in the following subsections.
 
\subsection{Preprocessing for Feedforward Neural Networks}\label{ssec:preprocFNNs}
Figure~\ref{app_fig:fnn_syst} illustrates the evaluation system for
three-layer feedforward neural networks (FNNs) 
used in the Fashion-MNIST classification task described in Section~\ref{ssec:fnn}.
The upper and lower blocks correspond to the preprocessing and main blocks,
respectively.
Each input image consists of $28\!\times\!28$ pixels, 
with pixel values represented as integers in the range $[0,255]$.
The image is reshaped into a 784-dimensional vector and normalized to
a real-valued vector in the range $[0,1]$.
The normalized vector is then transformed using the discrete Fourier transform (DFT),
yielding a 393-dimensional complex-valued feature vector comprising one DC
component, one Nyquist component, and 391 nonredundant positive-frequency components.
From the resulting DFT outputs,
$n$ components with the highest average power over the training dataset are selected, 
where $n$ matches the number of input ports in the linear units in the main block.
This selection ensures dimensional compatibility 
while preserving dominant spectral features.
Alternatively, from the perspective of physical implementation,
the feature vector may be constructed by selecting the $n$ lowest-frequency components.
This preprocessing maintains a balance between physical realizability and 
minimal feature engineering, 
allowing the performance differences to be attributed primarily to the main architectures.

\subsection{Preprocessing for Recurrent Neural Networks}\label{ssec:preprocRNNs}
Figure~\ref{app_fig:rnn_syst} illustrates the evaluation system for 
recurrent neural networks (RNNs) described in Section~\ref{ssec:rnn}.
The upper and lower blocks correspond to the preprocessing and main blocks, 
respectively.
Each input image consists of $28\!\times\!28$ pixels, 
with pixel values normalized to the range $[0,1]$.
From each image, 36 overlapping patches are extracted using a sliding window 
of size $8\!\times\!8$ with a stride of 4.
Each patch is vectorized in two scan orders, horizontal and vertical, to form
two 64-dimensional vectors.
Each vector is then transformed using the DFT, 
producing a 32-dimensional complex-valued feature vector
that includes one DC component and 31 positive-frequency components
(with the Nyquist component removed).
Concatenating the two transformed vectors yields a 64-dimensional complex-valued 
feature vector for each patch.
Applying this procedure to all patches produces a sequence of 36 vectors in
$\mathbb{C}^{64}$ for each input image.
The dimensionality and length of the sequence can be adjusted 
according to the number of input ports in the linear units of the main block.
In Section~\ref{ssec:rnn}, we use three (dimensionality, sequence-length)
configurations: $(64,36)$, $(128,18)$, and $(256,9)$.
This preprocessing enables sequential representations 
while maintaining compatibility with Fourier-based optical implementations.

\makeatletter
\@namedef{b@PRAppliedControl}{}
\renewcommand{\href}[2]{#2}
\bibliographystyle{apsrev4-2}

\end{document}